\documentclass[nohyperref]{article}

\usepackage{microtype}
\usepackage{graphicx}
\usepackage{subcaption}
\usepackage{booktabs} %
\usepackage{multirow}

\usepackage{tikz}
\tikzset{>=latex} %
\usepackage{pgfplots} %
\usepackage{xcolor}
\usepackage[outline]{contour} %
\contourlength{1.2pt}
\usetikzlibrary{positioning,calc}
\usetikzlibrary{backgrounds}%
\usepackage{hyperref}

\usepackage{colortbl}

\usepackage[accepted]{icml2023}

\usepackage{amsmath}
\usepackage{amssymb}
\usepackage{mathtools}
\usepackage{amsthm}

\usepackage[capitalize,noabbrev]{cleveref}
\Crefname{section}{\mbox{\S\hspace*{-0.25ex}}}{\mbox{\S\hspace*{-0.25ex}}}
\Crefname{equation}{Eq.}{Eqs.}
\Crefname{figure}{Fig.}{Figs.}
\Crefname{table}{Tab.}{Tabs.}
\Crefname{appendix}{\S$\!$}{\S$\!$}

\usepackage{caption}
\usepackage{enumitem} %
\setlist[itemize]{noitemsep,topsep=0ex}    %
\setlist{leftmargin=3mm}

\newcolumntype{P}[1]{>{\raggedright\arraybackslash}p{#1}}

\usepackage{alexdefs}
\usepackage[light,scaled=0.85]{roboto-mono}         %

\theoremstyle{plain}
\newtheorem{theorem}{Theorem}[section]

\theoremstyle{definition}
\newtheorem{definition}[theorem]{Definition}

\theoremstyle{remark}

\newtheorem*{rep@theorem}{\rep@title}
\newcommand{\newreptheorem}[2]{%
\newenvironment{rep#1}[1]{%
 \def\rep@title{#2 \ref{##1}}%
 \begin{rep@theorem}}%
 {\end{rep@theorem}}}
\makeatother
\newreptheorem{proposition}{Proposition}
\newreptheorem{theorem}{Theorem}

\usepackage[disable,textsize=tiny]{todonotes}
\newcommand{\D}{\mathcal{D}}
\newcommand{\Norm}{N}
\newcommand{\Exp}{\mathrm{Exp}}
\newcommand{\Skew}{\mathrm{Skew}}
\newcommand{\Df}{\D^f}
\newcommand{\Dgt}{\D^{gt}}
\newcommand{\metric}{S}

\newcommand{\nf}{m}
\newcommand{\ngt}{n}
\newcommand{\dvar}{d}
\newcommand{\prob}[1]{\mathbb{P}\left[#1\right]}
\newcommand{\ex}[2]{\mathbb{E}_{#1}\left[#2\right]}
\newcommand{\variance}[2]{\operatorname{Var}_{#1}\left[#2\right]}
\newcommand{\ror}[1]{\texttt{RoR}$_{#1}$}

\newcommand{\paragraphtight}[1]{\textbf{#1~~~}}

\makeatletter
\def\Cline#1#2{\@Cline#1#2\@nil}
\def\@Cline#1-#2#3\@nil{%
  \omit
  \@multicnt#1%
  \advance\@multispan\m@ne
  \ifnum\@multicnt=\@ne\@firstofone{&\omit}\fi
  \@multicnt#2%
  \advance\@multicnt-#1%
  \advance\@multispan\@ne
  \leaders\hrule\@height#3\hfill
  \cr}
\makeatother

\usepackage[subtle]{savetrees}

\icmltitlerunning{Regions of Reliability in the Evaluation of Multivariate Probabilistic Forecasts}

\begin{document}

\twocolumn[
\icmltitle{Regions of Reliability in the Evaluation of Multivariate Probabilistic Forecasts}

\icmlsetsymbol{equal}{*}

\begin{icmlauthorlist}
\icmlauthor{Étienne Marcotte}{snow}
\icmlauthor{Valentina Zantedeschi}{snow}
\icmlauthor{Alexandre Drouin}{snow}
\icmlauthor{Nicolas Chapados}{snow}
\end{icmlauthorlist}

\icmlaffiliation{snow}{ServiceNow Research, Montréal, Canada}

\icmlcorrespondingauthor{Étienne Marcotte}{etienne.marcotte@servicenow.com}

\icmlkeywords{Machine Learning, ICML}

\vskip 0.3in
]

\printAffiliationsAndNotice{}  %

\begin{abstract}
Multivariate probabilistic time series forecasts are commonly evaluated via proper scoring rules, i.e., functions that are minimal in expectation for the ground-truth distribution. 
However, this property is not sufficient to guarantee good discrimination in the non-asymptotic regime. 
In this paper, we provide the first systematic finite-sample study of proper scoring rules for time-series forecasting evaluation.
Through a power analysis,
we identify the ``region of reliability'' of a scoring rule, i.e., the set of practical conditions where it can be relied on to identify forecasting errors.
We carry out our analysis on a comprehensive synthetic benchmark, specifically designed to test several key discrepancies between ground-truth and forecast distributions, and we gauge the generalizability of our findings to real-world tasks with an application to an electricity production problem.
Our results reveal critical shortcomings in the evaluation of multivariate probabilistic forecasts as commonly performed in the literature.
\end{abstract}

\section{Introduction}

The forecasting of time-varying quantities is a fundamental component of decision-making in fields like economics, operation management, and healthcare~\cite{peterson2017introduction,heizer2023operations}. 
In this context, a proper characterization of uncertainty is key to reasoning about potential futures and their respective likelihood.
This has motivated the problem of \emph{multivariate probabilistic forecasting}, which consists in estimating the joint distribution of the future values of quantities of interest~\cite{gneiting2014probabilistic}.

In this setting, all estimates are not equal.
Depending on the application, certain kinds of errors, e.g., failures to properly model statistical dependencies between variables, can have a catastrophic impact on downstream decisions.
It is thus critical to develop methodological tools to assess the quality of distributional forecasts produced by statistical models.

For this, the literature has primarily focused on developing \emph{proper scoring rules}~\cite{gneiting2007strictly}, which are designed to reach a minimum when the forecast and the ground-truth distributions match.
Among them, the \emph{Negative Log-Likelihood} has been shown to be an optimal discriminator of erroneous distributions~\cite{neyman1933ix}.
However, the use of the likelihood is not always practical since many models do not allow for its efficient calculation.
Hence, time-series practitioners have turned to other proper scoring rules, such as the \emph{Continuous Ranked Probability Score} (CRPS; \citealt{matheson1976scoring}) to evaluate forecasts.
While proper in theory, the discriminative performance of such scoring rules in practical conditions, where the dimensionality of the problem is large and the sample size is small in comparison, is poorly understood.

This work aims to study the reliability of proper scoring rules for the evaluation of multivariate probabilistic forecasts in realistic finite-sample settings.
We quantify reliability as the statistical power of a rule at discriminating between data sampled respectively from the ground truth and an erroneous forecast.
We introduce a comprehensive benchmark to systematically measure the ability of scoring rules to detect failures in forecasts of practical interest.
Our results emphasize sets of conditions (in terms of problem dimensionality and Monte Carlo approximation quality) under which each scoring rule is reliable, dubbed \emph{regions of reliability},\footnote{We note that our regions of reliability are distinct from \emph{reliability diagrams}, which are graphical tools for checking forecast calibration.} and, most interestingly, reveal significant shortcomings such as the general inability of the studied scoring rules at detecting some basic forecasting errors.

\paragraphtight{Contributions:}
\begin{itemize}
    \item We propose a methodology, based on power analysis, to assess the reliability of proper scoring rules in the evaluation of multivariate probabilistic forecasts (\cref{sec:ror});
    \item We propose an extensive benchmark that reveals \emph{regions of reliability} for five common proper scoring rules and 19~types of forecasting errors (\cref{sec:benchmark});
    \item We show that our findings generalize to three real-world settings beyond this benchmark (\cref{sec:real-data});
    \item We present a critical review of experimental practices in the recent literature in light of these results and make recommendations for future developments in the field (\cref{sec:conclusion}).
\end{itemize}

\section{Background}
\label{sec:background}

\subsection{Multivariate Probabilistic Forecasting}
\label{subsec:multivariate-prob-fcast}

We consider the problem of discrete-time probabilistic forecasting of multivariate time series, where we seek to accurately estimate the joint distribution of some numerical quantities of interest over a future time horizon, given historical data.
Formally, let $X_t \in \reals^v$ be a random vector containing the values of $v$ variables at time $t$. The problem consists in accurately estimating the joint conditional distribution
\begin{equation}\label{eq:mvarforecast}
    P\left(  X_{t + 1},\,\,\ldots,\,\,X_{T} \mid X_1,\,\,\ldots,\,\,X_t  \right),
\end{equation}
where $X_1,\,\,\dots,\,\,X_t$ are past observations of the variables and $X_{t + 1},\,\,\dots,\,\,X_T$ are forecasted values up to time $T$. 
This is in contrast with point and quantile forecasting, which typically aim at estimating the conditional mean or quantiles of this distribution, respectively~\cite{West1997,cai2002quantiles}.

\subsection{Motivating Example in Decision-Making}\label{sec:motivating-example}

Many problems can only be effectively solved by accurately estimating \cref{eq:mvarforecast}.
Consider for instance the following real-world decision-making task, inspired by the \texttt{solar} dataset~\citep{godahewa2021monash}.
Suppose that one manages a network of solar power plants and that, one day in advance, needs to decide (i)~which plants to shut down for maintenance, and (ii)~how much electricity to commit to selling at every hour.
Any shortage in production would result in a hefty fine, which implies that just predicting the expectation of \cref{eq:mvarforecast} is not sufficient for the task.
Consider the following formalization of the problem:
\begin{equation}\label{eq:motivating}
\begin{aligned}
\max_{a_i,s_t} \quad & \Esp_{p \sim P}\left[\sum_{t} \biggl( s_t - \lambda \ \max\Bigl(0, \; s_t - \sum_i a_i p_{it} \Bigr)\biggr)\right],\\[-0.25ex]
\textrm{s.t.} \quad & \sum_i a_i \le M,~~~a_i \in \{0, 1\},~~~s_t \geq 0,
\end{aligned}
\end{equation}
where $a_i = 1$ indicates that plant $i \in \{1,\ldots,N\}$ is active (with at most $M$ active at once), $s_t$ is our production commitment for period $t \in \{1,\ldots,T\}$, $P_{it}$ is the distribution of electricity produced by plant $i$ during period $t$ (to be forecasted), $P$ represents the multivariate distribution for all plants and periods, and $\lambda > 1$ is the penalty factor for not delivering the promised electricity.%

Given a perfect estimate of \cref{eq:mvarforecast}, i.e., the future power production for each station, one could optimally solve this problem.
However, in practice, this distribution is unknown and must be estimated using imperfect forecasting models;
some imperfections could lead to critically bad solutions. %
For instance, if the forecast does not capture the statistical associations between stations, one could decide to only activate stations in one geographical area, making the total production vulnerable to local weather events.

It is thus essential to develop methodological tools to holistically assess the quality of probabilistic forecasts, not limited to their marginals or expectations. 
However, it is not straightforward to carry out such an evaluation as it requires knowledge of both the forecast and ground-truth distributions.
As the latter is, in practice, unknown, the community has turned to \emph{proper scoring rules}, i.e., evaluation criteria that only require samples/observations from these distributions.

\subsection{Proper Scoring Rules}
Denote $\Dgt$ the ground-truth process with \cref{eq:mvarforecast} as probability density function and $\Df$ an arbitrary forecast distribution over the same domain.
A \emph{scoring rule} is a function $\metric(y, \Df)$ that measures the loss incurred if event $y \sim \Dgt$ realizes under forecast $\Df$, assessing, for instance, how unlikely event $y$ is according to $\Df$.
An example of a scoring rule is the \emph{Energy Score}~\cite{gneiting2007strictly}, defined as
\begin{equation}
    \textrm{ES}(y,\Df) = 
        \Esp_{x \sim \Df}\|y - x\|_2^{p} - 
        \frac{1}{2} \Esp_{\substack{x \sim \Df\\x' \sim \Df}} \|x - x'\|_2^{p},
\end{equation}
with $\|\!\!\cdot\!\!\|_2$ the Euclidean norm and $p \in (0, 2)$ a parameter.

In general a scoring rule $S$ must obey some basic regularity conditions, such as being \emph{proper}, i.e., given a ground-truth distribution $\Dgt$ for any forecast $\Df$, we must have
\begin{equation}
\Esp_{y \sim \Dgt}[S(y,\Dgt)] \;\;\leq\;\; \Esp_{y \sim \Dgt}[S(y,\Df)].
\end{equation}
In other words, the scoring rule must achieve its minimum, in expectation over all realizations of $\Dgt$, for a forecast that perfectly matches the ground truth.
Further, a scoring rule is said to be \emph{strictly proper} if the minimum is unique.

While it is a necessary condition for a scoring rule to be minimal in the ground truth, 
it is important to note that being proper does not imply that a scoring rule will be able to detect any prediction flaw.
Several failure modes have been pointed out in the literature (e.g., \citet{hamill2001interpretation,gneiting2007probabilistic}), but 
an understudied issue is the behavior of proper scoring rules in the finite-sample regime.
Indeed, properness only holds in expectation, while in practice evaluation is conducted based on a few samples from $\Dgt$ (e.g., in a rolling window evaluation).
In what follows, we raise concerns about the practical reliability of common proper scoring rules, showing that, in realistic experimental conditions, they fail to distinguish between the ground-truth distribution and forecasts with significant imperfections.

\section{Related Works}

Early concern with forecast evaluation measures was driven in large part by the first forecasting competitions, such as the influential M-competitions~\cite{Makridakis:1979tt, Makridakis:1982we, Makridakis:2000tj}, which prior to M4~\cite{Makridakis:2018em} only focused on univariate point accuracy measures.
Such measures include the mean absolute, squared, and absolute percentage errors~\cite{Mahmoud:1984uy}.
The mean absolute scaled error was also proposed by~\citet{Hyndman:2006dt} as an improved scale-invariant measure. Interval forecasts~\cite{Chatfield:1993tm,Chatfield2001} are a mid-point between point and full probabilistic forecasts, and were part of the M4 and M5 competitions~\cite{Hewamalage2021look,Makridakis2022:uncertainty}.

\paragraphtight{Development of Scoring Rules for Stochastic Forecasts}
Given that point and interval forecasting accuracy metrics are incomplete for probabilistic assessments, the literature considered alternatives such as the Continuous Ranked Probability Score (CRPS)~\cite{matheson1976scoring,Winkler1996}, and introduced the Energy Score (ES)~\cite{gneiting2007strictly,gneiting2008assessing} as a multivariate generalization of the CRPS. \citet{Gneiting2011} studied weighting schemes for the CRPS aimed at improving its tail behavior. As limitations of these rules became better understood, new ones, such as the Variogram \cite{Scheuerer2015}, which is popular in the weather forecasting literature, have emerged. \citet{salinas2019high} introduced the CRPS-Sum as a simple scoring rule for multivariate time series.
In another direction, \citet{Ziel2019} proposed to split the forecast distribution into its marginals and a copula, allowing the use of specialized scoring rules on each component.

\paragraphtight{Assessment of Proper Scoring Rules}
Although a scoring rule cannot give a total ranking of forecasts for all possible applications of the target quantities \cite{Diebold1998}, much interest has been given to the specific discrimination abilities of particular scoring rules. In a univariate setting, \citet{hamill2001interpretation} illustrated a case where the rank histogram is uniform, yet every probabilistic forecast is biased; \citet{gneiting2007probabilistic} studied several rules in light of a proposed tuning approach to multivariate forecasts. \citet{Bao2007} used the Kullback-Leibler information criterion to compare univariate probabilistic forecasting models. 
\citet{Pinson2013} explored how well the Energy Score can distinguish between two bivariate Gaussian distributions, and gave a theoretical bound on how it fares as the number of variables increases.
\citet{Alexander2022} compared the Energy Score and the Variogram on multiple distributions built from real-world data. However, their work does not explore the impact of dimensionality and sampling size on robustness.
The effectiveness of the CRPS-Sum has been studied by \citet{Koochali2022}, who showed how it can fail to distinguish between naive and state-of-the-art forecasts on real-world data. 
In these studies, a systematic assessment of forecasting rules against known deviations in the finite-sample regime is lacking, preventing a clear understanding of their limitations. This is what the present work seeks to address.

\section{Regions of Reliability}\label{sec:ror}

As exemplified in our motivating example (\cref{sec:motivating-example}), the detection of certain discrepancies between forecasts and ground truth, e.g., adequately capturing the correlation structure, is of high practical importance. Nonetheless, the conditions in which proper scoring rules are used in practice do not always allow for the detection of such discrepancies. 
In what follows, we devise a methodology based on power analysis~(e.g.,~\citet{cohen1992statistical}) to assess the reliability of a scoring rule at evaluating forecasts.
Aside from being a well-established framework for carrying out significance tests, power analysis has the advantage of providing power laws for the minimal size of the ground-truth sample sufficient for obtaining a certain discriminatory power, which would have to be assessed empirically otherwise.

\subsection{Measuring Reliability via Power Analysis}\label{sec:power}

To quantify the reliability of a scoring rule, we conduct a power analysis that tests whether the rule can discriminate an incorrect forecast distribution from the ground-truth distribution, given samples from both.
That is, we measure the statistical power of the scoring rule on the task of rejecting the null hypothesis that the forecast and the ground truth are statistically indistinguishable
at a chosen significance level.
Note that, contrary to common applications of power analysis, we seek to assess the ability of a scoring rule to detect a known effect, rather than detecting a purported effect with a scoring rule that is known to be reliable.

Formally, consider a pair of ground-truth $\Dgt$ and forecast $\Df$ distributions over $d$ variables.
We examine the following random variable, which denotes the gap between ground truth and forecast according to the scoring function $\metric$,
\begin{equation}
    \Delta_\nf = \metric(y, \;X^{f}_\nf) - \metric(y, \;X^{gt}_\nf),
\end{equation}
with $y \sim \Dgt$ a realization of the ground truth, and $X^{gt}_\nf = \{x_i \sim \Dgt\}_{i=1}^\nf$ and $X^{f}_\nf = \{x_i \sim \Df\}_{i=1}^\nf$ random variables corresponding to samples of size $m$ of the ground-truth and forecast distributions, respectively.
We can empirically estimate the mean and variance of this random variable as
\begin{align}
    \mu_\nf = \ex{y, X^{gt}_\nf, X^{f}_\nf}{\Delta_\nf} &\approx \frac{1}{K} \sum_{k=1}^{K} \delta_\nf^{(k)}, \\[-1ex]
    \sigma_\nf^2 = \variance{y, X^{gt}_\nf, X^{f}_\nf}{\Delta_\nf} &\approx \frac{1}{K{-}1} \sum_{k=1}^{K} \left(\delta_\nf^{(k)} - \mu_\nf \right)^2, \label{eq:delta_variance}
\end{align}
where $\delta^{(k)}_\nf \sim \Delta_\nf$, which we measure through $K$ independent trials.~\footnote{In practice, we set $K=1000$.}
Our null hypothesis corresponds to assuming that the difference $\Delta_\nf$ has mean $\mu_\nf = 0$, i.e., the forecast is indistinguishable from the ground truth.

Instead of making assumptions on the distribution of $\Delta_\nf$,
we leverage the central limit theorem\footnote{We investigate how accurate this approximation is in \cref{app:validity_normal_assumption}.} and consider that the average of $\ngt$ independent replications of $\Delta_\nf$ (i.e., $\frac{1}{\ngt} \sum_{j=1}^{\ngt} \Delta_\nf^{(j)}$) approximately follows the normal distribution $H_\metric(\ngt, \nf) = \Norm\bigl(\mu_\nf, \sigma_\nf^2/\ngt\bigr)$, which we take to be the \emph{distribution under the alternative hypothesis}.
Similarly, we take $H_0(\ngt, \nf) = \Norm\bigl(0, \sigma_\nf^2/\ngt\bigr)$ to be the corresponding \emph{distribution under the null hypothesis}.
As illustrated in \cref{fig:power-scheme}, our power analysis boils down to studying how $H_0(\ngt, \nf)$ and $H_\metric(\ngt, \nf)$ overlap, i.e., the power of the scoring rule at setting them apart.

Concretely, we fix the level of significance to $\alpha=5\%$, i.e., the false positive rate (shaded blue area in \cref{fig:power-scheme}), and determine the corresponding critical value $t_\alpha \in \mathbb{R}^+$,
\begin{equation}
    \prob{H_0(\ngt, \nf) \geq t_\alpha} = \alpha.
\end{equation}
We then quantify the reliability of the scoring rule via its statistical power.
\begin{definition}[Statistical power]\label{def:power}
    The statistical power of a scoring rule $\metric$ is defined as its true positive rate given a significance level of $\alpha$ and is given by
\begin{equation}
    \prob{H_\metric(\ngt, \nf) \geq t_\alpha} = 1 - \beta,
\end{equation}
where $\beta$ is the false negative rate (shaded red area in \cref{fig:power-scheme}).
\end{definition}

\paragraphtight{Remarks:} Increasing the number of replications ($\ngt$) leads to a reduction in the variance of the distributions of $H_0$ and $H_S$, resulting in decreased overlap and thus, increased power (up to the perfect power of $1$ in the limit). In practice, this can be achieved to some extent by increasing the number of rolling windows used for evaluation in a time series setting (up to the limits of data availability). 
As for the sample size ($m$),
small values might inflate the variance estimated in \cref{eq:delta_variance}, resulting in increased overlap between $H_0$ and $H_S$ and reduced power. 
Finally, note that the dimensionality of the distributions ($d$) may degrade the ability of some scoring rules to detect discrepancies (e.g., due to the curse of dimensionality).

\pgfmathdeclarefunction{gauss}{3}{%
  \pgfmathparse{1/(#3*sqrt(2*pi))*exp(-((#1-#2)^2)/(2*#3^2))*0.62}%
}
\pgfmathdeclarefunction{cdf}{3}{%
  \pgfmathparse{1/(1+exp(-0.07056*((#1-#2)/#3)^3 - 1.5976*(#1-#2)/#3))}%
}
\pgfmathdeclarefunction{fq}{3}{%
  \pgfmathparse{1/(sqrt(2*pi*#1))*exp(-(sqrt(#1)-#2/#3)^2/2)}%
}
\pgfmathdeclarefunction{fq0}{1}{%
  \pgfmathparse{1/(sqrt(2*pi*#1))*exp(-#1/2))}%
}

\colorlet{mydarkblue}{blue!30!black}

\usepgfplotslibrary{fillbetween}
\usetikzlibrary{patterns}
\pgfplotsset{compat=1.12} %

\def\N{50}

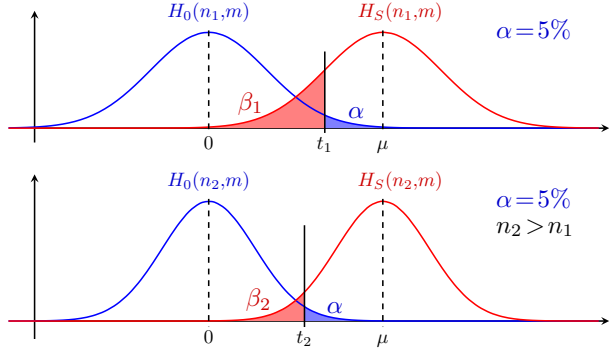
\begin{figure}
\resizebox{0.47\textwidth}{!}{
    \begin{tikzpicture}
      
      \def\q{5};
      \def\B{3};
      \def\S{6};
      \def\Bs{1.0};
      \def\Ss{1.0};
      \def\xmax{9};
      \def\ymin{{-0.15*gauss(\B,\B,\Bs)}};
      
      \begin{axis}[every axis plot post/.append style={
                   mark=none,domain={-0.05*(\xmax)}:{1.08*\xmax},samples=\N,smooth},
                   xmin={0}, xmax=\xmax,
                   ymin=\ymin, ymax={1.1*gauss(\B,\B,\Bs)},
                   axis lines=middle,
                   axis line style=thick,
                   enlargelimits=upper, %
                   ticks=none,
                   x label/.style={at={(current axis.right of origin)},anchor=north west},
                   width=0.7*\textwidth, height=0.5*\textwidth,
                   clip=false, %
                   y=200pt
                  ]

        \addplot[blue, name path=B,thick] {gauss(x,\B,\Bs)};
        \addplot[red,  name path=S,thick] {gauss(x,\S,\Ss)};
        \addplot[black,dashed,thick]
          coordinates {(\B, {1.02*gauss(\B,\B,\Bs)}) (\B,{-0.05*gauss(\B,\B,\Bs)})}
          node[below=-4pt] {\strut$0$};
        \addplot[black,dashed,thick]
          coordinates {(\S, {1.02*gauss(\S,\S,\Ss)}) (\S,{-0.05*gauss(\S,\S,\Ss)})}
          node[below=-4pt] {\strut$\mu$};
        \addplot[black,thick]
          coordinates {(\q, {0.80*gauss(\B,\B,\Bs)}) (\q,{-0.05*gauss(\B,\B,\Bs)})}
          node[below=-4pt] {\strut$t_1$};

        \path[name path=xaxis]
          (0,0) -- (\pgfkeysvalueof{/pgfplots/xmax},0); %
        \addplot[white!50!blue] fill between[of=xaxis and B, soft clip={domain=\q:\xmax}];
        \addplot[white!50!red]  fill between[of=xaxis and S, soft clip={domain=0:\q}];

        \node[above=2pt,black!20!blue] at (   \B,  {gauss(\B,\B,\Bs)}) {$H_0(\ngt_1,\nf)$};
        \node[above=2pt,black!20!red]  at (1.05*\S,{gauss(\S,\S,\Ss)}) {$H_\metric(\ngt_1, \nf)$};
        \node[right,black!20!blue,scale=1.3] at ({1.05*\q},{gauss(0.98*\q,\B,\Bs)}) {\strut$\alpha$};
        \node[right,black!20!blue,scale=1.3] at ({1.3*\S},{gauss(\B,\B,\Bs)}) {\strut$\alpha=5\%$};
        \node[right,black!20!red,scale=1.3] at ({0.67*\q},{gauss(0.87*\q,\S,\Ss)}) {\strut$\beta_1$};

      \end{axis}
    \end{tikzpicture}}
    \resizebox{0.47\textwidth}{!}{
    \begin{tikzpicture}
      
      \def\q{4.65};
      \def\B{3};
      \def\S{6};
      \def\Bs{0.8};
      \def\Ss{0.8};
      \def\xmax{9};
      \def\ymin{{-0.15*gauss(\B,\B,\Bs)}};
      
      \begin{axis}[every axis plot post/.append style={
                   mark=none,domain={-0.05*(\xmax)}:{1.08*\xmax},samples=\N,smooth},
                   xmin={0}, xmax=\xmax,
                   ymin=\ymin, ymax={1.1*gauss(\B,\B,\Bs)},
                   axis lines=middle,
                   axis line style=thick,
                   enlargelimits=upper, %
                   ticks=none,
                   x label/.style={at={(current axis.right of origin)},anchor=north west},
                   width=0.7*\textwidth, height=0.5*\textwidth,
                   clip=false, %
                   y=200pt
                  ]

        \addplot[blue, name path=B,thick] {gauss(x,\B,\Bs)};
        \addplot[red,  name path=S,thick] {gauss(x,\S,\Ss)};
        \addplot[black,dashed,thick]
          coordinates {(\B, {1.02*gauss(\B,\B,\Bs)}) (\B,{-0.05*gauss(\B,\B,\Bs)})}
          node[below=-4pt] {\strut$0$};
        \addplot[black,dashed,thick]
          coordinates {(\S, {1.02*gauss(\S,\S,\Ss)}) (\S,{-0.05*gauss(\S,\S,\Ss)})}
          node[below=-4pt] {\strut$\mu$};
        \addplot[black,thick]
          coordinates {(\q, {0.80*gauss(\B,\B,\Bs)}) (\q,{-0.05*gauss(\B,\B,\Bs)})}
          node[below=-4pt] {\strut$t_2$};

        \path[name path=xaxis]
          (0,0) -- (\pgfkeysvalueof{/pgfplots/xmax},0); %
        \addplot[white!50!blue] fill between[of=xaxis and B, soft clip={domain=\q:\xmax}];
        \addplot[white!50!red]  fill between[of=xaxis and S, soft clip={domain=0:\q}];

        \node[above=2pt,black!20!blue] at (   \B,  {gauss(\B,\B,\Bs)}) {$H_0(\ngt_2, \nf)$};
        \node[above=2pt,black!20!red]  at (1.05*\S,{gauss(\S,\S,\Ss)}) {$H_\metric(\ngt_2, \nf)$};
        \node[right,black!20!blue,scale=1.3] at ({1.05*\q},{gauss(1.*\q,\B,\Bs)}) {\strut$\alpha$};
        \node[right,black!20!blue,scale=1.3] at ({1.3*\S},{gauss(\B,\B,\Bs)}) {\strut$\alpha=5\%$};
        \node[right,black!20!red,scale=1.3] at ({0.75*\q},{gauss(0.97*\q,\S,\Ss)}) {\strut$\beta_2$};
        \node[right,black,scale=1.3] at ({1.3*\S},{gauss(\B,\B,\Bs)-0.065}) {\strut$\ngt_2>\ngt_1$};

      \end{axis}
    \end{tikzpicture}}

    \vspace*{-5mm}
    \caption{Schematic of our power analysis. It can be thought of as a classification problem, where the task is to distinguish between effects that are due to statistical noise (i.e., samples from the null hypothesis distribution $H_0$, in blue, or the negative class) and effects due to the discrepancy between ground truth and forecast as measured by the scoring rule $S$ (i.e., samples from the alternative hypothesis $H_\metric$, in red, or the positive class).
    The critical value $t_\alpha$ for the classification task is set so that the false positive rate $\alpha$ equals $0.05$. 
    The statistical power of a test corresponds then to the true positive rate $1{-}\beta$ and varies depending on three factors:
    (1) The number of trials $n$; (2) The sample size $m$; (3) The studied scoring rule $\metric$.
    For instance, by increasing the number of trials from the top ($n_1$) to the bottom ($n_2$) plots, the same scoring rule $\metric$ has better power, as the false negative rate decreases (from $\beta_1$ to $\beta_2$).
    On the same problem and for the same $n$ and $m$, different scoring rules can have different power, depending on their ability to capture the mismatch between ground truth and forecast. }
    \label{fig:power-scheme}
    \vspace*{-5mm}
\end{figure}

\subsection{Identifying Regions of Reliability}

By performing the power analysis described above for a variety of conditions: number of variables ($\dvar$), ground-truth sample size ($\ngt$), and forecast sample size ($\nf$), we can isolate the set of conditions (region) under which a scoring rule $S$ achieves a power of at least $1 - \beta$ for some $\beta$ of interest:

\begin{definition}[Region of reliability]
We define the region of reliability of level $1 - \beta$ for a scoring rule $S$, denoted by \ror{(1 - \beta)}, as the subset of $dom(d) \times dom(\ngt) \times dom(\nf)$, with $dom(\cdot)$ the domain of each factor, where $S$ achieves at least $1 - \beta$ statistical power as defined in~\cref{def:power}.
\end{definition}

Note that the above methodology cannot directly be applied to real-world datasets since their ground-truth distributions are unknown, making it infeasible to sample $\ngt$ ground-truth trajectories and to produce forecasts that are qualitatively different from them.
Hence, in what follows, we construct a synthetic benchmark that enables the controlled evaluation of scoring rules in a comprehensive set of experimental conditions where the ground-truth distribution is known, and we test the generalizability of our findings beyond these simulated cases on three temporal real datasets.

\section{Benchmark experiments}\label{sec:benchmark}

We propose a benchmark consisting of a comprehensive array of test cases, each made up of a ground-truth and forecast distributions that differ in a chosen feature (e.g., a statistical moment) by a controlled amount ($\varepsilon$).
These test cases are carefully selected to enable a systematic evaluation of scoring rules on specific discrepancies that may occur when building probabilistic forecasts for real-world tasks.

\subsection{Proper Scoring Rules}

We analyze five scoring rules that are common in the multivariate probabilistic forecasting literature: the Negative Log-Likelihood (NLL), the Continuous Ranked Probability Score (CRPS), the Energy Score (ES), the Variogram (VG), and the Dawid-Sebastiani score (DS).
For the CRPS, we use its average over the dimensions\footnote{Taking the average CRPS is often used in practice, although a multivariate extension of CRPS has been proposed \cite{gneiting2007strictly}.} and we consider two variations, which we denote by CRPS-Q (quantiles) and CRPS-E (expectations), where CRPS-Q is the most commonly used.
Similarly, for the ES, we consider the complete numerical approximation (ES-Full) and a faster approximation (ES-Partial).
Detailed definitions of these rules and their respective parameters can be found in \cref{app:scoring-rules}.

\subsection{Test Cases and Tuning} \label{sec:test_cases_and_calibration}

\begin{table}
\caption{Nomenclature of test cases where distributions vary in their marginal distributions. Each test case is defined by distributions with a specific kind of marginals, a subset of dimensions that change, a parameter to modify them (denoted $\varepsilon$ for generality), and a shift direction.
For instance, Exponential(All,$\mu \uparrow$) designates a test case where the ground-truth distribution consists of multiple independent Exponential distributions with mean 1, while the forecast distribution is the same but with mean $\mu = \varepsilon > 1$ for all dimensions.}
\smallskip\small
\begin{tabular}{p{0.55\linewidth}p{0.35\linewidth}}
\toprule
  \textbf{Marginal distribution} & \textbf{Dimension subset} \\
\midrule
  \textbf{Normal} 
  $\Norm(\mu,\sigma^2)$, the Normal distribution with mean $\mu = 0$ and standard deviation $\sigma = 1$. &
  \textbf{Single}  Only the first dimension is modified in the forecast. \\[24pt]
  \textbf{Exponential} 
  $\Exp(\lambda = 1 / \mu)$, the Exponential distribution with mean $\mu = 1$. &
  \textbf{All}  All dimensions are modified identically in the forecast. \\[24pt]\Cline{2-2}{0.8pt}
  \multirow{3}{*}{\shortstack[l]{\textbf{Skew Normal} \\ $\Skew(\mu=0, \sigma^2=1, \alpha)$, the Skew \\ Normal distribution with mean equal \\ to 0, variance equal to 1, and shape \\ parameter $\alpha$.}}
   & \rule{0pt}{2ex}\textbf{Direction} \\\cline{2-2}
  & $\uparrow$  The parameter is increased in the forecast.\\[12pt]
   & $\downarrow$  The parameter is decreased in the forecast.\\
\bottomrule
\end{tabular}
\label{tab:test_enumeration_marginals}
\vspace*{-5mm}
\end{table}

\begin{table}
\caption{Nomenclature of test cases based on multivariate Gaussian distributions, with means set to 0 and variances to 1, but varying correlations.
For each test case, we decide whether it is the ground truth or the forecast whose covariance matrix is not the identity, and which entries of said covariance matrix are set to $\varepsilon$.
For instance, Full Cov(Extra) describes a test where the ground-truth distribution is a multivariate Gaussian distribution with 0-mean and identity covariance matrix, while the forecast distribution is a multivariate Gaussian distribution of 0-mean and covariance matrix equal to 1 on its diagonal and $\varepsilon$ elsewhere.}
\smallskip\small
\begin{tabular}{P{0.51\linewidth}P{0.4\linewidth}}
\toprule
  \textbf{Covariance matrix} & \textbf{Non-trivial covariance} \\
\midrule
  \textbf{Full Cov} 
  $\Sigma_{ab} = \varepsilon \quad \forall a \neq b$. All correlations are equal to some positive constant. &
  \textbf{Missing} The ground-truth covariance matrix is not the identity. \\[24pt]
  \textbf{Checker Cov} 
  $\Sigma_{ab} = (-1)^{a+b} \varepsilon $\newline$ \forall a \neq b$. All correlations alternate between $+\varepsilon$ and $-\varepsilon$. &
  \textbf{Extra}  The Forecast distribution covariance matrix is not the identity. \\[24pt]
  \textbf{Block Cov} \newline
  $\Sigma_{2h,2h+1} = \Sigma_{2h+1,2h} = \varepsilon $\newline$ \forall h \in \{1,\ldots,\dvar/2\}$. The correlation matrix is block-diagonal with blocks of size 2-by-2. &
  \\
\bottomrule
\end{tabular}
\label{tab:test_enumeration_correlations}
\vspace*{-5mm}
\end{table}

We consider 19 test cases, or discrepancy types, categorized into (i)~distributions that differ in their marginal distributions (detailed in \cref{tab:test_enumeration_marginals}), (ii)~distributions that differ in their covariance structure (detailed in \cref{tab:test_enumeration_correlations}), and (iii)~multivariate Gaussian mixture distributions where the forecast has one more or one fewer mixture component than the ground truth, denoted by \emph{Mixture (Missing)} and \emph{Mixture (Extra)}, respectively.
An explicit definition of each test case can be found in \cref{app:perturbations}.
\paragraphtight{Tuning} In all cases, we tune the magnitude of the discrepancy between distributions ($\varepsilon$), based on the performance of the NLL scoring rule.
More precisely, we fix $\varepsilon$ such that the NLL has a statistical power of $1 - \beta = 0.8$ for a significance level of $\alpha=0.05$.
This ensures that each test case is meaningful in that the magnitude of the effect is sufficient to be detected (most of the time) by the NLL. The scoring rules are thus evaluated w.r.t.\ their ability to serve as surrogates for the NLL in forecast evaluation.
In~\cref{app:results-general-remarks} we report additional results not related to the NLL power.

\subsection{Results} \label{sec:exp_results}

\begin{figure*}[h]
    \centering
    \includegraphics[width=0.9\linewidth]{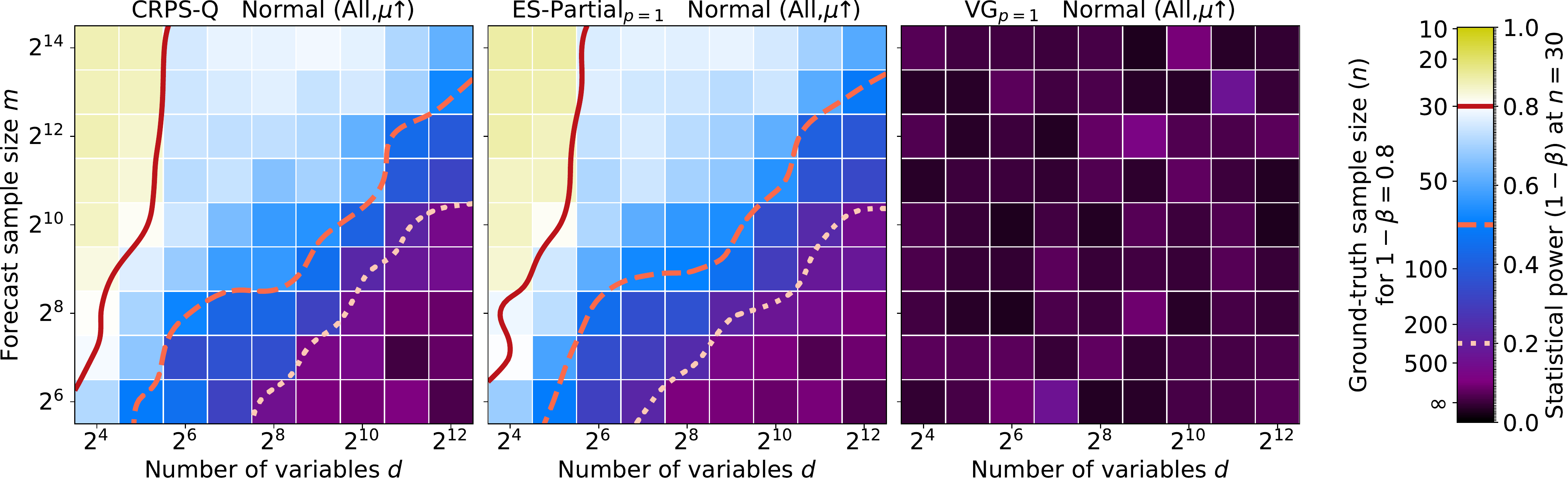}
    \caption{Statistical power of \emph{CRPS-Q} (left), \emph{ES-Partial} (middle) and \emph{VG} (right) for detecting a mismatch in mean on all dimensions, depending on the problem dimensionality ($\dvar$) and the forecast sample size ($\nf$). 
    }
    \label{fig:result_crps_quantile__wrong_mean_single}
\end{figure*}

\begin{figure*}[h]
    \centering
    \includegraphics[width=0.9\linewidth]{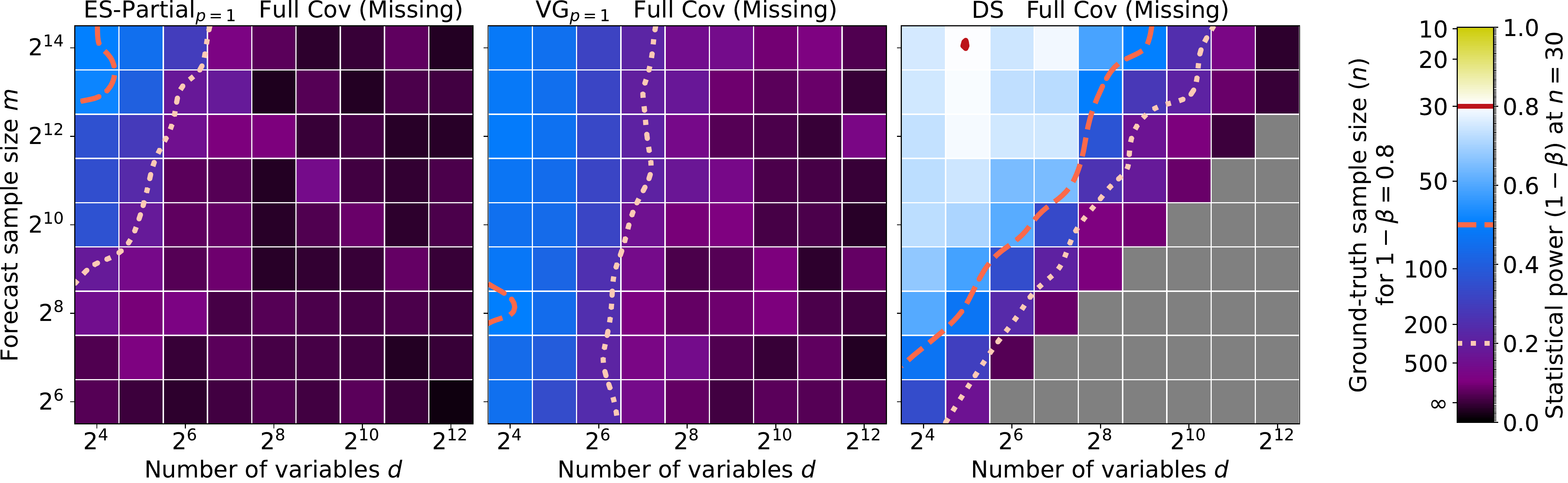}
    \caption{Statistical power of \emph{ES-Partial} (left), \emph{VG} (middle) and \emph{DS} (right) at detecting that the forecast is missing the positive correlations between variables, depending on the problem dimensionality ($\dvar$) and the forecast sample size ($\nf$). 
    \emph{DS} cannot be computed when $\dvar \ge \nf$, so the corresponding area is greyed out.
    }
    \label{fig:energy_1__missing_covariance_full}
\end{figure*}

\begin{figure*}[h]
    \centering
    \includegraphics[width=0.9\linewidth]{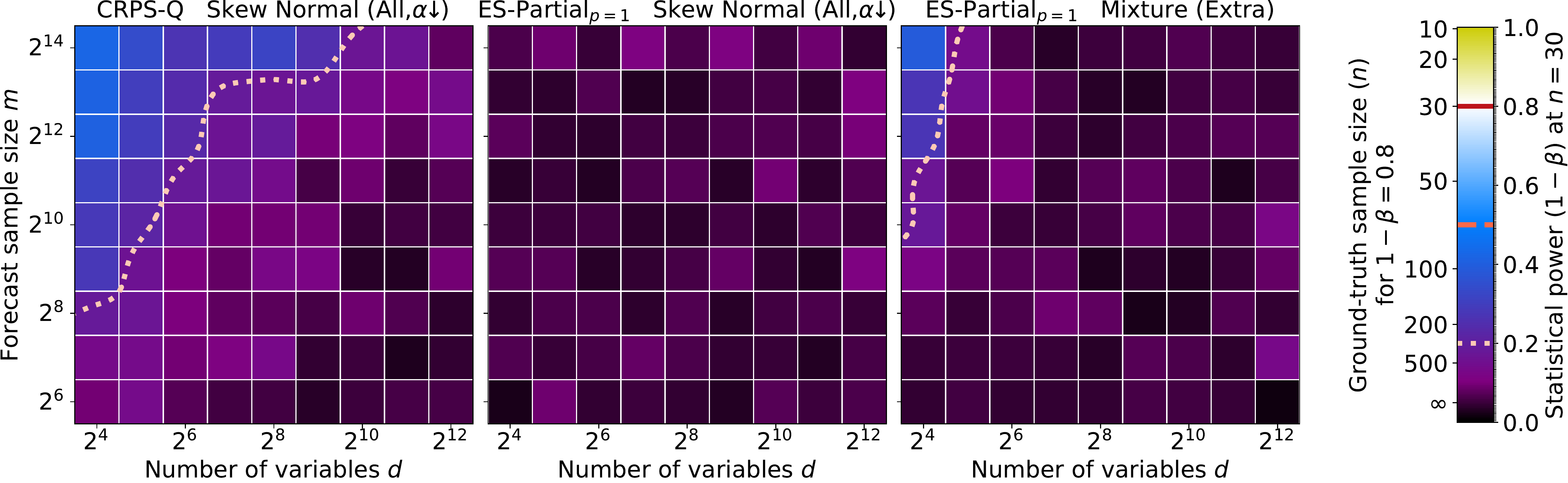}
    \caption{Statistical power of \emph{CRPS-Q} (left) and \emph{ES-Partial} (middle) at detecting that the forecast is not capturing the skewness of the ground-truth distribution, and of \emph{ES-Partial} (right) at detecting that the forecast is predicting an additional mode, all depending on the problem dimensionality ($\dvar$) and the forecast sample size ($\nf$).
    }
    \label{fig:result_crps_quantile__missing_skew_all}
\end{figure*}

\begin{figure}[h]
    \centering
    \includegraphics[width=0.95\linewidth]{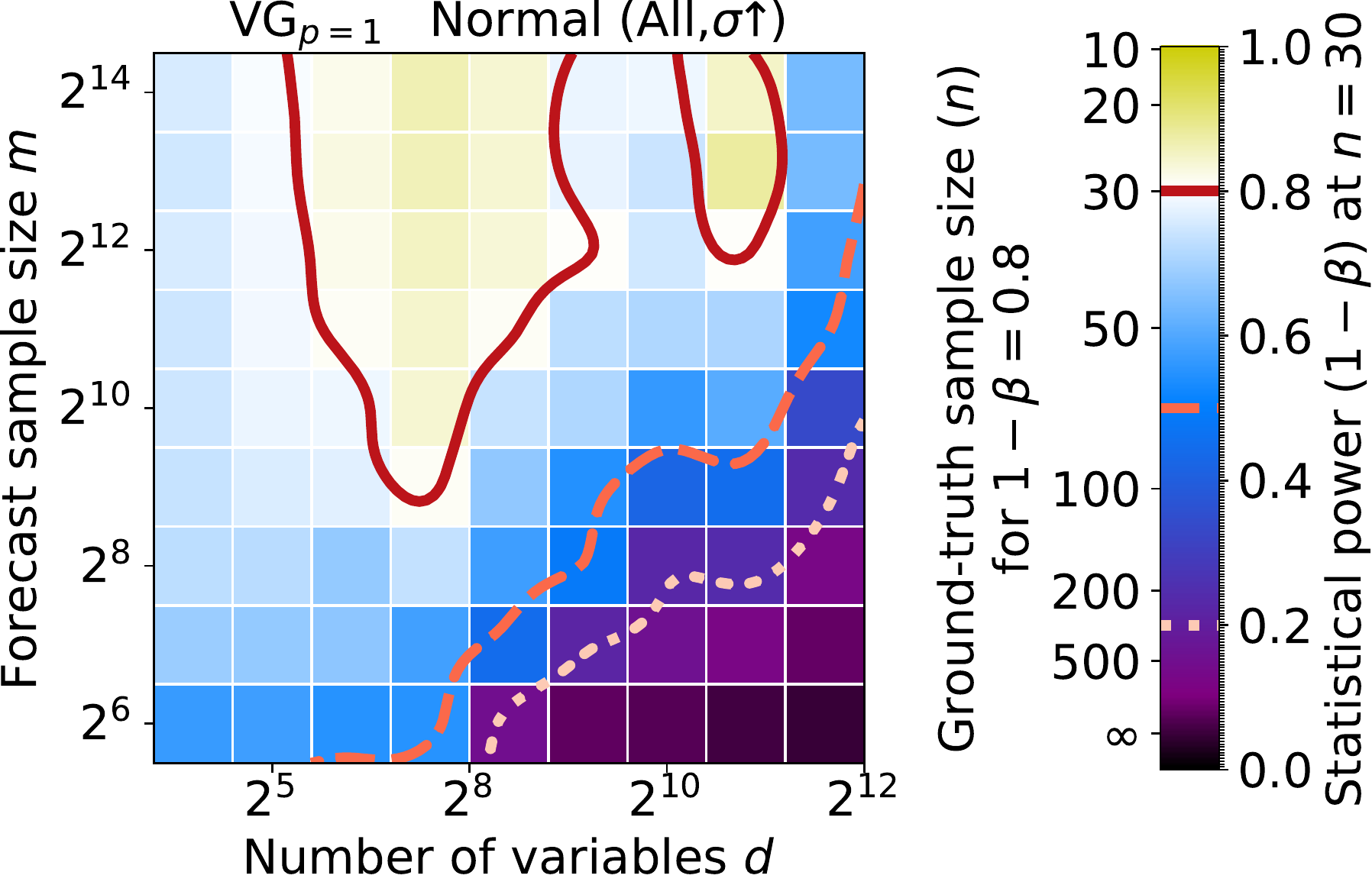}
    \caption{Non-Monotonicity of statistical power of \emph{VG} at correctly detecting that a forecast marginal variance is lower than the ground-truth one on a single dimension.}
    \label{fig:result_variogram_non_monotonic}
    \vspace*{-4mm}
\end{figure}

We start by emphasizing results of key interest by illustrating regions of reliability for specific scoring rules and test cases in \cref{fig:result_crps_quantile__wrong_mean_single,fig:energy_1__missing_covariance_full,fig:result_crps_quantile__missing_skew_all,fig:result_variogram_non_monotonic}. %
We then report a summary of all results in \cref{fig:best_mean_power_subset} and detailed results in \cref{app:results}.

\paragraphtight{Visualizing RoRs} As regions of reliability are defined over three axes (number of variables $\dvar$, ground-truth sample size $\ngt$, and forecast sample size $\nf$), we plot the cross section corresponding to setting $\ngt=30$ and show how the statistical power of a scoring rule varies with $\dvar$ and $\nf$ using a heatmap.
We choose $\ngt=30$ since (i)~it corresponds to a rolling window evaluation setting of realistic length, and (ii)~the central limit theorem commonly holds around this sample size.
Nonetheless, we note that insights on reliability for larger $\ngt$ can be inferred from such visualizations since the values of the heatmap can be interchangeably read as the statistical power at $\ngt=30$ (higher is better) or as the minimal $\ngt$ required to achieve 80\% power (lower is better).
For ease of readability, we also plot the contour lines for \ror{0.8} (solid), \ror{0.5} (dashed), and \ror{0.2} (dotted) computed based on a kernel-smoothed estimation of the measured statistical power.

\paragraphtight{Detection of Incorrect Mean}
\cref{fig:result_crps_quantile__wrong_mean_single} shows the statistical power of CRPS-Q, ES-Partial, and VG when the ground truth and the forecast means differ in a single dimension.
When $\nf > \dvar$, the CRPS-Q and ES-Partial are able to detect such discrepancies:
their \ror{0.8}, which indicates a power greater or equal to that of the NLL, spans $\dvar \le 2^5 \;\cap\; \nf \ge 2^{10}$.
However, when $\nf < \dvar$, a setting that is ubiquitous in real-world benchmark datasets~\cite{godahewa2021monash}, their power drops below 20\% and we find that $\ngt = 300$ ground-truth samples (e.g., rolling windows) would be needed for these scoring rules to perform on par with the NLL.
In contrast, VG never succeeds at detecting the discrepancy, with empty \ror{0.8} and \ror{0.5}, and a power below 20\% even when $m \gg d$.
We obtain similar results on the test cases with incorrect means for all dimensions (cf.~\Cref{app:results}).

\paragraphtight{Detection of Missing Correlations}
\cref{fig:energy_1__missing_covariance_full} shows the statistical power of ES-Partial, DS, and VG when the ground truth and forecast differ only in their correlation structure.
DS is the only scoring rule with a non-empty, albeit very small, \ror{0.8} ($\dvar = 2^5,\; \nf=2^{14}$).
However, this scoring rule makes parametric assumptions\footnote{The Dawid-Sebastiani scoring rule approximates the multivariate Gaussian NLL (see \cref{app:dawid-sebastiani}), which matches the data.}
that match the underlying distributions, giving it a significant advantage.
As for ES-Partial, its performance is far from that of the NLL, with a small \ror{0.5} ($\dvar = 2^4 \;\cap\; \nf \geq 2^{13}$) and a larger \ror{0.2} ($\dvar \leq 2^6 \;\cap\; \nf \geq 2^9$).
VG performs comparably but, interestingly, it seems less affected by the forecast sample size ($m$), with a \ror{0.2} spanning all values of $m$ ($\dvar \leq 2^7$).
Nevertheless, we find that the ground-truth sample size ($\ngt$) needed for ES-Partial and VG to perform on par with the NLL is of the order of hundreds or thousands, a setting that is highly impractical.

\paragraphtight{Detection of Incorrect Higher Moments}
\cref{fig:result_crps_quantile__missing_skew_all} shows the power of CRPS-Q and ES-Partial when the distributions differ in higher moments, i.e., their mean and covariance are identical.
We illustrate two cases: (i)~differences in skewness (\cref{fig:result_crps_quantile__missing_skew_all}, left-middle) and (ii)~the inclusion of an extra distribution mode in the forecast (\cref{fig:result_crps_quantile__missing_skew_all}, right).
In all cases, the scoring rules completely underperform in comparison to the NLL.
For differences in skewness, CRPS-Q is the only scoring rule that shows any statistical power, reaching a power of around 20\% for very high sample sizes ($\nf \ge 2^{13}$) or very low numbers of dimensions ($\dvar \le 2^5$).
As for ES-Partial, it reaches power values close to the false positive rate ($\alpha = 5\%$), indicating that its scores are essentially random.
Finally, for the case where the forecast contains an extra mode, ES-Partial is the only scoring rule that was found to achieve non-negligible power.
The results indicate a small \ror{0.2} spanning $d \le 2^5 \;\cap\; n \ge 2^{10}$ and that an impractical $n$ of the order of hundreds would be required to match NLL performance.

\paragraphtight{Extrapolating these Results}
Can these results be extrapolated beyond the ranges considered for $\dvar$, $\nf$, $\ngt$?
Statistical power generally increases with $\nf$ and decreases with $\dvar$ and this is clearly reflected in our analysis.
However, this does not necessarily happen monotonically, as illustrated in \cref{fig:result_variogram_non_monotonic}.
We provide an explanation for this in \cref{app:results-general-remarks}.

\paragraphtight{Summary of Results} 
Finally, we summarize the results of our analysis in \cref{fig:best_mean_power_subset} by listing, for each scoring rule and test case, the maximal power (over $m$) averaged over all values of $d$.
This measure gives insights into the performance of a scoring rule in our benchmark, regardless of the dimensionality.
For instance, a value of $0.8$ indicates that a scoring rule typically performs on par with the NLL, whereas values $< 0.8$ indicate that it underperforms.
Here are a few notable observations:
\begin{itemize}

\item Most scoring rules are only capable of detecting errors in the marginal distributions, as revealed by their small power in the test cases that induce errors in covariance.

\item For covariance-related test cases, DS shows the greatest power, followed by VG. Yet, they both significantly underperform the NLL. DS's stronger performance is likely due to its Gaussian parametric assumptions (see \cref{app:dawid-sebastiani}), a thesis supported by its failure in the mixture test cases.

\item VG is complementary to the CRPS and ES in our settings. When VG achieves lower power, CRPS/ES generally achieves higher power and vice versa. The only exceptions are the mixture test cases, where all scoring rules fail.

\item The various implementations of CRPS and ES perform comparably. This indicates that the computational gain of CRPS-Q and ES-Partial does not negatively affect their reliability.

\end{itemize}

\paragraphtight{Generalizability to Real Data}
The regions of reliability that we identify in the simulated benchmark provide valuable insights for practitioners. In fact, even if the distributions used in the benchmark do not match their data, observing a low-power result in one setting is indicative of a potential problem.
This means that there exists at least one simple setting where the scoring rule fails to detect an error, indicating that one should take additional measures to ensure the validity of one's findings (see \cref{sec:conclusion}).
As such, the regions of reliability calculated on simulated data serve as a necessary sanity check, i.e., the scoring rules that do not pass the tests proposed herein for a given setting should not be relied upon in practical settings.
In the next section, we evaluate whether our findings also hold for real-world temporal data.

\section{Application to Real Temporal Data}\label{sec:real-data}

We now explore the generalizability of our findings to more realistic temporal data distributions in the context of our motivating example (see \cref{sec:background}).
Since our methodology can only be applied when the ground-truth distribution is known, we fit the TACTiS multivariate probabilistic forecasting model \cite{pmlr-v162-drouin22a} to the \texttt{solar-10min} dataset ($d=9864$, $n=30$, $m=100$; details in \cref{app:real-data}) and consider the learned distribution to be the ground truth.
We then produce forecast distributions with three kinds of errors, which are likely to affect the resulting profit, \cref{eq:motivating}, namely: (i)~breaking all correlations between the variables, which prevents one from strategically selecting power stations to put into maintenance (e.g., based on location), (ii)~multiplying all variables by $1.05$, which causes an overestimation of power production, resulting in penalties, and (iii)~adding $0.05$ to all variables, which has a similar effect on profit.\footnote{For scale comparison, the average mean value is equal to $3.04$ in the forecast used for this study.}
The TACTiS model was chosen because of the ease of removing all dependencies between variables when computing its NLL, as it is a copula-based model.

Given that this is a real-world setting, we cannot tune the difficulty of the task consisting of detecting broken correlations.
For the multiplicative and additive errors, we picked their parameters such that their respective decreases in profit are in the same order of magnitude as the one for broken correlations.
These profit decreases are computed with the model presented in \cref{eq:motivating}, with $M=50$, and $\lambda=10$.
This results in test cases that are somewhat easier than those studied in the benchmark, as reflected by the NLL achieving a power of $1.0$ (instead of $0.8$) in each case.
Nonetheless, all the other scoring rules fail in at least one case:
\begin{itemize}
    \item \textbf{Breaking Correlations:} Profit decrease: 2.3\%. As expected, the CRPS fails to capture this, achieving a power of $0.05$ (see~\Cref{fig:full_page_missing_covariance_full}). ES and VG achieve a perfect power of $1.0$, which is surprising given that $d \gg m$, but may be explained by the simplicity of the problem (see~\cref{fig:calib-corr} for a study of statistical power as a function of the difficulty of the problem). 
    \item \textbf{Multiplying by a Constant:} Profit decrease: 3.0\%. The CRPS, ES, and VG all detect the discrepancy, with powers of $0.85$, $0.75$ and $0.75$, respectively. This result is not surprising given that these rules each showed significant power at detecting increases in mean in exponential distributions in \cref{fig:full_page_wrong_exponential_all_higher}, which is the closest analog in our benchmark (more details in~\cref{app:real-data}).
    \item \textbf{Adding a Constant:} Profit decrease: 5.1\%. The CRPS succeeds at detecting the error, with a power of $1.0$. However, the ES performs poorly, with a power of $0.19$ and VG completely fails, with a power of $0.05$. The results for the CRPS and VG are in line with those in \cref{fig:full_page_wrong_mean_all} for the test case where the marginal means were modified. As for the poor performance of ES, it may be explained by a large $d$, which places this rule outside of a region of reliability in our benchmark results.
\end{itemize}

While limited in scale, this experiment suggests that our benchmark results are informative, as most observations transfer to this more realistic setting.
In \cref{app:more_real_data}, we expand the analysis to two additional real-world datasets and arrive at similar conclusions.

\section{Discussion}\label{sec:conclusion}
This work studied the reliability of common proper scoring rules at detecting a variety of errors of practical significance in multivariate probabilistic time-series forecasting.
For this purpose, we introduced the notion of regions of reliability, i.e., the experimental conditions under which a proper scoring rule can reliably be used for evaluation, and devised a methodology to identify them.
Our proposed methodology consists in assessing the statistical power of a scoring rule (\cref{sec:ror}) on a comprehensive benchmark (\cref{sec:benchmark}) of test cases that were carefully selected to be of practical relevance.
All test cases were tuned with respect to the Negative Log-Likelihood in order to assess whether existing scoring rules could serve as reliable substitutes for it.
Our results paint a clear picture: although they are theoretically grounded in the asymptotic regime, none of the considered proper scoring rules is a good surrogate for the NLL, across all test cases, in the small-sample regimes that are common in practice.
What is more, none of these scoring rules is able to reliably detect errors in modeling the statistical dependencies across variables, which is a fundamental goal of \emph{multivariate} forecasting.

\begin{figure}
    \vspace*{-2mm}
    \centering
    \includegraphics[width=\linewidth]{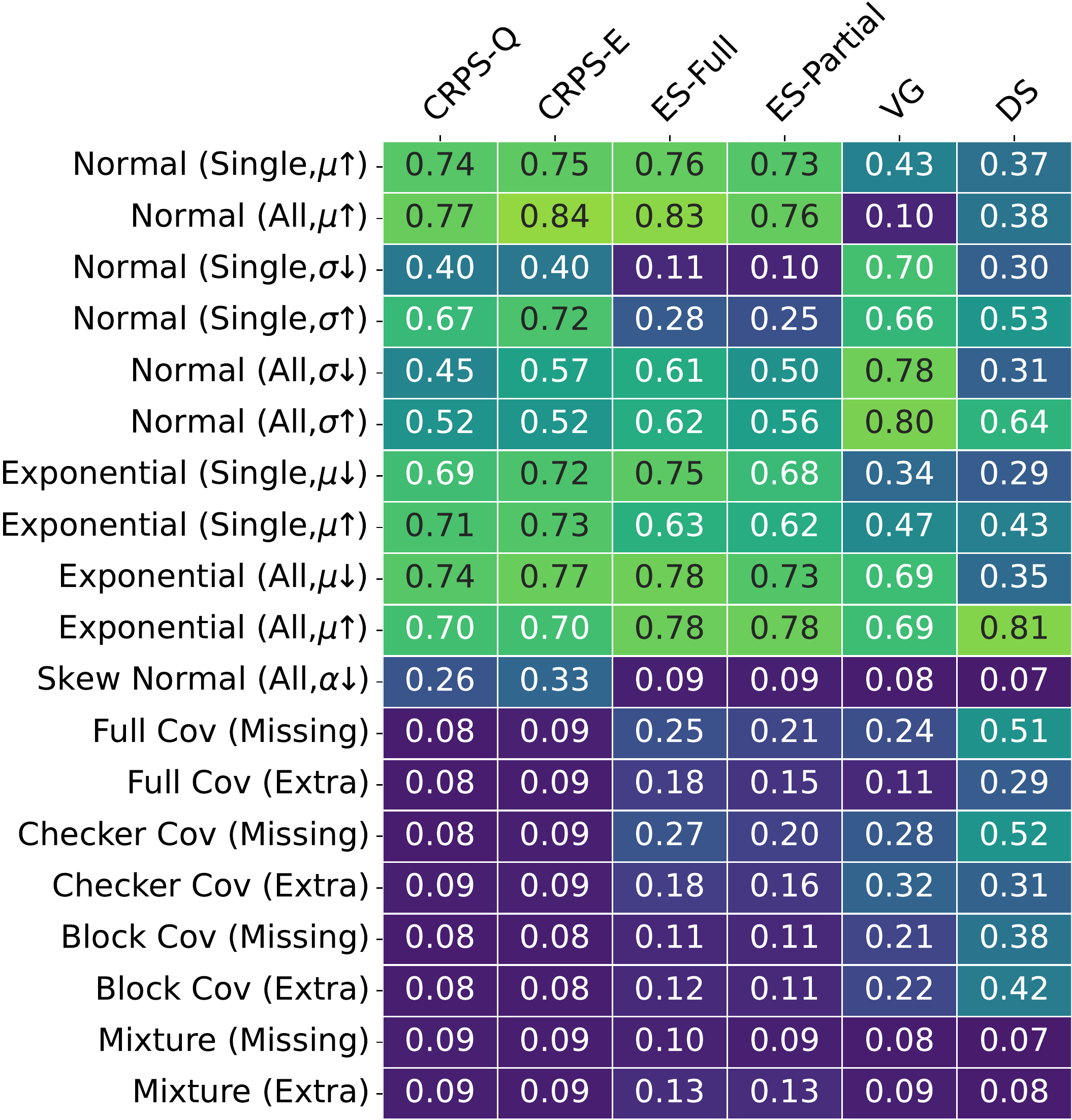}
    \caption{Summary of results: maximal statistical power ($1 - \beta$) over the sample size $m$, averaged over the number of variables $d$.}
    \label{fig:best_mean_power_subset}
\end{figure}

What are the implications of these findings for the time-series community?
A review of the recent literature (e.g., \citet{salinas2019high, rasul2021multivariate,NEURIPS2020_1f47cef5,pmlr-v139-rasul21a,NEURIPS2021_cfe8504b,nguyen2021temporal,NEURIPS2021_c68bd905,pmlr-v162-drouin22a}) reveals that 
most contributions are benchmarked by performing rolling-window evaluation on the following datasets: \texttt{electricity} ($d = 8880$, $n = 7$), \texttt{exchange} ($d = 240$, $n = 5$), \texttt{solar} ($d = 3288$, $n = 7$), \texttt{taxi} ($d = 29136$, $n = 57$), \texttt{traffic} ($d = 23112$, $n = 7$), \texttt{wikipedia} ($d = 60000$, $n = 5$), with $m \in [100, 1000]$ depending on the experimental choices of each study.
Notice that, in all cases, the sample size ($m$) and the number of evaluation windows ($n$) are small in comparison to the dimensionality of the data ($d$).
Our results suggest that, in these regimes, the assessed scoring rules are generally unreliable, i.e., they detect fewer or none of the errors that the NLL would be able to capture.
This observation is alarming, as it puts into question the reliability of how progress is currently measured in the field.

\paragraphtight{Practical Recommendations} We emphasize the need to evaluate models in settings where the number of ground truth samples ($n$) and the number of forecast samples ($m$) are significantly larger than current standard practice in the literature.
As described in \cref{sec:power}, both of these quantities have a direct effect on the statistical power of scoring rules.
On the one hand, increasing $n$ always improves the statistical power (see \cref{eq:delta_variance}), but is not always a practical solution as it requires collecting more data or reducing the training set in favor of the test set.
Nonetheless, the large size of the aforementioned datasets should be amenable to such a change.
For instance, the \texttt{electricity} dataset contains observations at $5790$ time points, a meager $7$ of which are typically used for evaluation.
On the other hand, increasing $m$ is only guaranteed to increase power up to an asymptote that is not necessarily $\beta = 0$, but it is a viable option that comes at a reasonable computational cost.
Therefore, we recommend taking action on both quantities.
Furthermore, when planning an experiment on real-world data, we recommend that practitioners query our results, with the appropriate values for $d$, $m$, and $n$, and look at the power for the kinds of errors that would be detrimental to their downstream task.
If a scoring rule has low power in our simulated benchmark, they should proceed with caution, as there exists at least one data distribution where the rule is unreliable.
Finally, we align with the common recommendation of complementing evaluation with other diagnostic tools, such as inspecting calibration and sharpness~\citep{gneiting2007probabilistic} and plotting correlations~\citep{pmlr-v162-drouin22a}.

\paragraphtight{Limitations and Future Work}
Our results are of empirical nature. As such, we cannot guarantee their validity beyond the considered domains (especially for $d$ and $m$), which are constrained for computational reasons.
This is particularly true in light of the non-monotonic behavior of the statistical power on certain test cases, as reported for instance in~\Cref{fig:result_variogram_non_monotonic}.
Deriving finite-sample theoretical guarantees would be key to characterizing the domains not studied in the current work and is a promising direction for future work.
Overall, our analysis highlights the need for developing new scoring rules, with better finite-sample behavior.
By providing a simulated benchmark of diverse test cases with known statistical properties, our work enables researchers to assess and compare the performance of different scoring rules in a concrete and standardized way.
Our hope is that our work will stimulate the development of new scoring rules, and more extensive benchmarks, to better capture different aspects of prediction quality, and ultimately enable objective assessment of progress in the field.\footnote{Benchmark data and supporting code are available at \url{https://github.com/ServiceNow/regions-of-reliability}.}
As some scoring rules appear complementary (e.g., CRPS + VG, see \cref{sec:benchmark}), a promising direction for future work would be to use our benchmark to learn combinations of scoring rules that achieve high power across a variety of settings.

\paragraphtight{Broader Impact}
Our research examines the reliability of scoring rules currently used in the evaluation of multivariate time series forecasting models. Our findings highlight the need for the scientific community to reassess how progress is measured in this field. By raising awareness of this issue, our research should encourage scientists to supplement their model evaluation with additional measures of correctness. This will stimulate the development of better evaluation protocols, leading to more accurate detection and tracking of innovation in the field. Ultimately, our research should lead to better screening of the shortcomings of forecasting models used in practical applications of machine learning, which have become increasingly ubiquitous in our daily lives. While there is a potential for this work to delay the deployment of innovative solutions due to concerns over forecast accuracy, we believe that the positive benefits to science will ultimately outweigh potential short-term negatives.

\bibliography{multivariate_metrics.bib,forecast_performance.bib}

\begin{thebibliography}{40}
\providecommand{\natexlab}[1]{#1}
\providecommand{\url}[1]{\texttt{#1}}
\expandafter\ifx\csname urlstyle\endcsname\relax
  \providecommand{\doi}[1]{doi: #1}\else
  \providecommand{\doi}{doi: \begingroup \urlstyle{rm}\Url}\fi

\bibitem[Alexander et~al.(2022)Alexander, Coulon, Han, and Meng]{Alexander2022}
Alexander, C., Coulon, M., Han, Y., and Meng, X.
\newblock Evaluating the discrimination ability of proper multi-variate scoring
  rules.
\newblock \emph{Annals of Operations Research}, 2022.
\newblock \doi{10.1007/s10479-022-04611-9}.

\bibitem[Bao et~al.(2007)Bao, Lee, and Salto\u{g}lu]{Bao2007}
Bao, Y., Lee, T.-H., and Salto\u{g}lu, B.
\newblock Comparing density forecast models.
\newblock \emph{Journal of Forecasting}, 26\penalty0 (3):\penalty0 203–225,
  2007.

\bibitem[Cai(2002)]{cai2002quantiles}
Cai, Z.
\newblock Regression quantiles for time series.
\newblock \emph{Econometric Theory}, 18\penalty0 (1):\penalty0 169--192, 2002.
\newblock ISSN 02664666, 14694360.
\newblock URL \url{http://www.jstor.org/stable/3533031}.

\bibitem[Chatfield(1993)]{Chatfield:1993tm}
Chatfield, C.
\newblock Calculating interval forecasts.
\newblock \emph{Journal of Business \& Economic Statistics}, 11\penalty0
  (2):\penalty0 121–135, 1993.

\bibitem[Chatfield(2001)]{Chatfield2001}
Chatfield, C.
\newblock Prediction intervals for time-series forecasting.
\newblock In Armstrong, J.~S. (ed.), \emph{Principles of Forecasting: A
  Handbook for Researchers and Practitioners}, pp.\  475--494, Boston, MA,
  2001. Springer US.
\newblock \doi{10.1007/978-0-306-47630-3_21}.

\bibitem[Cohen(1992)]{cohen1992statistical}
Cohen, J.
\newblock Statistical power analysis.
\newblock \emph{Current directions in psychological science}, 1992.

\bibitem[de~B\'{e}zenac et~al.(2020)de~B\'{e}zenac, Rangapuram, Benidis,
  Bohlke-Schneider, Kurle, Stella, Hasson, Gallinari, and
  Januschowski]{NEURIPS2020_1f47cef5}
de~B\'{e}zenac, E., Rangapuram, S.~S., Benidis, K., Bohlke-Schneider, M.,
  Kurle, R., Stella, L., Hasson, H., Gallinari, P., and Januschowski, T.
\newblock Normalizing {Kalman} filters for multivariate time series analysis.
\newblock In Larochelle, H., Ranzato, M., Hadsell, R., Balcan, M., and Lin, H.
  (eds.), \emph{Advances in Neural Information Processing Systems}, volume~33,
  pp.\  2995--3007. Curran Associates, Inc., 2020.
\newblock URL
  \url{https://proceedings.neurips.cc/paper/2020/file/1f47cef5e38c952f94c5d61726027439-Paper.pdf}.

\bibitem[Diebold et~al.(1998)Diebold, Gunther, and Tay]{Diebold1998}
Diebold, F.~X., Gunther, T.~A., and Tay, A.~S.
\newblock Evaluating density forecasts with applications to financial risk
  management.
\newblock \emph{International Economic Review}, 39\penalty0 (4):\penalty0 863,
  1998.

\bibitem[Drouin et~al.(2022)Drouin, Marcotte, and
  Chapados]{pmlr-v162-drouin22a}
Drouin, A., Marcotte, E., and Chapados, N.
\newblock {TACT}i{S}: Transformer-attentional copulas for time series.
\newblock In Chaudhuri, K., Jegelka, S., Song, L., Szepesvari, C., Niu, G., and
  Sabato, S. (eds.), \emph{Proceedings of the 39th International Conference on
  Machine Learning}, volume 162 of \emph{Proceedings of Machine Learning
  Research}, pp.\  5447--5493. PMLR, 17--23 Jul 2022.
\newblock URL \url{https://proceedings.mlr.press/v162/drouin22a.html}.

\bibitem[Gneiting \& Katzfuss(2014)Gneiting and
  Katzfuss]{gneiting2014probabilistic}
Gneiting, T. and Katzfuss, M.
\newblock Probabilistic forecasting.
\newblock \emph{Annual Review of Statistics and Its Application}, 1:\penalty0
  125--151, 2014.

\bibitem[Gneiting \& Raftery(2007)Gneiting and Raftery]{gneiting2007strictly}
Gneiting, T. and Raftery, A.~E.
\newblock Strictly proper scoring rules, prediction, and estimation.
\newblock \emph{Journal of the American statistical Association}, 102\penalty0
  (477):\penalty0 359--378, 2007.

\bibitem[Gneiting \& Ranjan(2011)Gneiting and Ranjan]{Gneiting2011}
Gneiting, T. and Ranjan, R.
\newblock Comparing density forecasts using threshold- and quantile-weighted
  scoring rules.
\newblock \emph{Journal of Business \& Economic Statistics}, 29\penalty0
  (3):\penalty0 411–422, 2011.

\bibitem[Gneiting et~al.(2007)Gneiting, Balabdaoui, and
  Raftery]{gneiting2007probabilistic}
Gneiting, T., Balabdaoui, F., and Raftery, A.~E.
\newblock Probabilistic forecasts, calibration and sharpness.
\newblock \emph{Journal of the Royal Statistical Society: Series B (Statistical
  Methodology)}, 69\penalty0 (2):\penalty0 243--268, 2007.

\bibitem[Gneiting et~al.(2008)Gneiting, Stanberry, Grimit, Held, and
  Johnson]{gneiting2008assessing}
Gneiting, T., Stanberry, L.~I., Grimit, E.~P., Held, L., and Johnson, N.~A.
\newblock Assessing probabilistic forecasts of multivariate quantities, with an
  application to ensemble predictions of surface winds.
\newblock \emph{Test}, 17:\penalty0 211--235, 2008.

\bibitem[Godahewa et~al.(2021)Godahewa, Bergmeir, Webb, Hyndman, and
  Montero-Manso]{godahewa2021monash}
Godahewa, R., Bergmeir, C., Webb, G.~I., Hyndman, R.~J., and Montero-Manso, P.
\newblock Monash time series forecasting archive.
\newblock In \emph{Neural Information Processing Systems Track on Datasets and
  Benchmarks}, 2021.
\newblock forthcoming.

\bibitem[Hamill(2001)]{hamill2001interpretation}
Hamill, T.~M.
\newblock Interpretation of rank histograms for verifying ensemble forecasts.
\newblock \emph{Monthly Weather Review}, 2001.

\bibitem[Heizer et~al.(2023)Heizer, Render, and Munson]{heizer2023operations}
Heizer, J., Render, B., and Munson, C.
\newblock \emph{Operations Management: Sustainability and Supply Chain
  Management}.
\newblock Pearson, 14th edition, 2023.

\bibitem[Hewamalage et~al.(2021)Hewamalage, Montero-Manso, Bergmeir, and
  Hyndman]{Hewamalage2021look}
Hewamalage, H., Montero-Manso, P., Bergmeir, C., and Hyndman, R.~J.
\newblock A look at the evaluation setup of the {M5} forecasting competition,
  2021.
\newblock URL \url{https://arxiv.org/abs/2108.03588}.

\bibitem[Hyndman \& Koehler(2006)Hyndman and Koehler]{Hyndman:2006dt}
Hyndman, R.~J. and Koehler, A.~B.
\newblock Another look at measures of forecast accuracy.
\newblock \emph{International Journal of forecasting}, 22\penalty0
  (4):\penalty0 679–688, 2006.

\bibitem[Koochali et~al.(2022)Koochali, Schichtel, Dengel, and
  Ahmed]{Koochali2022}
Koochali, A., Schichtel, P., Dengel, A., and Ahmed, S.
\newblock Random noise vs. state-of-the-art probabilistic forecasting methods:
  A case study on {CRPS-Sum} discrimination ability.
\newblock \emph{Applied Sciences}, 12\penalty0 (10):\penalty0 5104, 2022.

\bibitem[Mahmoud(1984)]{Mahmoud:1984uy}
Mahmoud, E.
\newblock Accuracy in forecasting: A survey.
\newblock \emph{Journal of Forecasting}, 3\penalty0 (2):\penalty0 139–159,
  1984.

\bibitem[Makridakis \& Hibon(1979)Makridakis and Hibon]{Makridakis:1979tt}
Makridakis, S. and Hibon, M.
\newblock Accuracy of forecasting: An empirical investigation.
\newblock \emph{Journal of the Royal Statistical Society. Series A (General)},
  142\penalty0 (2):\penalty0 97–145, 1979.

\bibitem[Makridakis \& Hibon(2000)Makridakis and Hibon]{Makridakis:2000tj}
Makridakis, S. and Hibon, M.
\newblock The {M3}-competition: results, conclusions and implications.
\newblock \emph{International Journal of forecasting}, 16\penalty0
  (4):\penalty0 451–476, 2000.

\bibitem[Makridakis et~al.(1982)Makridakis, Andersen, Carbone, Fildes, Hibon,
  Lewandowski, Newton, Parzen, and Winkler]{Makridakis:1982we}
Makridakis, S., Andersen, A., Carbone, R., Fildes, R., Hibon, M., Lewandowski,
  R., Newton, J., Parzen, E., and Winkler, R.
\newblock The accuracy of extrapolation (time series) methods: Results of a
  forecasting competition.
\newblock \emph{Journal of Forecasting}, 1\penalty0 (2):\penalty0 111–153,
  1982.

\bibitem[Makridakis et~al.(2018)Makridakis, Spiliotis, and
  Assimakopoulos]{Makridakis:2018em}
Makridakis, S., Spiliotis, E., and Assimakopoulos, V.
\newblock The {M4} competition: Results, findings, conclusion and way forward.
\newblock \emph{International Journal of forecasting}, 34\penalty0
  (4):\penalty0 802–808, 2018.

\bibitem[Makridakis et~al.(2022)Makridakis, Spiliotis, Assimakopoulos, Chen,
  Gaba, Tsetlin, and Winkler]{Makridakis2022:uncertainty}
Makridakis, S., Spiliotis, E., Assimakopoulos, V., Chen, Z., Gaba, A., Tsetlin,
  I., and Winkler, R.~L.
\newblock The {M5} uncertainty competition: Results, findings and conclusions.
\newblock \emph{International Journal of Forecasting}, 38\penalty0
  (4):\penalty0 1365--1385, 2022.
\newblock ISSN 0169-2070.
\newblock \doi{https://doi.org/10.1016/j.ijforecast.2021.10.009}.
\newblock URL
  \url{https://www.sciencedirect.com/science/article/pii/S0169207021001722}.
\newblock Special Issue: {M5} competition.

\bibitem[Matheson \& Winkler(1976)Matheson and Winkler]{matheson1976scoring}
Matheson, J.~E. and Winkler, R.~L.
\newblock Scoring rules for continuous probability distributions.
\newblock \emph{Management science}, 22\penalty0 (10):\penalty0 1087--1096,
  1976.

\bibitem[Neyman \& Pearson(1933)Neyman and Pearson]{neyman1933ix}
Neyman, J. and Pearson, E.~S.
\newblock On the problem of the most efficient tests of statistical hypotheses.
\newblock \emph{Philosophical Transactions of the Royal Society of London.
  Series A, Containing Papers of a Mathematical or Physical Character},
  231\penalty0 (694-706):\penalty0 289--337, 1933.

\bibitem[Nguyen \& Quanz(2021)Nguyen and Quanz]{nguyen2021temporal}
Nguyen, N. and Quanz, B.
\newblock Temporal latent auto-encoder: A method for probabilistic multivariate
  time series forecasting.
\newblock In \emph{Proceedings of the AAAI Conference on Artificial
  Intelligence}, volume~35, pp.\  9117--9125, 2021.

\bibitem[Peterson(2017)]{peterson2017introduction}
Peterson, M.
\newblock \emph{An introduction to decision theory}.
\newblock Cambridge University Press, 2017.

\bibitem[Pinson \& Tastu(2013)Pinson and Tastu]{Pinson2013}
Pinson, P. and Tastu, J.
\newblock Discrimination ability of the energy score.
\newblock Technical report, Technical University of Denmark, DTU Compute
  Technical Report-2013 No. 15, 2013.

\bibitem[Rasul et~al.(2021{\natexlab{a}})Rasul, Seward, Schuster, and
  Vollgraf]{pmlr-v139-rasul21a}
Rasul, K., Seward, C., Schuster, I., and Vollgraf, R.
\newblock Autoregressive denoising diffusion models for multivariate
  probabilistic time series forecasting.
\newblock In Meila, M. and Zhang, T. (eds.), \emph{Proceedings of the 38th
  International Conference on Machine Learning}, volume 139 of
  \emph{Proceedings of Machine Learning Research}, pp.\  8857--8868. PMLR,
  18--24 Jul 2021{\natexlab{a}}.
\newblock URL \url{https://proceedings.mlr.press/v139/rasul21a.html}.

\bibitem[Rasul et~al.(2021{\natexlab{b}})Rasul, Sheikh, Schuster, Bergmann, and
  Vollgraf]{rasul2021multivariate}
Rasul, K., Sheikh, A.-S., Schuster, I., Bergmann, U.~M., and Vollgraf, R.
\newblock Multivariate probabilistic time series forecasting via conditioned
  normalizing flows.
\newblock In \emph{International Conference on Learning Representations},
  2021{\natexlab{b}}.
\newblock URL \url{https://openreview.net/forum?id=WiGQBFuVRv}.

\bibitem[Salinas et~al.(2019)Salinas, Bohlke-Schneider, Callot, Medico, and
  Gasthaus]{salinas2019high}
Salinas, D., Bohlke-Schneider, M., Callot, L., Medico, R., and Gasthaus, J.
\newblock High-dimensional multivariate forecasting with low-rank {Gaussian}
  copula processes.
\newblock \emph{Advances in neural information processing systems}, 32, 2019.

\bibitem[Scheuerer \& Hamill(2015)Scheuerer and Hamill]{Scheuerer2015}
Scheuerer, M. and Hamill, T.~M.
\newblock Variogram-based proper scoring rules for probabilistic forecasts of
  multivariate quantities.
\newblock \emph{Monthly Weather Review}, 143\penalty0 (4):\penalty0
  1321–1334, 2015.

\bibitem[Tang \& Matteson(2021)Tang and Matteson]{NEURIPS2021_c68bd905}
Tang, B. and Matteson, D.~S.
\newblock Probabilistic transformer for time series analysis.
\newblock In Ranzato, M., Beygelzimer, A., Dauphin, Y., Liang, P., and Vaughan,
  J.~W. (eds.), \emph{Advances in Neural Information Processing Systems},
  volume~34, pp.\  23592--23608. Curran Associates, Inc., 2021.
\newblock URL
  \url{https://proceedings.neurips.cc/paper/2021/file/c68bd9055776bf38d8fc43c0ed283678-Paper.pdf}.

\bibitem[Tashiro et~al.(2021)Tashiro, Song, Song, and
  Ermon]{NEURIPS2021_cfe8504b}
Tashiro, Y., Song, J., Song, Y., and Ermon, S.
\newblock {CSDI}: Conditional score-based diffusion models for probabilistic
  time series imputation.
\newblock In Ranzato, M., Beygelzimer, A., Dauphin, Y., Liang, P., and Vaughan,
  J.~W. (eds.), \emph{Advances in Neural Information Processing Systems},
  volume~34, pp.\  24804--24816. Curran Associates, Inc., 2021.
\newblock URL
  \url{https://proceedings.neurips.cc/paper/2021/file/cfe8504bda37b575c70ee1a8276f3486-Paper.pdf}.

\bibitem[West \& Harrison(1997)West and Harrison]{West1997}
West, M. and Harrison, J.
\newblock \emph{Bayesian Forecasting and Dynamic Models}.
\newblock Springer, second edition, 1997.

\bibitem[Winkler(1996)]{Winkler1996}
Winkler, R.~L.
\newblock Scoring rules and the evaluation of probabilities.
\newblock \emph{Test}, 5\penalty0 (1):\penalty0 1–60, 1996.

\bibitem[Ziel \& Berk(2019)Ziel and Berk]{Ziel2019}
Ziel, F. and Berk, K.
\newblock Multivariate forecasting evaluation: On sensitive and strictly proper
  scoring rules.
\newblock \emph{arXiv}, pp.\  1910.07325v1, 2019.

\end{thebibliography}
\bibliographystyle{icml2023}

\newpage
\appendix
\onecolumn

\section{Proper Scoring Rules}\label{app:scoring-rules}

\subsection{Continuous Ranked Probability Score}

The Continuous Ranked Probability Score (CRPS) \cite{matheson1976scoring} can be written in multiple ways.
In particular, it can be written as a comparison between the forecast distribution cumulative density function $\Phi_{\Df}(x)$ and the realization $y \sim \Dgt$:
\begin{equation}
\text{CRPS}(y,\Df) = \int_{-\infty}^{\infty} \left(\Phi_{\Df}(x) - \mathbf{1}[x - y]\right)^2 dx,
\end{equation}
where $\mathbf{1}[x]$ is the Heaviside function.
It has been shown \cite{gneiting2007strictly} that CRPS can be rewritten in a form akin to the Energy Score:
\begin{equation}
\text{CRPS}(y,\Df) = \Esp_{x \sim \Df}|y - x| - \frac{1}{2} \Esp_{\substack{x \sim \Df\\ x' \sim \Df}}|x - x'|, \label{eq:crps_def_energy}
\end{equation}
or using quantiles: %
\begin{equation}
\text{CRPS}(y,\Df) = 2 \int_{q \in [0, 1]} \left(\mathbf{1}[\Phi^{-1}_{\Df}(q) - y] - q\right) \left(\Phi^{-1}_{\Df}(q) - y\right) dq. \label{eq:crps_def_quantiles}
\end{equation}

Since the CRPS is only defined for univariate distribution, the CRPS of a multivariate distribution is taken as the average CRPS over all dimensions.
While the univariate CRPS is a strictly proper scoring rule, the multivariate version is only proper, as it does not capture the correlations between dimensions.

In our numerical experiments, we use numerical approximations of \cref{eq:crps_def_energy,eq:crps_def_quantiles} to compute the CRPS.
The CRPS-Q (quantiles) uses the quantiles from 0.05 to 0.95 in steps of 0.05:
\begin{equation}
\text{CRPS-Q}(y,\Df) \approx \frac{1}{|Q|} \sum_{q \in Q} \left(\mathbf{1}[\Phi^{-1}_{\Df}(q) - y] - q\right) \left(\Phi^{-1}_{\Df}(q) - y\right),
\end{equation}
while the CRPS-E (expectations) uses the $\nf$ samples from $\Df$ directly:
\begin{equation}
\text{CRPS-E}(y,\Df) \approx \frac{1}{\nf} \sum_i |y - x_i| - \frac{1}{\nf (\nf - 1)} \sum_{i < i'} |x_i - x_{i'}|.
\end{equation}
Due to the double sum, CRPS-E has complexity $O(\dvar \nf^2)$, while CRPS-Q has $O(\dvar \nf \log \nf$) since quantiles can be efficiently computed after sorting the samples according to their values.

\subsection{Energy Score}

The Energy Score (ES) \cite{gneiting2007strictly} can be considered a multivariate generalization of the CRPS.
Given a parameter $0 < p < 2$ (commonly called $\beta$ in the literature, which we changed to avoid confusion with the false negative rate of \Cref{sec:power}), the ES is defined as:
\begin{equation}
\end{equation}
where $\|z\|_2$ denotes the Euclidean norm.
The ES is a strictly proper scoring rule.

We use two numerical approximations for the Energy Score:
\begin{equation}
\text{ES-Full}_p(y,\Df) \approx \frac{1}{\nf} \sum_i \|y - x_i\|_2^{p} - \frac{1}{\nf (\nf - 1)} \sum_{i < i'} \|x_i - x_{i'}\|_2^{p},
\end{equation}
which uses the full amount of data available in the sample, but has $O(\dvar \nf^2)$ complexity; and
\begin{equation}
\text{ES-Partial}_p(y,\Df) \approx \frac{1}{\nf} \sum_i \|y - x_i\|_2^{p} - \frac{1}{\nf} \sum_{i=1}^{\nf / 2} \|x_i - x_{i + \nf / 2}\|_2^{p},
\end{equation}
which uses each data point from the sample only once in the second sum, thus reducing the computing time to $O(\dvar \nf)$.

\subsection{Variogram}

The Variogram (VG) \cite{Scheuerer2015}, for a given parameter $p$, is computed as follows:
\begin{equation}
\text{VG}_p(y,\Df) = \sum_{a,b} \left( \left| y_a - y_b \right|^p - \Esp_{x \sim \Df}[\left| x_a - x_b \right|^p] \right)^2,
\end{equation}
where the sum is over all pairs of dimensions of the problem, indexed by $a$ and $b$, resulting in a $O(\dvar^2 \nf)$ complexity.
VG is a proper scoring rule but is not strictly proper since it is invariant to translations in the forecast.

\subsection{Negative Log-Likelihood}

Given the Probability Density Function (PDF) $p_\Df(x)$ of the forecast $\Df$, the negative log-likelihood (NLL) is defined as:
\begin{equation}
\text{NLL}(y,\Df) = -\log p_\Df(y).
\end{equation}
While the negative log-likelihood is strictly proper and has many other theoretical properties, it cannot be straight-forwardly estimated from a finite sample of $\Df$, so it often requires access to the PDF to be computed.

\subsection{Dawid-Sebastiani}\label{app:dawid-sebastiani}

The Dawid-Sebastiani score is computed from the first two moments of $\Df$, its mean $\mu_\Df$ and its covariance matrix $\Sigma_\Df$, as follows
\begin{equation}
\text{DS}(y,\Df) = \log \left|\det \Sigma_\Df \right| + (y - \mu_\Df)^T \Sigma_{\Df}^{-1} (y - \mu_\Df).
\end{equation}
The Dawid-Sebastiani score is very close to the negative log-likelihood for multivariate Gaussian distributions.
Unlike the log-likelihood, it can be computed from a finite sample of $\Df$ by using said sample to estimate $\mu_\Df$ and $\Sigma_\Df$.
However, if the rank of $\Sigma_{\Df}$ is upper bounded by $\nf$ ($\nf \le \dvar$), $\Sigma_{\Df}$ is guaranteed to not be full rank, hence its inverse (the concentration matrix $\Sigma_{\Df}^{-1}$) does not exist.
Thus, the Dawid-Sebastiani score is only defined for $\nf > \dvar$.

\section{Perturbations Used in the Benchmark}\label{app:perturbations}

In this section, we detail the test cases of our benchmark. 
Each test case consists of a couple of ground-truth $\Dgt$ and forecast $\Df$ distributions, specifically conceived to test a particular quality of the scoring rules under scrutiny, for instance, the ability to detect errors in a statistical moment of a significant magnitude.
All tests are parameterized by the dimensionality $\dvar$ (number of variables), and a scale parameter $\varepsilon$.

\subsection{Incorrect Marginals}

For all these distributions, each dimension of either $\Df$ or $\Dgt$ is an independent variable.
$\Df$ or $\Dgt$ are thus completely characterized by their respective marginal distributions $\{\Df_a\}_{a=1}^\dvar$ and $\{\Dgt_a\}_{a=1}^\dvar$.

\paragraph{Normal (Single, $\mu \uparrow$)}
\begin{equation}
\begin{aligned}
    \Dgt_1 & \sim \Norm(\varepsilon, 1) &  \\
    \Dgt_a & \sim \Norm(0, 1) & \forall a \neq 1 \\
    \Df_a & \sim \Norm(0, 1) & \forall a
\end{aligned}
\end{equation}

\paragraph{Normal (All, $\mu \uparrow$)}
\begin{equation}
\begin{aligned}
    \Dgt_a & \sim \Norm(\varepsilon, 1) & \forall a \\
    \Df_a & \sim \Norm(0, 1) & \forall a
\end{aligned}
\end{equation}

\paragraph{Normal (Single, $\sigma \uparrow$) and Normal (Single, $\sigma \downarrow$)}
\begin{equation}
\begin{aligned}
    \Dgt_1 & \sim \Norm(0, \varepsilon^2) & \\
    \Dgt_a & \sim \Norm(0, 1) & \forall a \neq 1 \\
    \Df_a & \sim \Norm(0, 1) & \forall a
\end{aligned}
\end{equation}

\paragraph{Normal (All, $\sigma \uparrow$) and Normal (All, $\sigma \downarrow$)}
\begin{equation}
\begin{aligned}
    \Dgt_a & \sim \Norm(0, \varepsilon^2) & \forall a \\
    \Df_a & \sim \Norm(0, 1) & \forall a
\end{aligned}
\end{equation}

\paragraph{Exponential (Single, $\mu \uparrow$) and Exponential (Single, $\mu \downarrow$)}
\begin{equation}
\begin{aligned}
    \Dgt_1 & \sim \Exp(1 / \varepsilon) & \\
    \Dgt_a & \sim \Exp(1) & \forall a \neq 1 \\
    \Df_a & \sim \Exp(1) & \forall a
\end{aligned}
\end{equation}
where $\Exp(1)(\lambda)$ is an exponential distribution with rate of change $\lambda$, and mean $1 / \lambda$.

\paragraph{Exponential (All, $\mu \uparrow$) and Exponential (All, $\mu \downarrow$)}
\begin{equation}
\begin{aligned}
    \Dgt_a & \sim \Exp(1 / \varepsilon) & \forall a \\
    \Df_a & \sim \Exp(1) & \forall a
\end{aligned}
\end{equation}

\paragraph{Skew Normal (All, $\alpha \downarrow$)}
\begin{equation}
\begin{aligned}
    \Dgt_a & \sim \Skew(\xi(\varepsilon), \omega(\varepsilon), \varepsilon) & \forall a \\
    \Df_a & \sim \Norm(0, 1) & \forall a
\end{aligned}
\end{equation}
where $\Skew(\xi, \omega, \alpha)$ is a skew normal distribution with location $\xi$, scale $\omega$, and shape $\alpha$ as parameters.
$\xi$ and $\omega$ are chosen such that the resulting distribution has a mean of 0 and a variance of 1.
Note that we recover the $\Norm(0,1)$ distribution when $\varepsilon = 0$.

\subsection{Incorrect Correlations}

For all these distributions, only the copulas of $\Df$ and $\Dgt$ differ, so their marginals are always identical.
For simplicity, we only selected multivariate normal distributions, due to the ease by which their marginals can be selected to always be $\Norm(0,1)$.
Therefore, all ground-truth distributions in this section are of the form $\Dgt \sim N(0, \Sigma_{\Dgt})$ and all forecast distributions are of the form $\Df \sim N(0, \Sigma_{\Df})$, for various covariance matrices $\Sigma_{\Dgt}$ and $\Sigma_{\Df}$.

\paragraph{Full Cov (Missing)}
\begin{equation}
\begin{aligned}
    \Sigma_{\Dgt,aa} & = 1 & \forall a \\
    \Sigma_{\Dgt,ab} & = \varepsilon & \forall a \neq b \\
    \Sigma_{\Df,aa} & = 1 & \forall a \\
    \Sigma_{\Df,ab} & = 0 & \forall a \neq b
\end{aligned}
\end{equation}

\paragraph{Full Cov (Extra)}
\begin{equation}
\begin{aligned}
    \Sigma_{\Dgt,aa} & = 1 & \forall a \\
    \Sigma_{\Dgt,ab} & = 0 & \forall a \neq b \\    
    \Sigma_{\Df,aa} & = 1 & \forall a \\
    \Sigma_{\Df,ab} & = \varepsilon & \forall a \neq b
\end{aligned}
\end{equation}

\paragraph{Checker Cov (Missing)}
\begin{equation}
\begin{aligned}
    \Sigma_{\Dgt,aa} & = 1 & \forall a \\
    \Sigma_{\Dgt,ab} & = (-1)^{a+b} \varepsilon & \forall a \neq b \\
    \Sigma_{\Df,aa} & = 1 & \forall a \\
    \Sigma_{\Df,ab} & = 0 & \forall a \neq b
\end{aligned}
\end{equation}

\paragraph{Checker Cov (Extra)}
\begin{equation}
\begin{aligned}
    \Sigma_{\Df,aa} & = 1 & \forall a \\
    \Sigma_{\Df,ab} & = 0 & \forall a \neq b \\
    \Sigma_{\Dgt,aa} & = 1 & \forall a \\
    \Sigma_{\Dgt,ab} & = (-1)^{a+b} \varepsilon & \forall a \neq b
\end{aligned}
\end{equation}

\paragraph{Block Cov (Missing)}
\begin{equation}
\begin{aligned}
    \Sigma_{\Dgt} & =
        \begin{pmatrix}
            \bar{\Sigma}_\Dgt\\
            &\ddots\\
            &&\bar{\Sigma}_\Dgt
        \end{pmatrix} \\
    \bar{\Sigma}_\Dgt & =
        \begin{pmatrix}
            1 & \varepsilon \\
            \varepsilon & 1 \\
        \end{pmatrix} \\
    \Sigma_{\Df} & = I
\end{aligned}
\end{equation}

\paragraph{Block Cov (Extra)}
\begin{equation}
\begin{aligned}
    \Sigma_{\Dgt} & = I \\
    \Sigma_{\Df} & =
        \begin{pmatrix}
            \bar{\Sigma}_\Df\\
            &\ddots\\
            &&\bar{\Sigma}_\Df
        \end{pmatrix} \\
    \bar{\Sigma}_\Df & =
        \begin{pmatrix}
            1 & \varepsilon \\
            \varepsilon & 1 \\
        \end{pmatrix}
\end{aligned}
\end{equation}

\subsection{Mixture of Distributions}

\paragraph{Mixture (Missing)}
\begin{equation}
\begin{aligned}
    \Dgt & \sim \text{Mixture}\left(\frac{\Norm(\varepsilon, I)}{2} + \frac{\Norm(-\varepsilon, I)}{2}\right) \\
    \Df & \sim \Norm(0, I + \varepsilon^2)
\end{aligned}
\end{equation}
The mean and covariance matrix of $\Df$ has been chosen to have the same mean and covariance as the mixture used for $\Dgt$.

\paragraph{Mixture (Extra)}
\begin{equation}
\begin{aligned}
    \Dgt & \sim \Norm(0, I + \varepsilon^2) \\
    \Df & \sim \text{Mixture}\left(\frac{\Norm(\varepsilon, I)}{2} + \frac{\Norm(-\varepsilon, I)}{2}\right)
\end{aligned}
\end{equation}
The mean and covariance matrix of $\Dgt$ has been chosen to have the same mean and covariance as the mixture used for $\Df$.

\subsection{Tuning Results}

\begin{table}
\centering
\caption{Values of the $\varepsilon$ parameter from the tuning procedure which requires the NLL to have a statistical power $1 - \beta$ equals to 80\%, as described in \cref{sec:test_cases_and_calibration}. The values of $\varepsilon$ which make the forecast distribution identical to the ground-truth distribution are 0 for Normal $\mu$, Skew Normal $\alpha$, all covariance test cases, and the mixture test cases; and 1 for Normal $\sigma$ and Exponential $\mu$.}
\label{tab:calibration_results}
\begin{tabular}{lrrrrrrrrr}
\toprule
$\mathbf{\dvar}$ &   \textbf{16}   &   \textbf{32}   &   \textbf{64}   &   \textbf{128}  &   \textbf{256}  &   \textbf{512}  &   \textbf{1024} &   \textbf{2048} &   \textbf{4096} \\
\midrule
\textbf{Normal (Single,$\mu \uparrow$)       } & 0.9079 & 0.9079 & 0.9079 & 0.9079 & 0.9079 & 0.9079 & 0.9079 & 0.9079 & 0.9079 \\
\textbf{Normal (All,$\mu \uparrow$)          } & 0.2270 & 0.1605 & 0.1135 & 0.0802 & 0.0567 & 0.0401 & 0.0284 & 0.0201 & 0.0142 \\
\textbf{Normal (Single,$\sigma \downarrow$)  } & 0.5799 & 0.5799 & 0.5799 & 0.5799 & 0.5799 & 0.5799 & 0.5799 & 0.5799 & 0.5799 \\
\textbf{Normal (Single,$\sigma \uparrow$)    } & 2.4514 & 2.4514 & 2.4514 & 2.4514 & 2.4514 & 2.4514 & 2.4514 & 2.4514 & 2.4514 \\
\textbf{Normal (All,$\sigma \downarrow$)     } & 0.8584 & 0.8963 & 0.9248 & 0.9458 & 0.9612 & 0.9723 & 0.9803 & 0.9860 & 0.9901 \\
\textbf{Normal (All,$\sigma \uparrow$)       } & 1.1855 & 1.1254 & 1.0860 & 1.0596 & 1.0415 & 1.0291 & 1.0204 & 1.0144 & 1.0101 \\
\textbf{Exponential (Single,$\mu \downarrow$)} & 0.4481 & 0.4487 & 0.4447 & 0.4481 & 0.4538 & 0.4528 & 0.4463 & 0.4463 & 0.4493 \\
\textbf{Exponential (Single,$\mu \uparrow$)  } & 3.0032 & 3.0395 & 2.9980 & 3.0000 & 3.0316 & 3.0303 & 3.0327 & 3.0497 & 3.0514 \\
\textbf{Exponential (All,$\mu \downarrow$)   } & 0.8028 & 0.8539 & 0.8932 & 0.9233 & 0.9451 & 0.9609 & 0.9721 & 0.9800 & 0.9859 \\
\textbf{Exponential (All,$\mu \uparrow$)     } & 1.2666 & 1.1778 & 1.1209 & 1.0838 & 1.0584 & 1.0411 & 1.0289 & 1.0202 & 1.0142 \\
\textbf{Skew Normal (All,$\alpha \downarrow$)  } & 2.3987 & 1.8090 & 1.4738 & 1.2036 & 1.0149 & 0.8744 & 0.7532 & 0.6555 & 0.5748 \\
\textbf{Full Cov (Missing)                   } & 0.2055 & 0.1218 & 0.0680 & 0.0363 & 0.0188 & 0.0096 & 0.0048 & 0.0024 & 0.0012 \\
\textbf{Full Cov (Extra)                     } & 0.1268 & 0.0629 & 0.0312 & 0.0155 & 0.0077 & 0.0039 & 0.0019 & 0.0010 & 0.0005 \\
\textbf{Checker Cov (Missing)                } & 0.2055 & 0.1218 & 0.0680 & 0.0363 & 0.0188 & 0.0096 & 0.0048 & 0.0024 & 0.0012 \\
\textbf{Checker Cov (Extra)                  } & 0.1268 & 0.0629 & 0.0312 & 0.0155 & 0.0077 & 0.0039 & 0.0019 & 0.0010 & 0.0005 \\
\textbf{Block Cov (Missing)                  } & 0.3058 & 0.2214 & 0.1585 & 0.1128 & 0.0800 & 0.0567 & 0.0401 & 0.0284 & 0.0201 \\
\textbf{Block Cov (Extra)                    } & 0.3201 & 0.2268 & 0.1605 & 0.1135 & 0.0802 & 0.0567 & 0.0401 & 0.0284 & 0.0201 \\
\textbf{Mixture (Missing)                    } & 0.5906 & 0.4151 & 0.2974 & 0.2083 & 0.1480 & 0.1053 & 0.0739 & 0.0519 & 0.0367 \\
\textbf{Mixture (Extra)                      } & 0.8020 & 0.5749 & 0.4052 & 0.2909 & 0.2040 & 0.1456 & 0.1032 & 0.0727 & 0.0516 \\
\bottomrule
\end{tabular}
\end{table}

The tuned values of $\varepsilon$ are listed in \cref{tab:calibration_results} for all test cases and dimensions under consideration in this paper.
We computed $\ex{y, X^{gt}_\nf, X^{f}_\nf}{\Delta_\nf}$ and $\variance{y, X^{gt}_\nf, X^{f}_\nf}{\Delta_\nf}$ analytically for the Gaussian test cases, which gave us a numerically precise function of the NLL statistical power in term of $\varepsilon$.
Since this function is strictly monotonic in the ranges we are interested in, we used a bisection algorithm to quickly find the correct tuned values.

For the non-Gaussian test cases, $\ex{y, X^{gt}_\nf, X^{f}_\nf}{\Delta_\nf}$ and $\variance{y, X^{gt}_\nf, X^{f}_\nf}{\Delta_\nf}$ were estimated numerically with a sample size of 10'000.
The sampling was done with a fixed random number generator seed, to avoid numerical jitter which breaks the monotonicity condition which we relied upon in the bisection algorithm.
The randomness inherent to this procedure explains why $\varepsilon$ is not constant for Exponential (Single,$\mu \downarrow$) and Exponential (Single,$\mu \uparrow$) even though they should be if the tuning was done exactly.

\section{Extensive Benchmark Results}\label{app:results}

Our results for all of our test cases and scoring rules are presented in \cref{fig:result_variogram,fig:full_page_wrong_mean_single,fig:full_page_wrong_mean_all,fig:full_page_wrong_std_single_lower,fig:full_page_wrong_std_single_higher,fig:full_page_wrong_std_all_lower,fig:full_page_wrong_std_all_higher,fig:full_page_wrong_exponential_single_lower,fig:full_page_wrong_exponential_single_higher,fig:full_page_wrong_exponential_all_lower,fig:full_page_wrong_exponential_all_higher,fig:full_page_missing_skew_all,fig:full_page_missing_covariance_full,fig:full_page_extra_covariance_full,fig:full_page_missing_covariance_checker,fig:full_page_extra_covariance_checker,fig:full_page_missing_covariance_block,fig:full_page_extra_covariance_block,fig:full_page_missing_mixture,fig:full_page_extra_mixture}.
To help distinguish \ror{0.8}, \ror{0.5}, and \ror{0.2}, we added contour lines for $1 - \beta = 0.8$, $1 - \beta = 0.5$, and $1 - \beta = 0.2$.
Since directly using the raw values would result in quite rough contour lines, we first smoothed our data using a Radial Basis Function interpolation (with the Thin Plate Spline function), with $\log_2 d$ and $\log_2 m$ as the independent variables.

\subsection{Additional Remarks}\label{app:results-general-remarks}

\paragraph{Non-Monotonic Statistical Power}
While statistical power generally improves when increasing $\nf$, we notice that it is not necessarily monotonic over the number of dimensions of our tests.
Consider the results of \cref{fig:result_variogram_non_monotonic} as an example.
VG's precision is highest around $d = 2^7$, and is lower for both higher and lower dimensionality.
We explain this unexpected behavior by the way we tuned the ground-truth and forecast distributions of our test cases, such that the negative log-likelihood has a statistical power of 80\%.
This non-monotonicity can be the result of the negative log-likelihood and the variogram having a different dependency on the dimensionality, making problems with a higher number of variables not necessarily harder than problems with a lower number of variables.

\paragraph{Asymmetry of Test Cases}
\begin{figure*}[h]
    \centering
    \includegraphics[width=\linewidth]{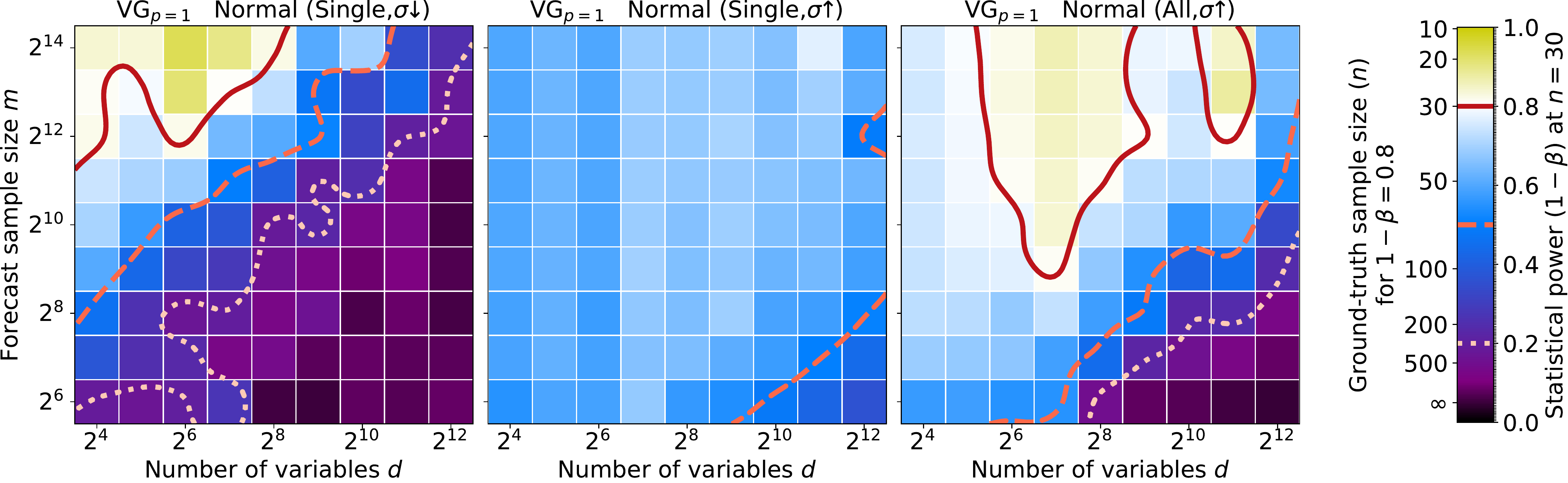}
    \caption{Statistical power of \emph{VG} at correctly detecting that forecast's marginal variances are different from the ground-truth's ones: higher (left) or lower (middle) on a single dimension, or lower on all dimensions (right).}
    \label{fig:result_variogram}
\end{figure*}
In most of our tests, a score is not in general equally powerful at detecting whether forecasts underestimate a feature than at detecting forecasts that overestimate the same feature.
For instance, \cref{fig:result_variogram} reports the statistical power of VG for a forecast that either underestimates or overestimates the standard deviation on a single dimension.
While this scoring rule has a decent statistical power almost everywhere (true positive rate above 50\% and minimal $\ngt$ around $50$) in the former case, it requires many more samples ($\nf \ge 2^4 \dvar$)
to reach the same power that it has in the latter case.
However, the scoring rule continues to improve as the number of samples increases when the forecast is sharp, eventually achieving greater power than the negative log-likelihood ($1 - \beta > 0.8$) for $\nf \ge 2^{13} \cap \dvar \leq 2^7$.

\subsection{Study for Varying Problem Complexity}
\begin{figure}[p]
    \centering
    \includegraphics[width=0.6\linewidth]{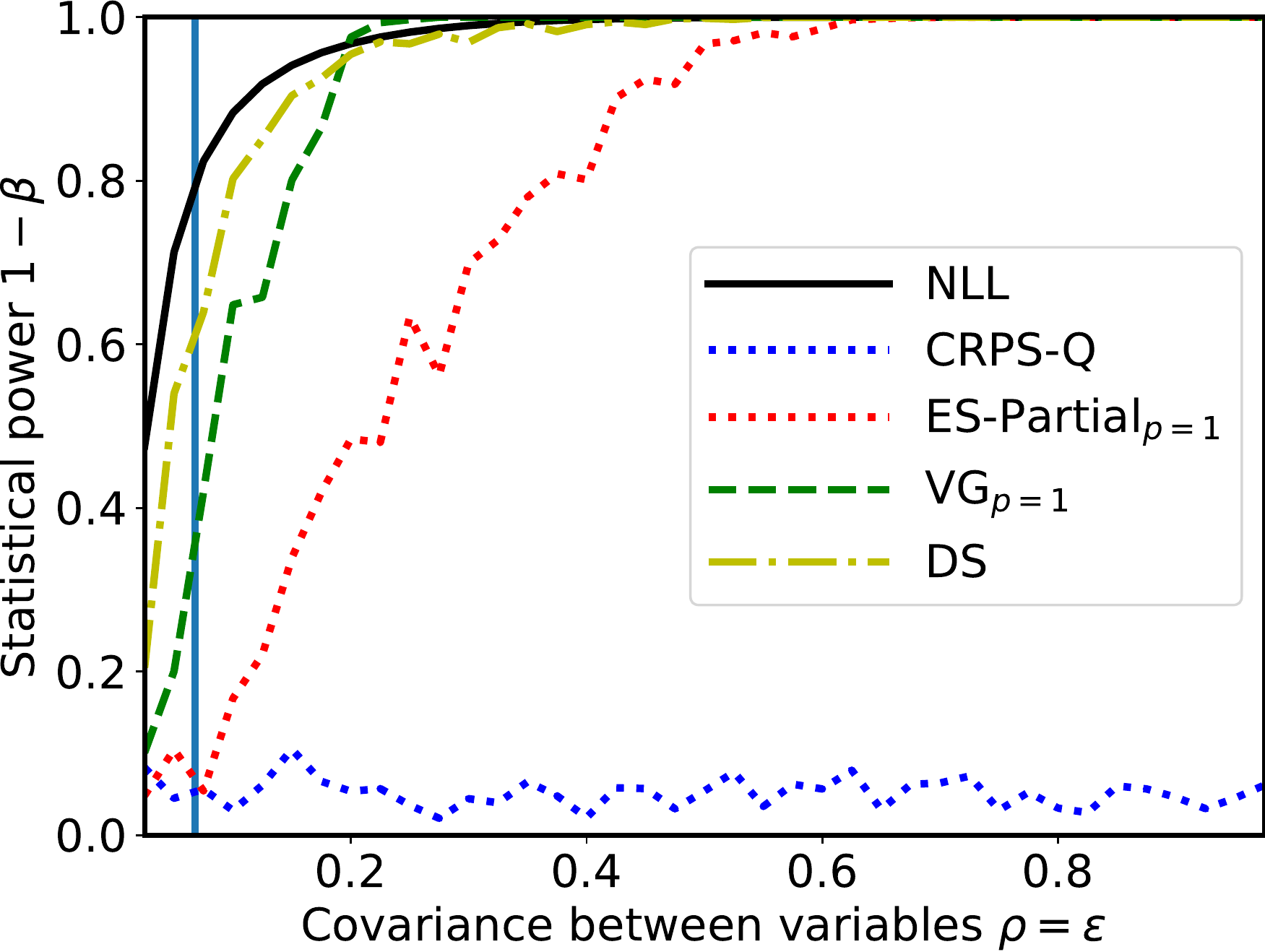}
    \caption{Statistical power of all scoring rules as a function of the simplicity of the problem (the higher $\varepsilon$ the simpler the task) for $\ngt = 30$, $\nf = 2^{12}$ and $\dvar=16$.
    The test case corresponds to a forecast with all independent variables, while they are all positively correlated in the ground truth, for Normal distribution marginals.
    The vertical line marks the value of $\varepsilon$ at which NLL has 80\% power, corresponding to the main setting of our other studies.}
    \label{fig:calib-corr}
\end{figure}

The results presented so far are obtained on test cases tuned w.r.t. the NLL, i.e., on ground-truth and forecast distributions generated so that the NLL can distinguish them with 80\% statistical power.
The rationale was to ensure that (i) the synthetic discrepancies could be captured at least by one scoring rule and (ii) the difficulty of the different test cases was comparable.
Doing so we however limited our analysis to a single (albeit compelling) level of difficulty.
In~\cref{fig:calib-corr} we study the impact of this additional factor on the power of the scoring rules, for the test case where the ground-truth correlations are dropped (\textbf{Full Cov (Missing)}).
We control the difficulty of the problem by varying the correlation parameter $\varepsilon$:
the higher $\varepsilon$ the bigger the discrepancy, hence the easier it is to distinguish the two distributions.
We observe that, apart from the CRPS which is known to be insensitive to this particular type of discrepancy, all scoring rules converge to perfect statistical power. 
However, they show different convergence rates, with the DS and VG being significantly more reliable than the ES on the most difficult settings ($\varepsilon \le 0.4$). 

\subsection{Sizes of the Regions of Reliability}
\begin{figure}[ht]
    \centering
    \includegraphics[width=0.5\linewidth]{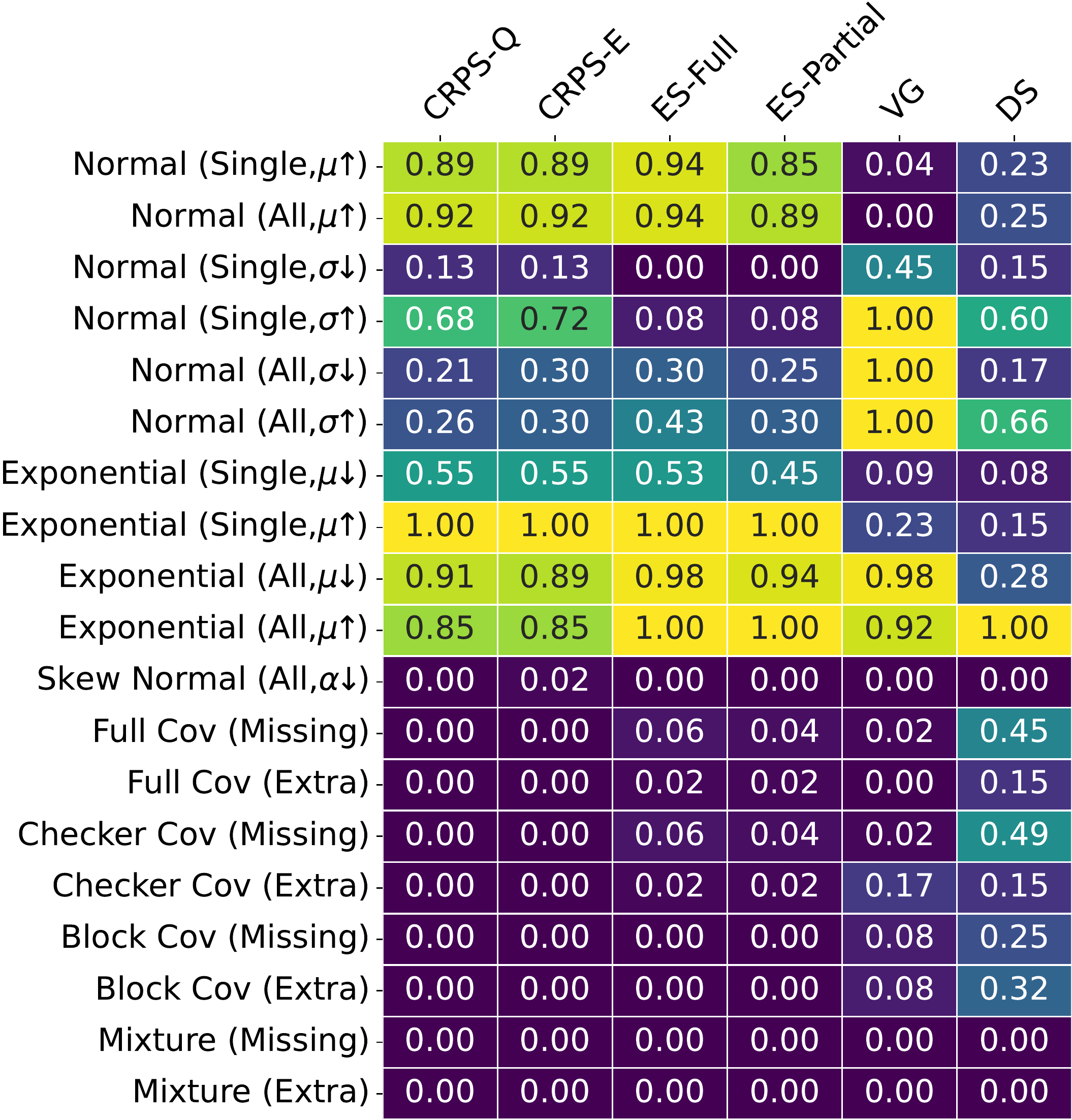}
    \caption{Summary of results: proportion of our experiments that are in \ror{0.5} amongst those where the sample size is higher than the number of variables $m > d$.}
    \label{fig:ror_50_ratio_subset}
\end{figure}

As an alternative point of view to assess the quality of the various scoring rules, \cref{fig:ror_50_ratio_subset} shows how large each \ror{0.5} are in our experiments.
Very similar conclusions can be taken from it than from \cref{fig:best_mean_power_subset}, since the complementarity between scoring rules, and which test cases cause issues for them, are still quite apparent.
However, it reveals test cases where some scoring rules never reach the reasonable $1 - \beta = 0.5$ threshold.

\begin{figure}[p]
    \centering
    \includegraphics[width=0.9\linewidth]{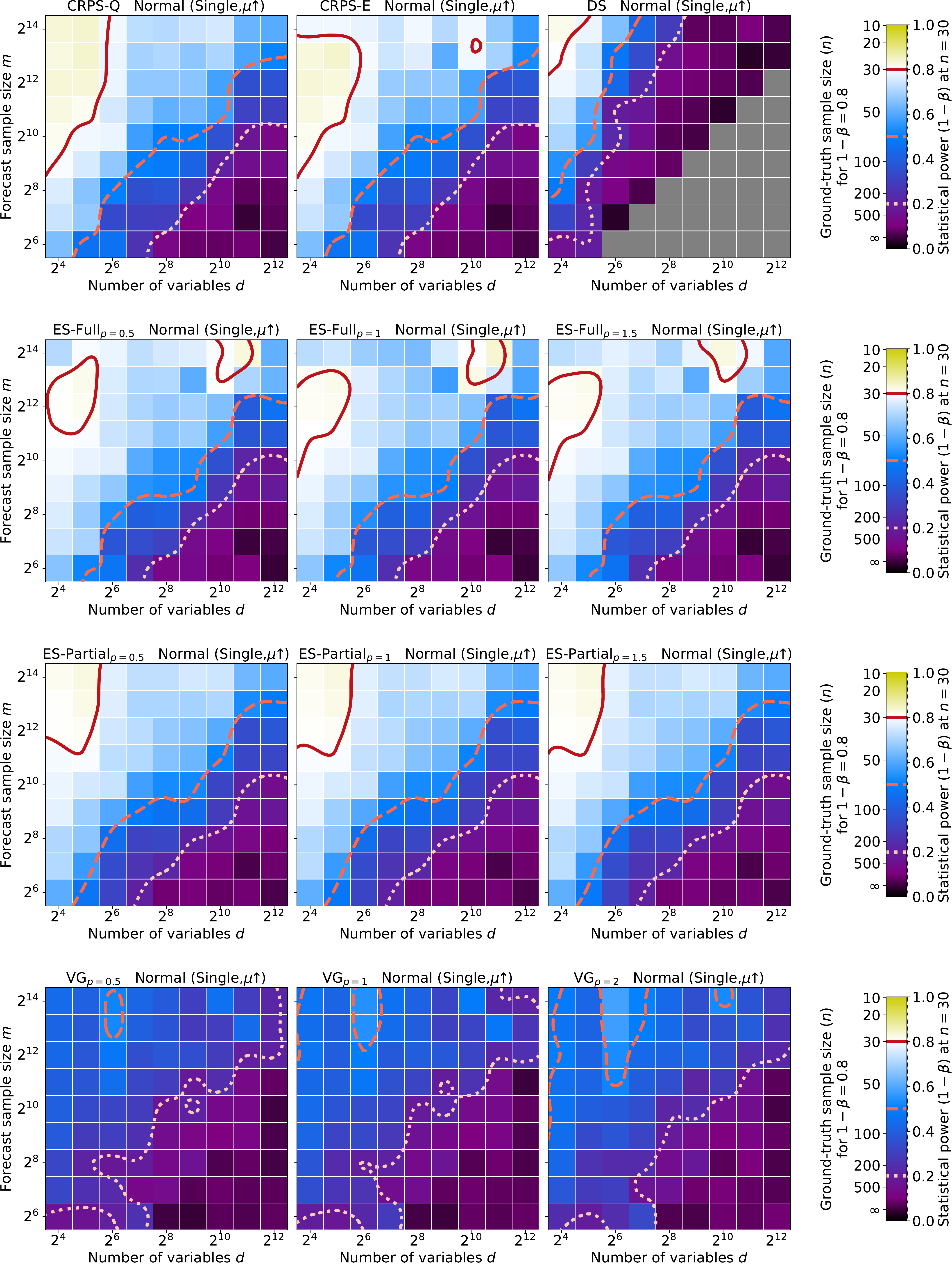}
    \caption{Statistical power of all scoring rules at correctly detecting that forecast's marginal mean is different than the ground-truth's one for a single dimension, for Normal distribution marginals.
    DS cannot be computed when $d \ge m$, so the corresponding area is greyed out.}
    \label{fig:full_page_wrong_mean_single}
\end{figure}

\begin{figure}[p]
    \centering
    \includegraphics[width=0.9\linewidth]{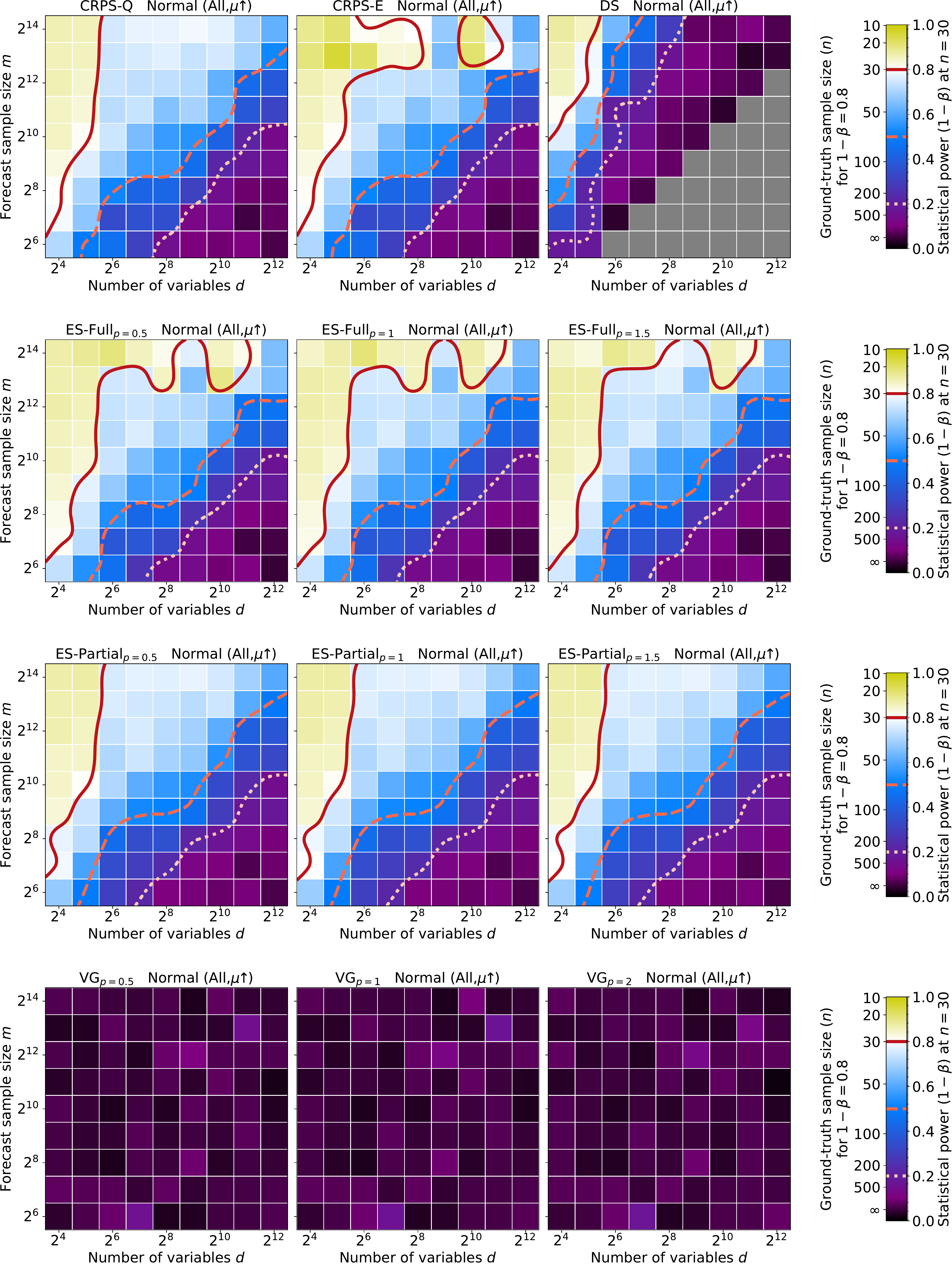}
    \caption{Statistical power of all scoring rules at correctly detecting that forecast's marginal means are different than the ground-truth's ones for all single dimensions, for Normal distribution marginals.
    DS cannot be computed when $d \ge m$, so the corresponding area is greyed out.}
    \label{fig:full_page_wrong_mean_all}
\end{figure}

\begin{figure}[p]
    \centering
    \includegraphics[width=0.9\linewidth]{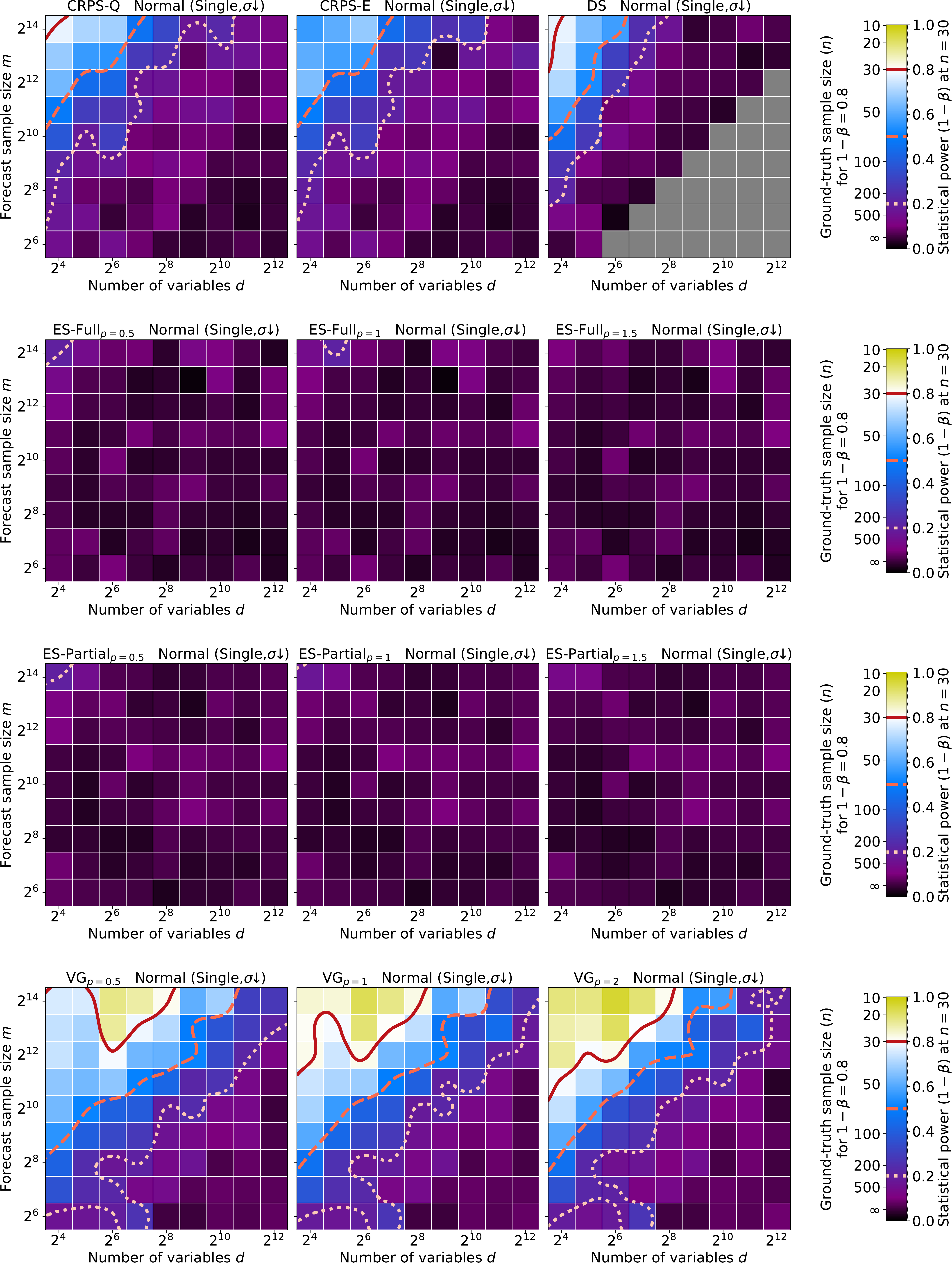}
    \caption{Statistical power of all scoring rules at correctly detecting that forecast's marginal standard deviation is lower than the ground-truth's one for a single dimension, for Normal distribution marginals.
    DS cannot be computed when $d \ge m$, so the corresponding area is greyed out.}
    \label{fig:full_page_wrong_std_single_lower}
\end{figure}

\begin{figure}[p]
    \centering
    \includegraphics[width=0.9\linewidth]{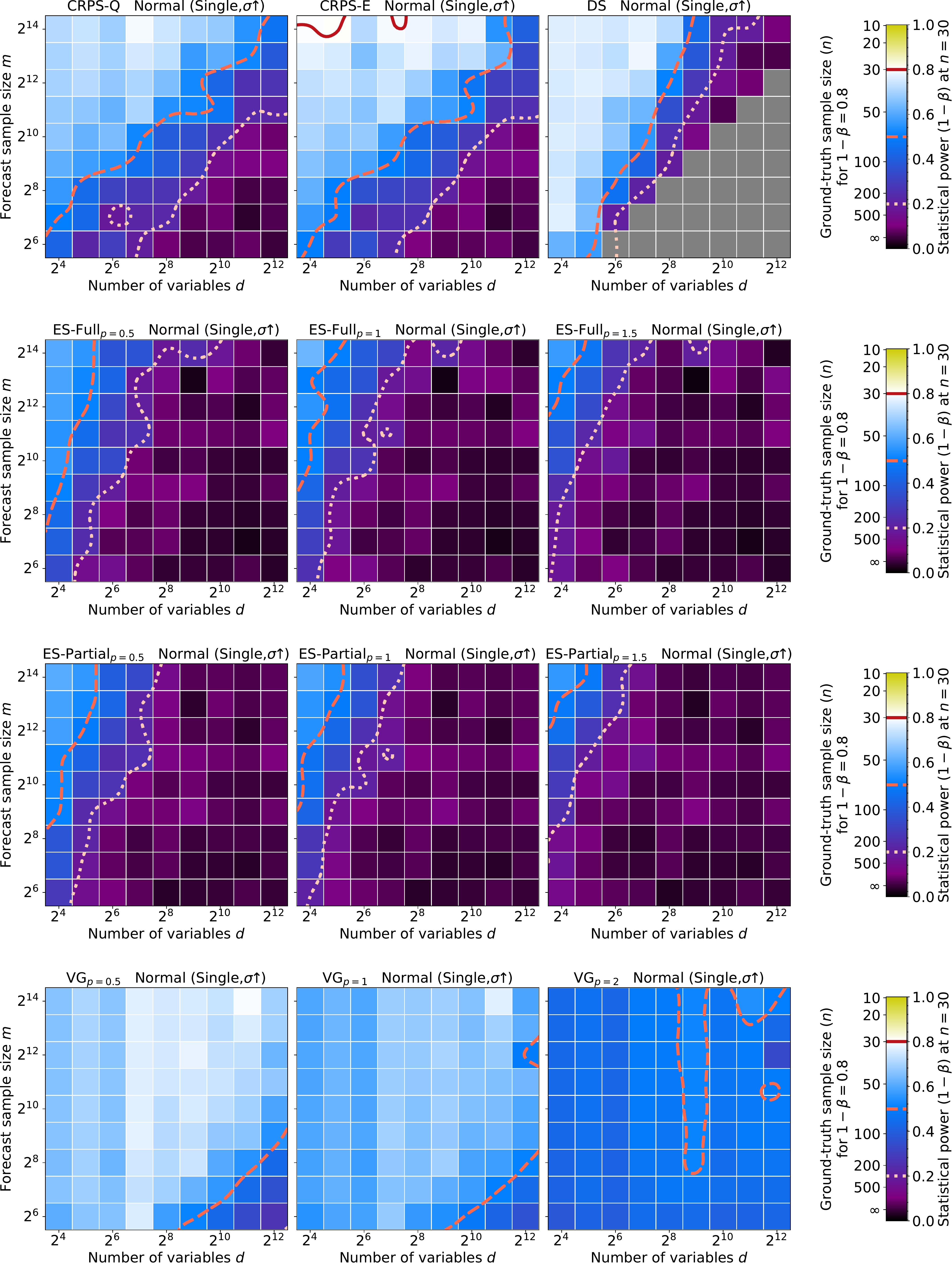}
    \caption{Statistical power of all scoring rules at correctly detecting that forecast's marginal standard deviation is higher than the ground-truth's one for a single dimension, for Normal distribution marginals.
    DS cannot be computed when $d \ge m$, so the corresponding area is greyed out.}
    \label{fig:full_page_wrong_std_single_higher}
\end{figure}

\begin{figure}[p]
    \centering
    \includegraphics[width=0.9\linewidth]{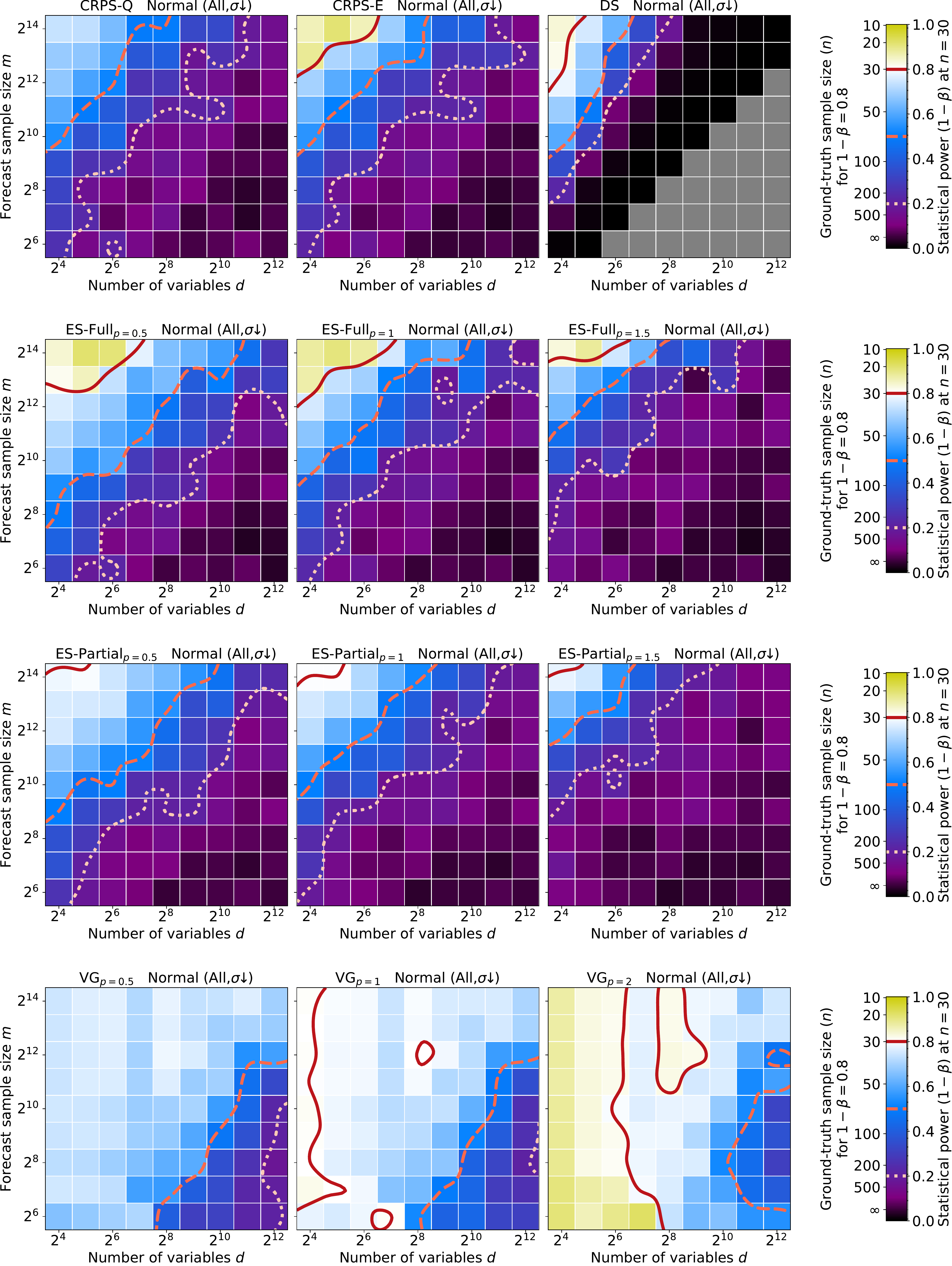}
    \caption{Statistical power of all scoring rules at correctly detecting that forecast's marginal standard deviations are lower than the ground-truth's ones for all dimensions, for Normal distribution marginals.
    DS cannot be computed when $d \ge m$, so the corresponding area is greyed out.}
    \label{fig:full_page_wrong_std_all_lower}
\end{figure}

\begin{figure}[p]
    \centering
    \includegraphics[width=0.9\linewidth]{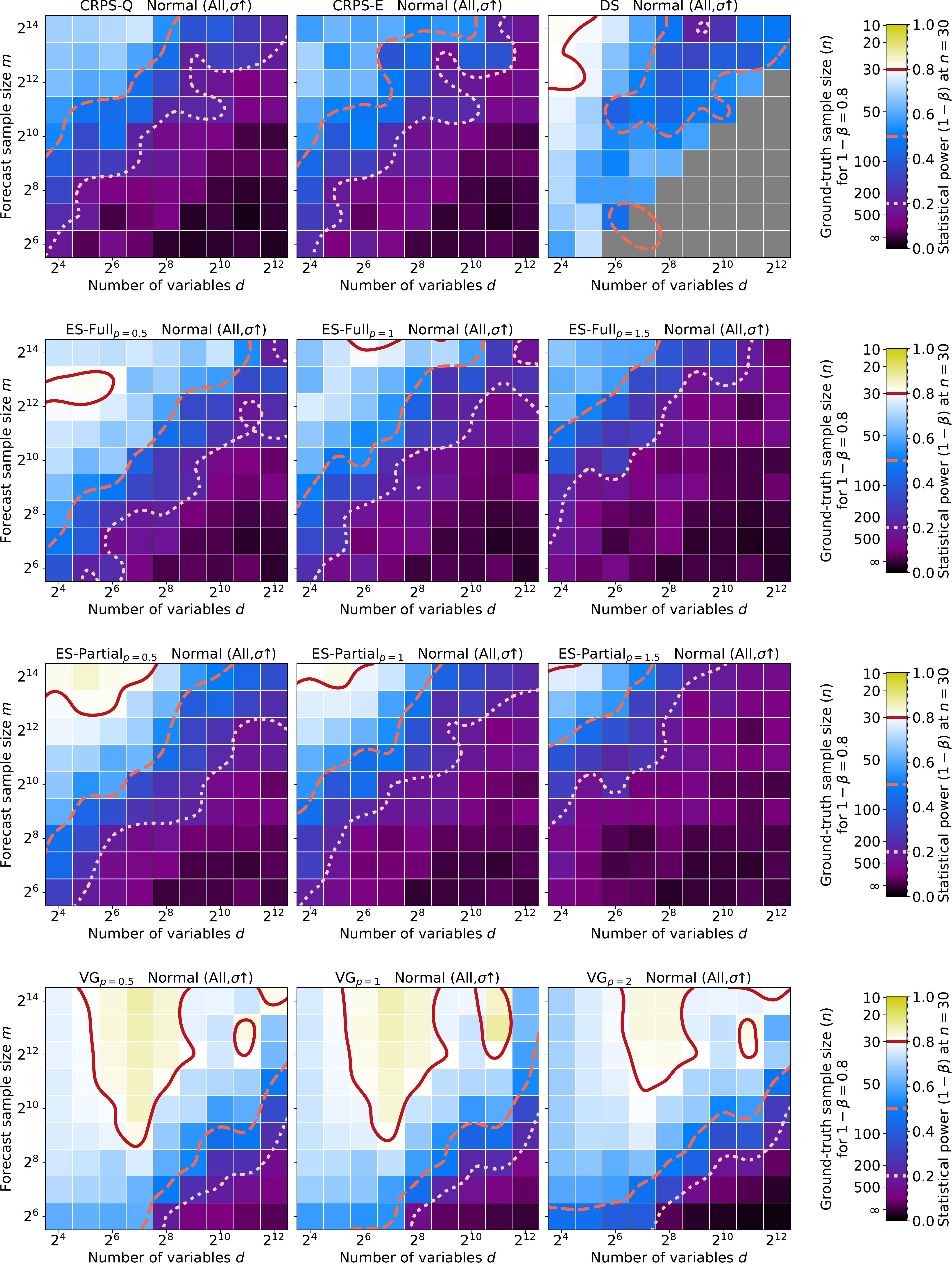}
    \caption{Statistical power of all scoring rules at correctly detecting that forecast's marginal standard deviations are lower than the ground-truth's ones for all dimensions, for Normal distribution marginals.
    DS cannot be computed when $d \ge m$, so the corresponding area is greyed out.}
    \label{fig:full_page_wrong_std_all_higher}
\end{figure}

\begin{figure}[p]
    \centering
    \includegraphics[width=0.9\linewidth]{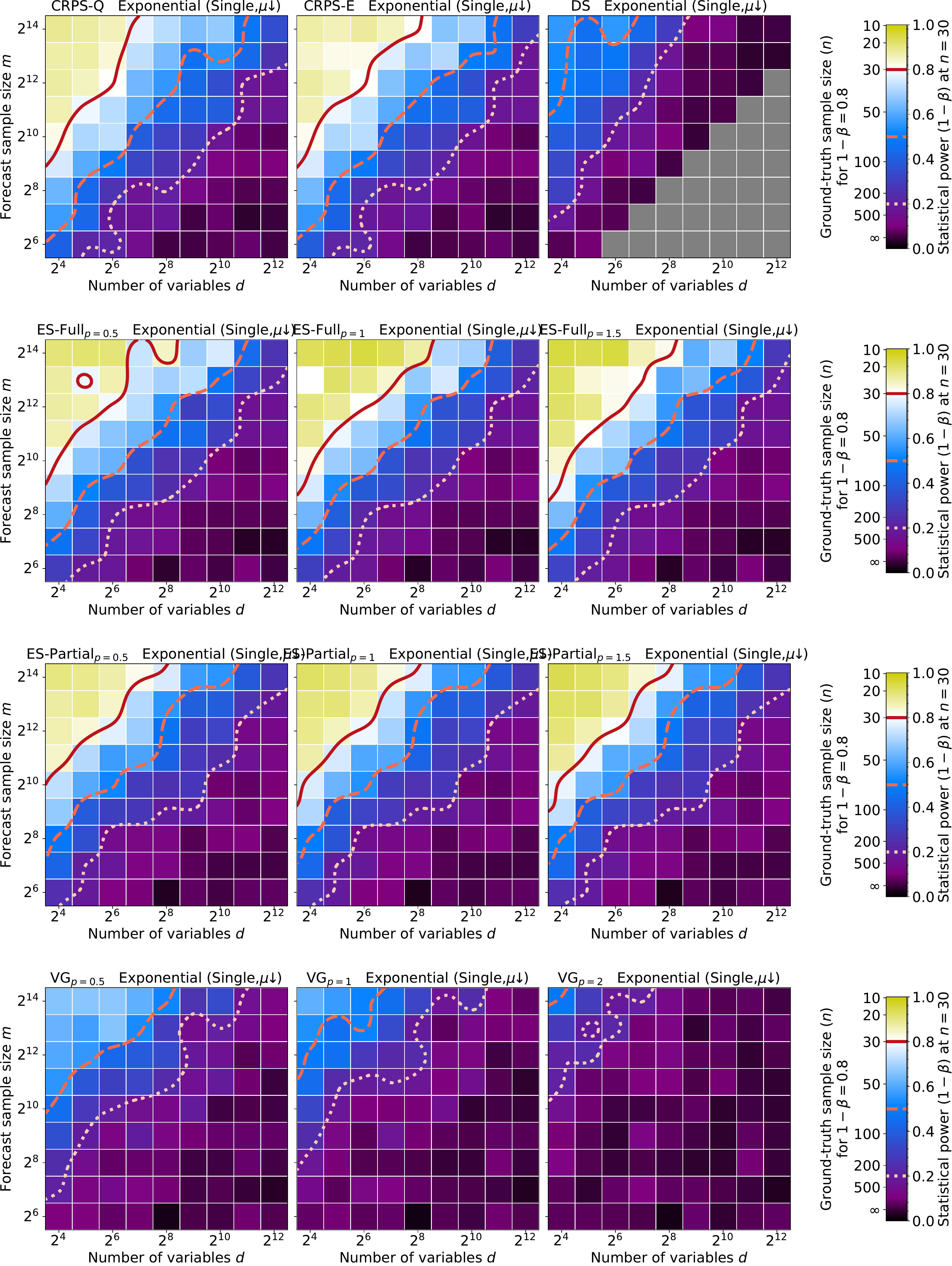}
    \caption{Statistical power of all scoring rules at correctly detecting that forecast's marginal mean is lower than the ground-truth's one for a single dimension, for Exponential distribution marginals.
    DS cannot be computed when $d \ge m$, so the corresponding area is greyed out.}
    \label{fig:full_page_wrong_exponential_single_lower}
\end{figure}

\begin{figure}[p]
    \centering
    \includegraphics[width=0.9\linewidth]{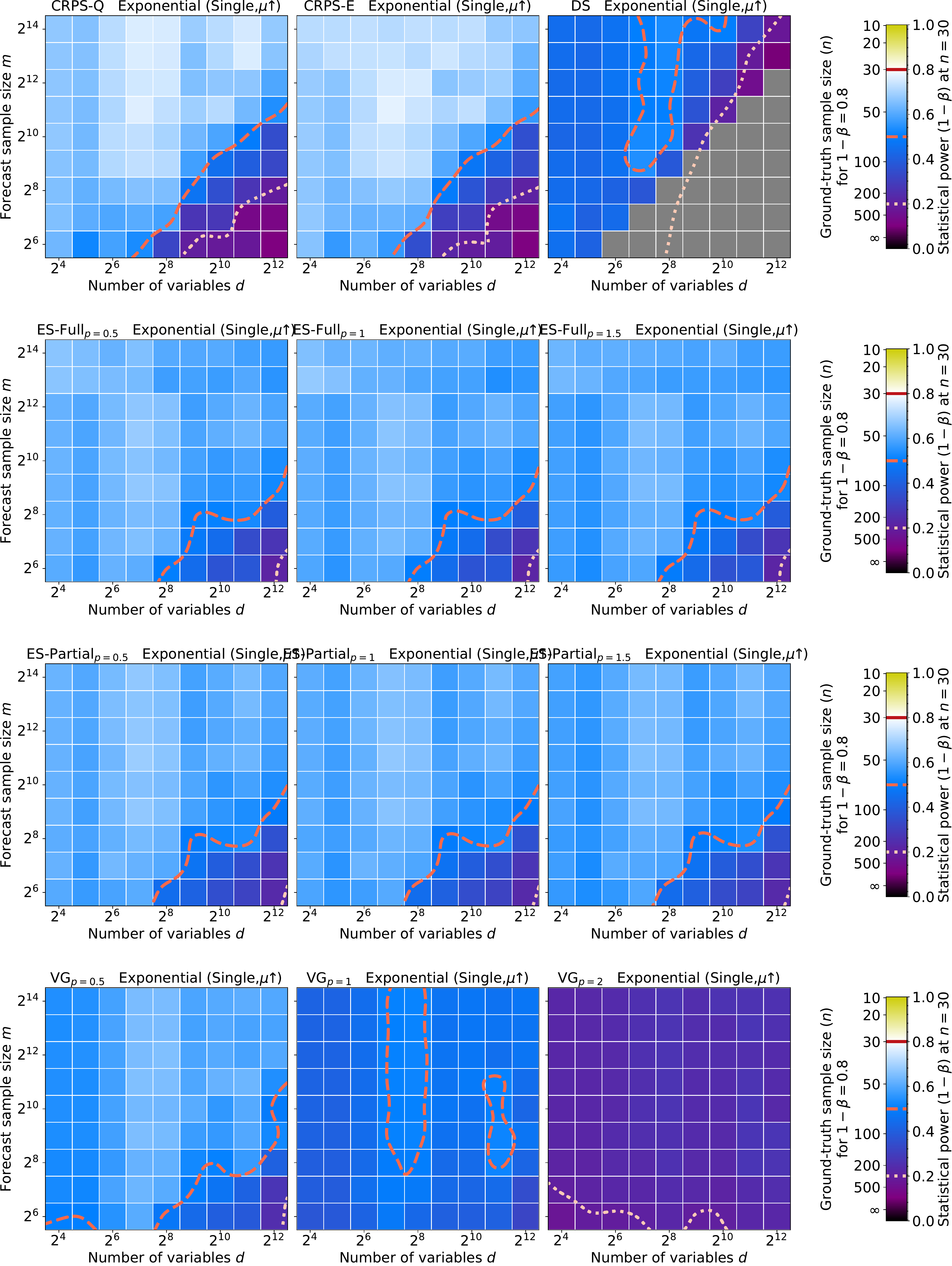}
    \caption{Statistical power of all scoring rules at correctly detecting that forecast's marginal mean is higher than the ground-truth's one for a single dimension, for Exponential distribution marginals.
    DS cannot be computed when $d \ge m$, so the corresponding area is greyed out.}
    \label{fig:full_page_wrong_exponential_single_higher}
\end{figure}

\begin{figure}[p]
    \centering
    \includegraphics[width=0.9\linewidth]{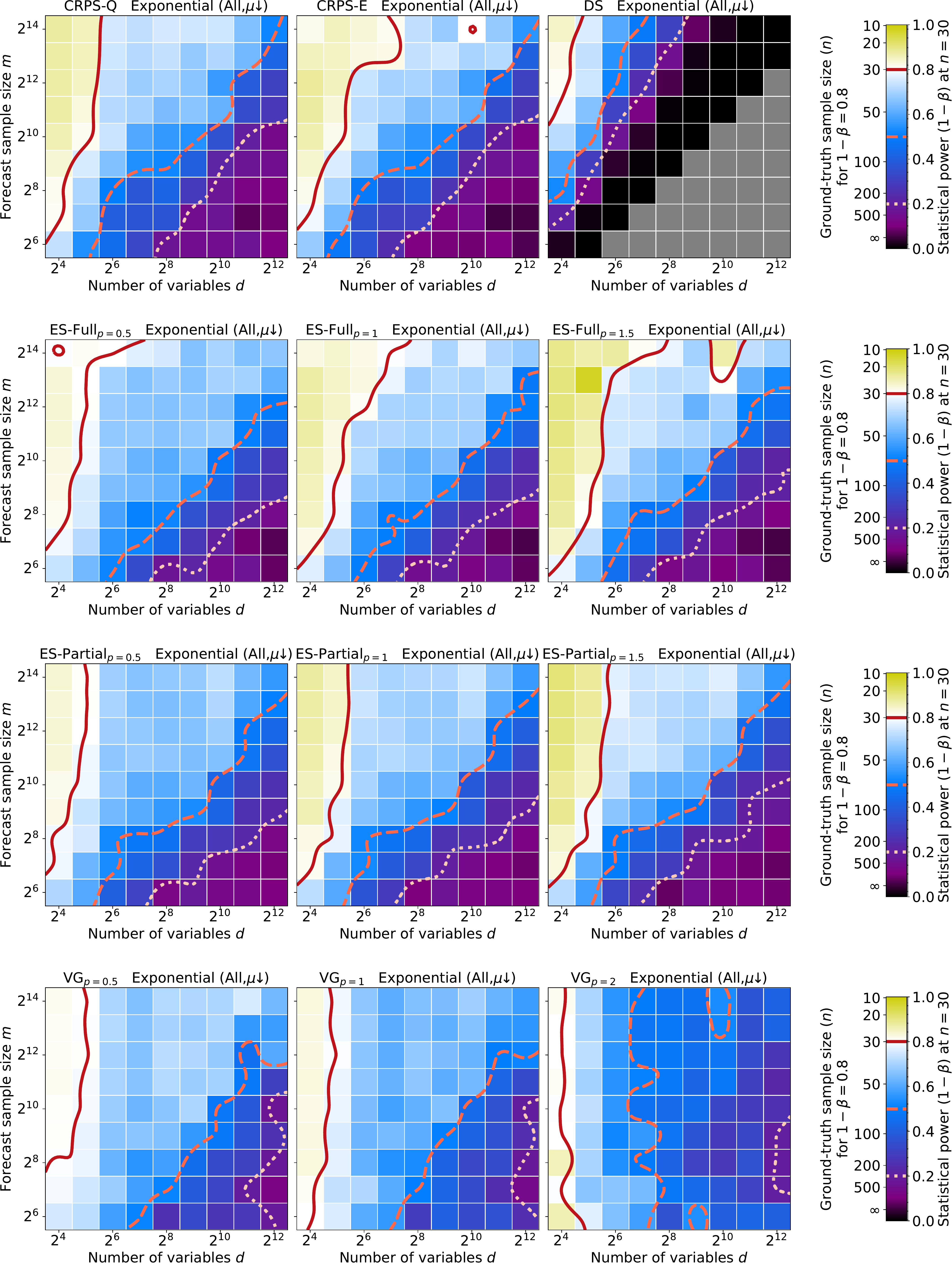}
    \caption{Statistical power of all scoring rules at correctly detecting that forecast's marginal means are lower than the ground-truth's ones for all dimensions, for Exponential distribution marginals.
    DS cannot be computed when $d \ge m$, so the corresponding area is greyed out.}
    \label{fig:full_page_wrong_exponential_all_lower}
\end{figure}

\begin{figure}[p]
    \centering
    \includegraphics[width=0.9\linewidth]{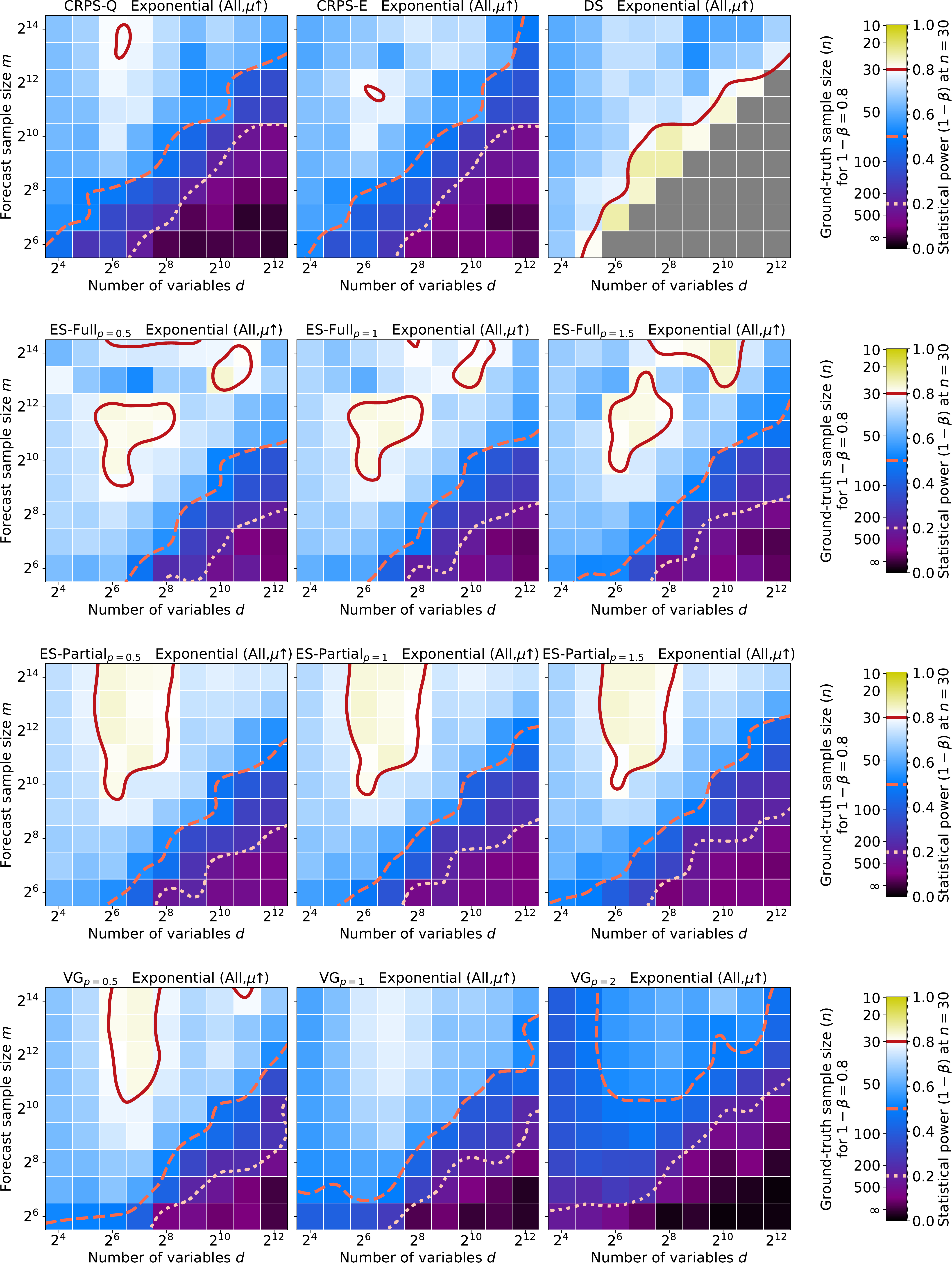}
    \caption{Statistical power of all scoring rules at correctly detecting that forecast's marginal means are higher than the ground-truth's ones for all dimensions, for Exponential distribution marginals.
    DS cannot be computed when $d \ge m$, so the corresponding area is greyed out.}
    \label{fig:full_page_wrong_exponential_all_higher}
\end{figure}

\begin{figure}[p]
    \centering
    \includegraphics[width=0.9\linewidth]{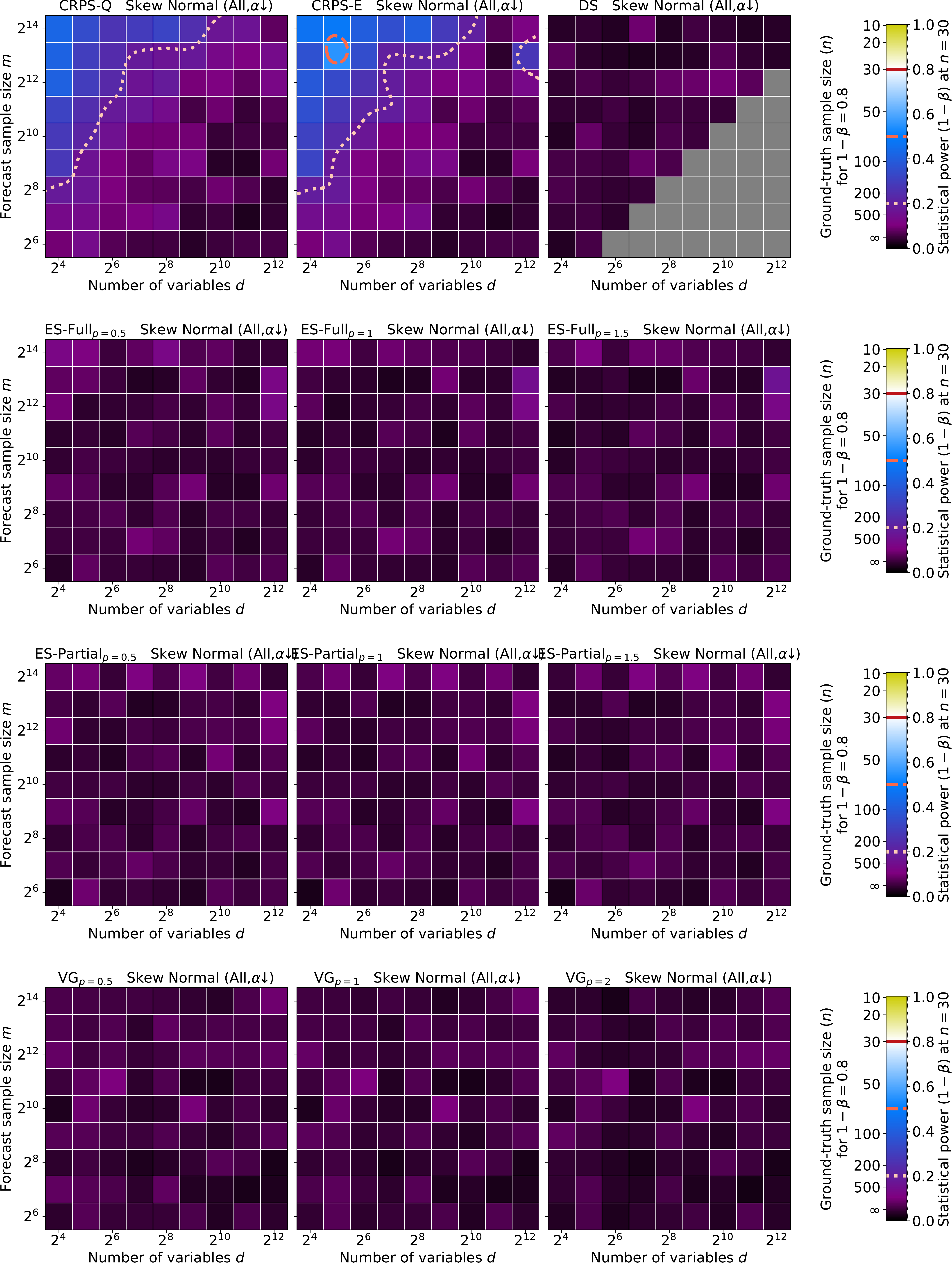}
    \caption{Statistical power of all scoring rules at correctly detecting that forecast's marginal skewness are lower than the ground-truth's ones for all dimensions, for Skew Normal distribution marginals.
    DS cannot be computed when $d \ge m$, so the corresponding area is greyed out.}
    \label{fig:full_page_missing_skew_all}
\end{figure}

\begin{figure}[p]
    \centering
    \includegraphics[width=0.9\linewidth]{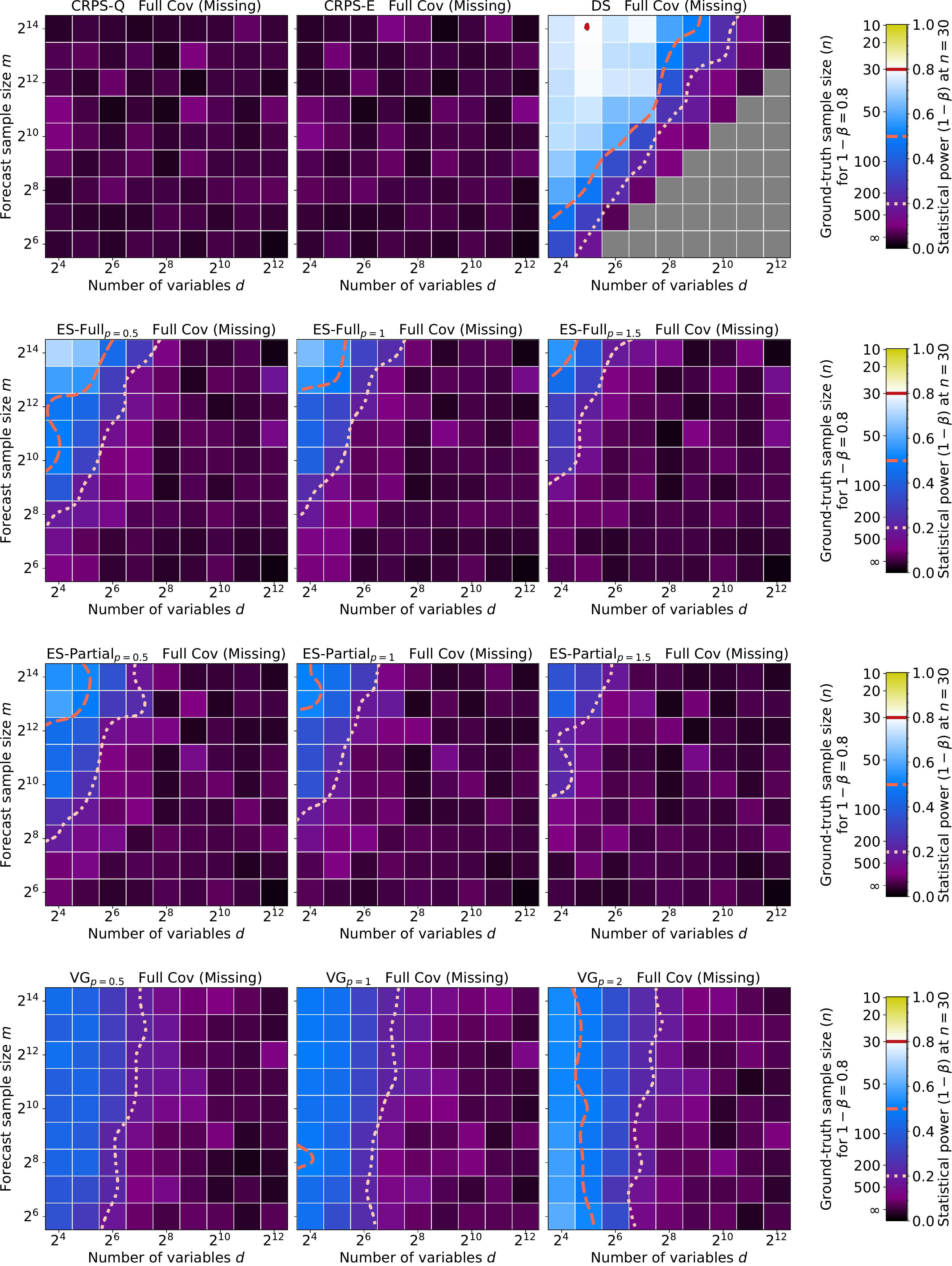}
    \caption{Statistical power of all scoring rules at correctly detecting that all variables are independent in the forecast, while they are all positively correlated in the ground truth, for Normal distribution marginals.
    DS cannot be computed when $d \ge m$, so the corresponding area is greyed out.}
    \label{fig:full_page_missing_covariance_full}
\end{figure}

\begin{figure}[p]
    \centering
    \includegraphics[width=0.9\linewidth]{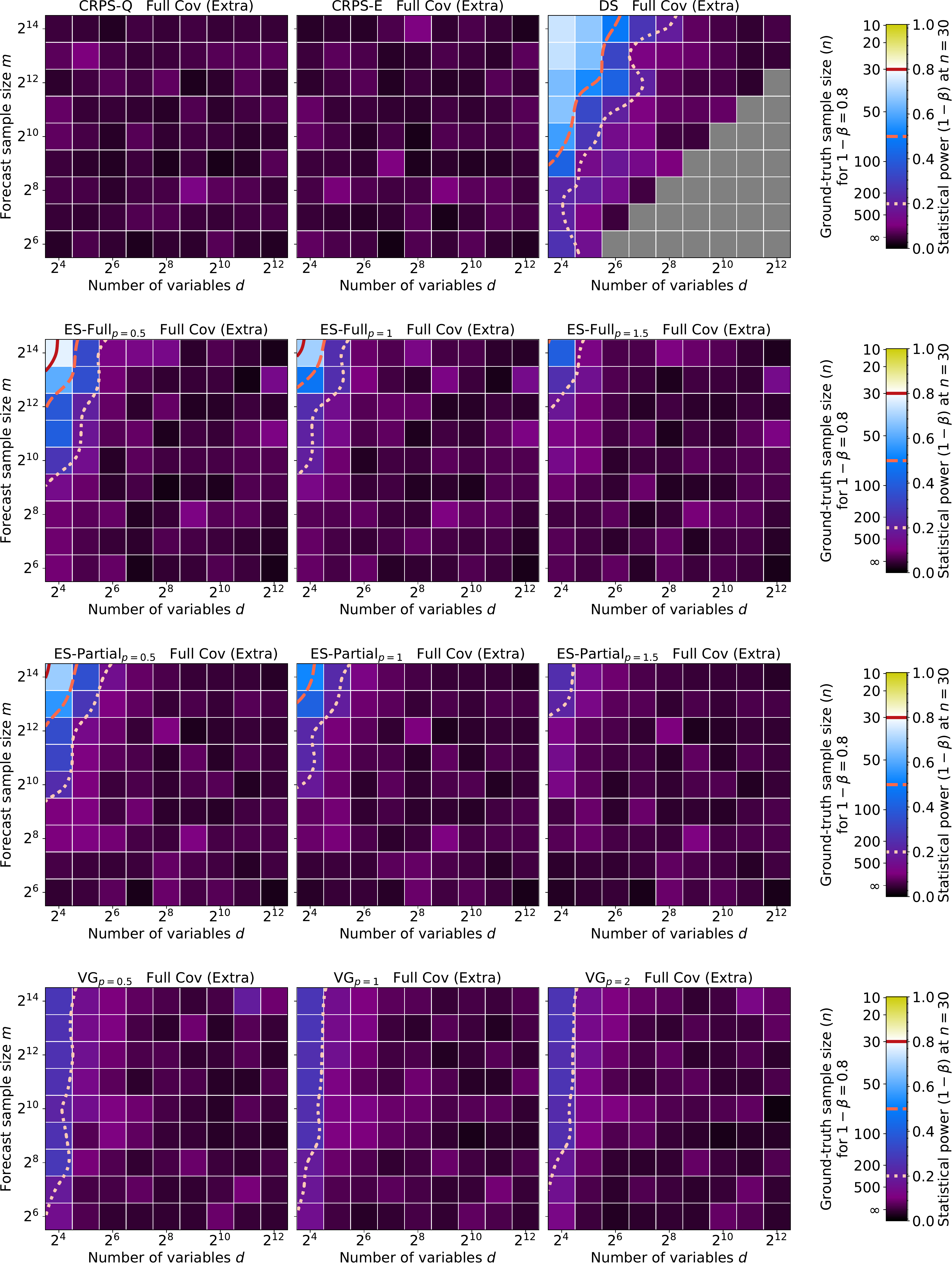}
    \caption{Statistical power of all scoring rules at correctly detecting that all variables are positively correlated in the forecast, while they are all independent in the ground truth, for Normal distribution marginals.
    DS cannot be computed when $d \ge m$, so the corresponding area is greyed out.}
    \label{fig:full_page_extra_covariance_full}
\end{figure}

\begin{figure}[p]
    \centering
    \includegraphics[width=0.9\linewidth]{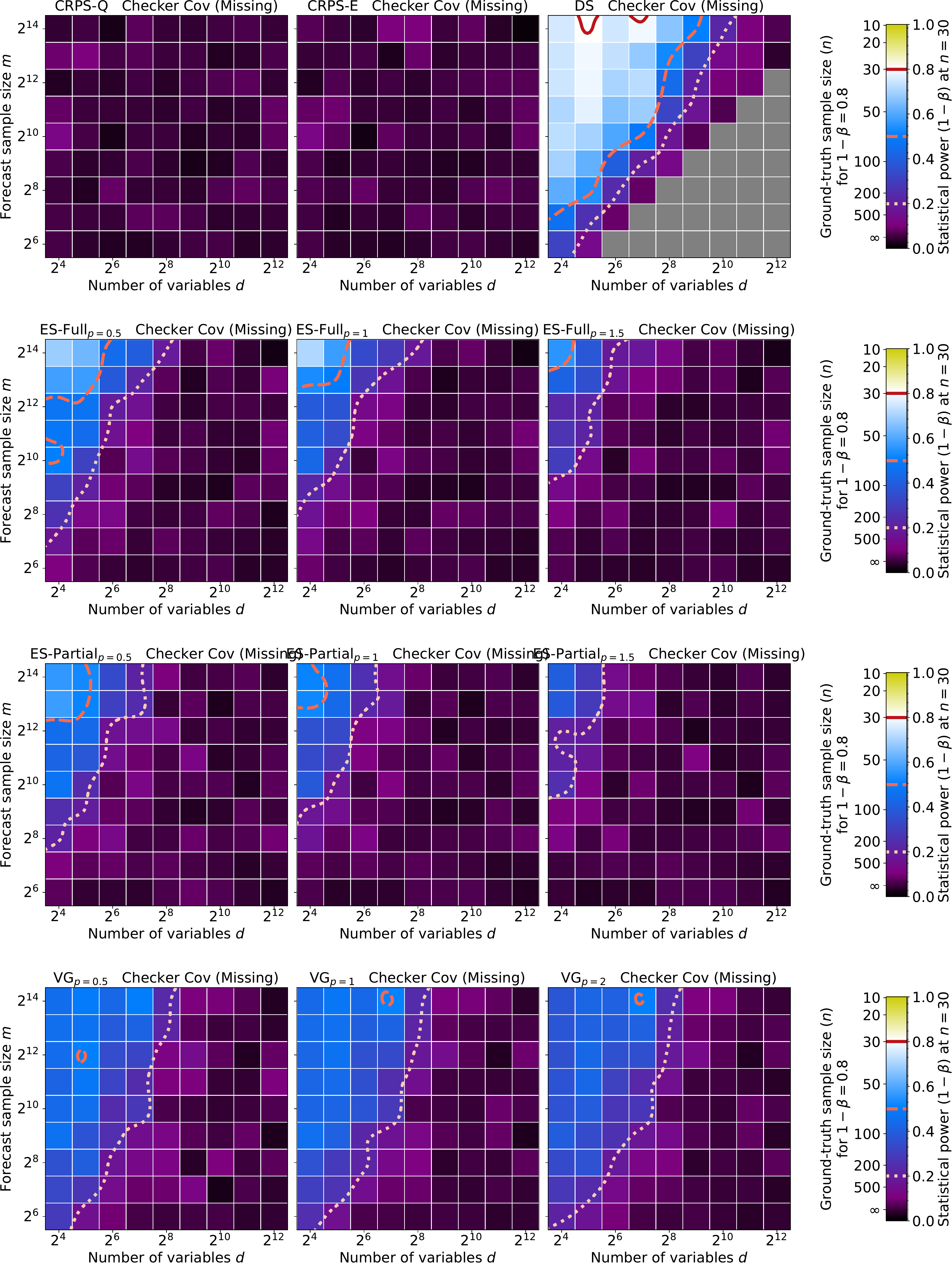}
    \caption{Statistical power of all scoring rules at correctly detecting that all variables are independent in the forecast, while they are all either positively or negatively correlated in the ground truth, for Normal distribution marginals.
    DS cannot be computed when $d \ge m$, so the corresponding area is greyed out.}
    \label{fig:full_page_missing_covariance_checker}
\end{figure}

\begin{figure}[p]
    \centering
    \includegraphics[width=0.9\linewidth]{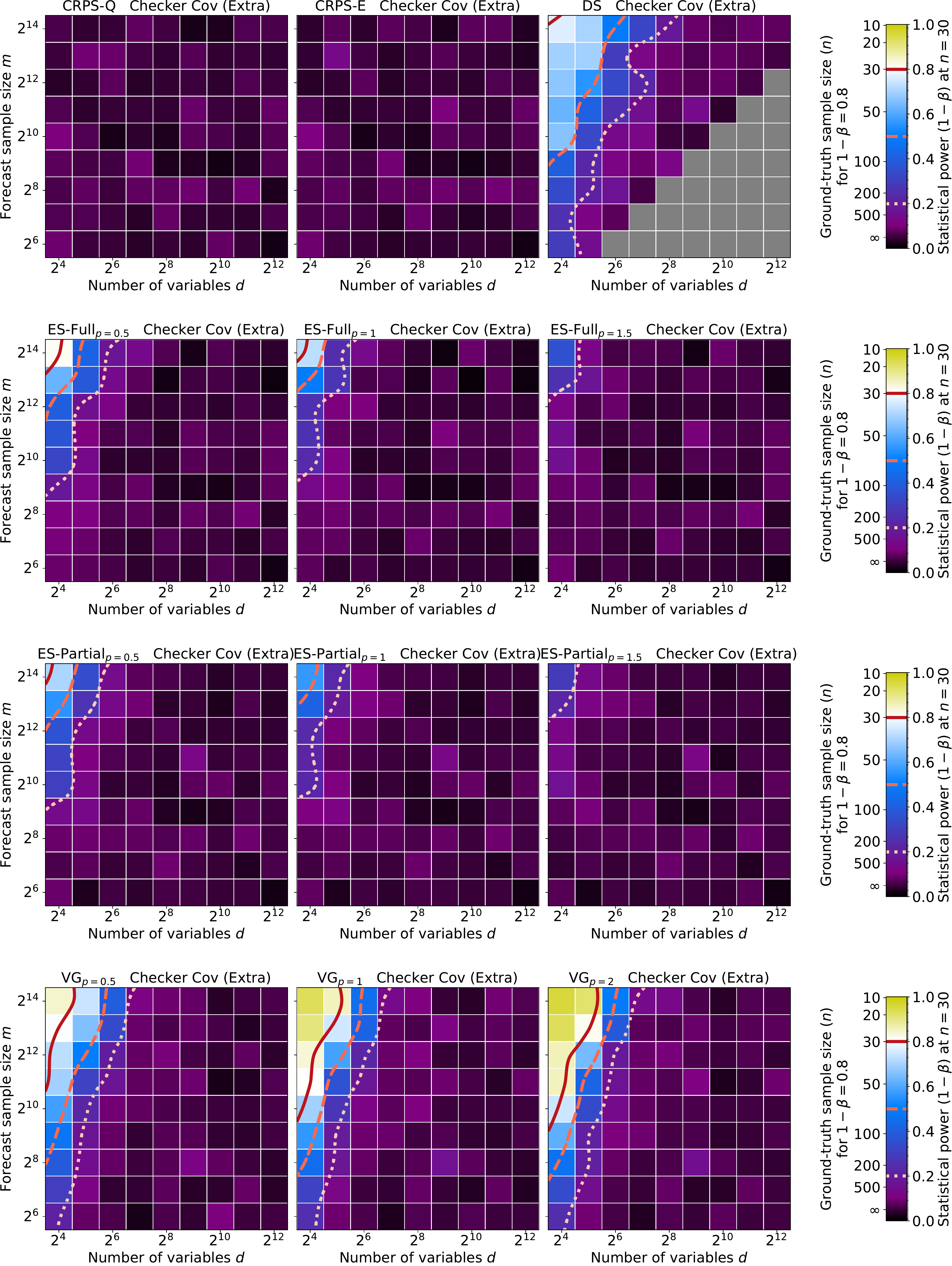}
    \caption{Statistical power of all scoring rules at correctly detecting that all variables are either positively or negatively correlated in the forecast, while they are all independent in the ground truth, for Normal distribution marginals.
    DS cannot be computed when $d \ge m$, so the corresponding area is greyed out.}
    \label{fig:full_page_extra_covariance_checker}
\end{figure}

\begin{figure}[p]
    \centering
    \includegraphics[width=0.9\linewidth]{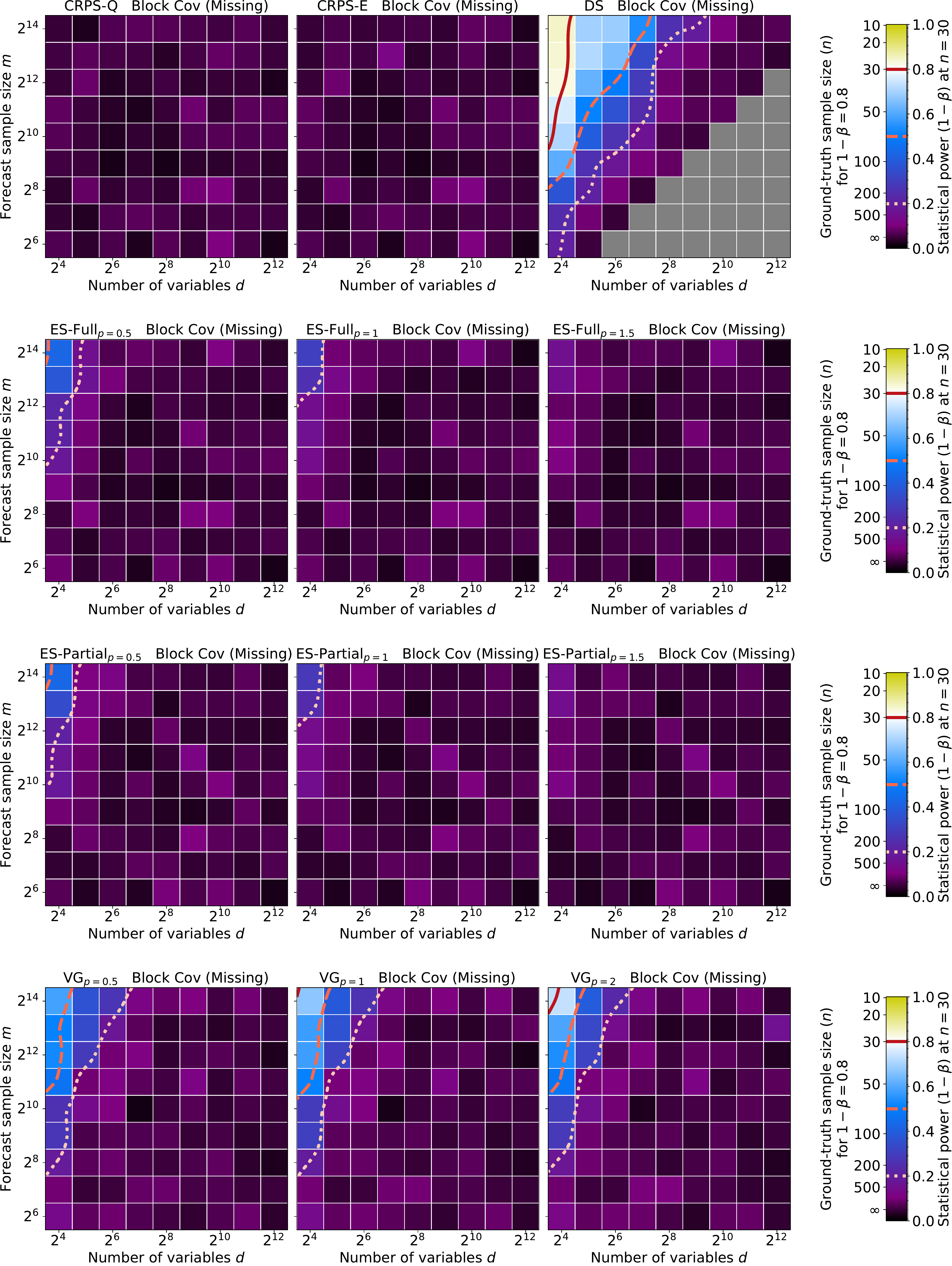}
    \caption{Statistical power of all scoring rules at correctly detecting that all variables are independent in the forecast, while each pair are positively correlated in the ground truth, for Normal distribution marginals.
    DS cannot be computed when $d \ge m$, so the corresponding area is greyed out.}
    \label{fig:full_page_missing_covariance_block}
\end{figure}

\begin{figure}[p]
    \centering
    \includegraphics[width=0.9\linewidth]{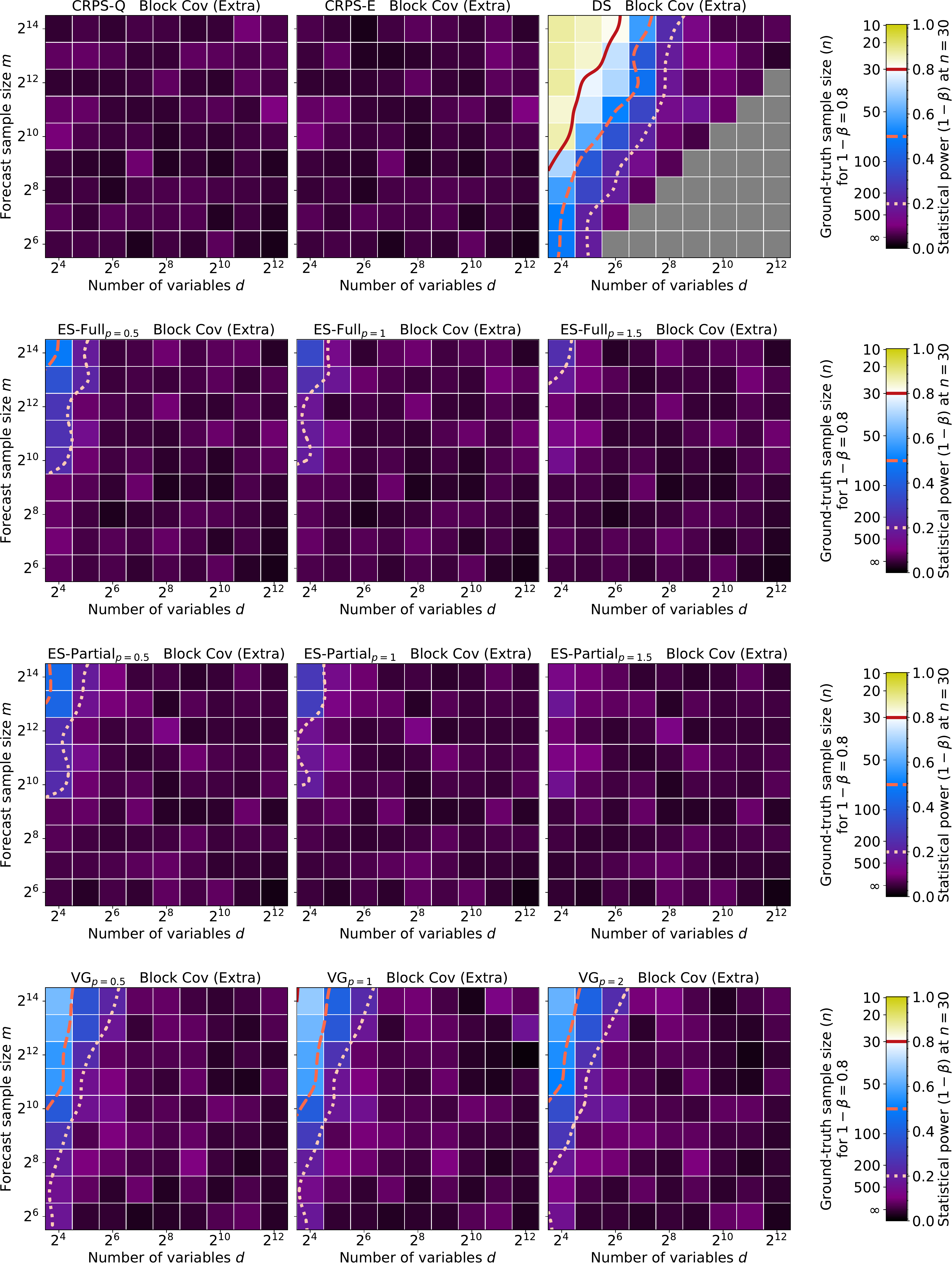}
    \caption{Statistical power of all scoring rules at correctly detecting that all pairs of variables are either positively correlated in the forecast, while they are all independent in the ground truth, for Normal distribution marginals.
    DS cannot be computed when $d \ge m$, so the corresponding area is greyed out.}
    \label{fig:full_page_extra_covariance_block}
\end{figure}

\begin{figure}[p]
    \centering
    \includegraphics[width=0.9\linewidth]{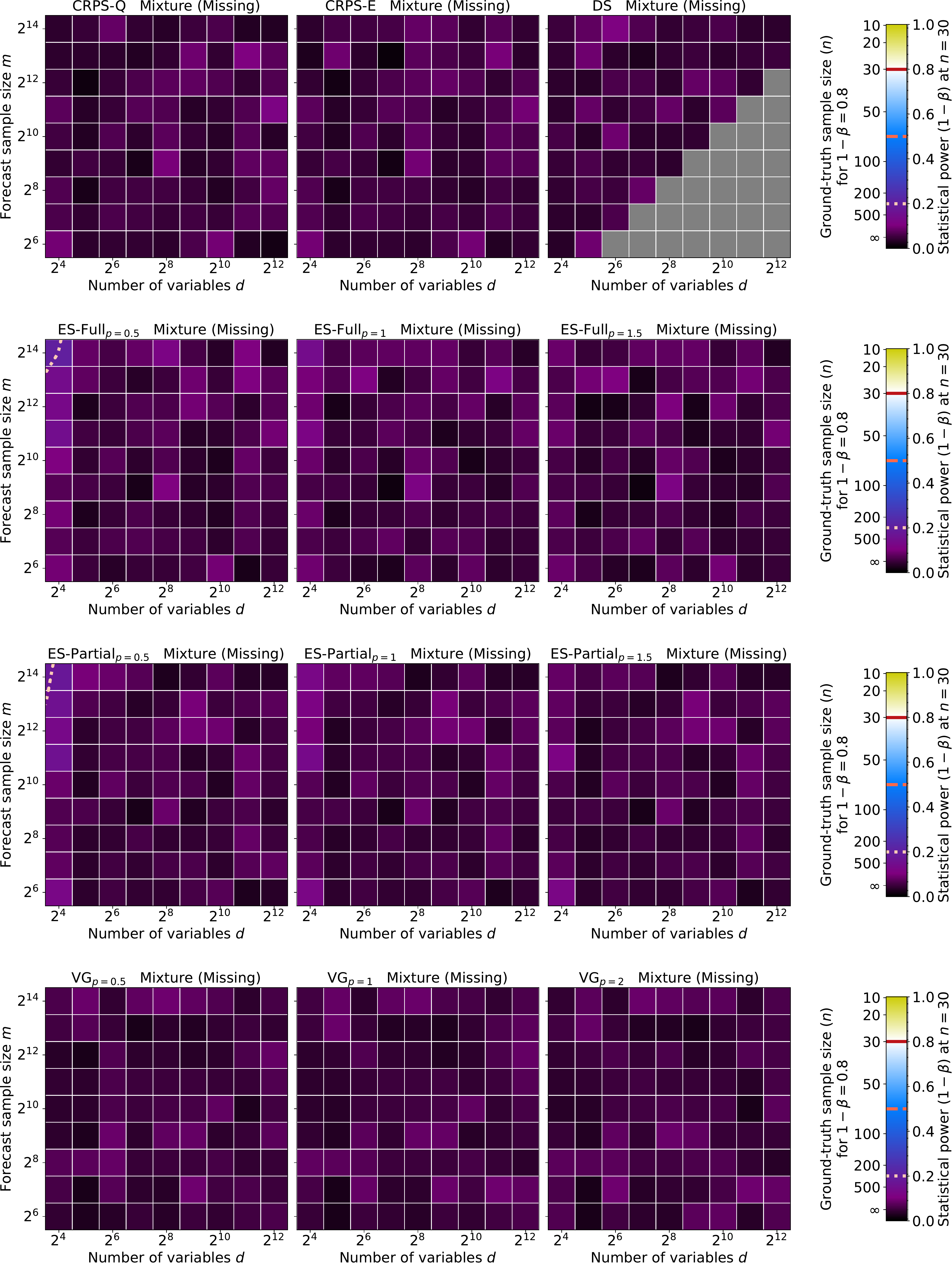}
    \caption{Statistical power of all scoring rules at correctly detecting that the forecast's distribution only contains a single mode, while the ground-truth's one contains two.
    DS cannot be computed when $d \ge m$, so the corresponding area is greyed out.}
    \label{fig:full_page_missing_mixture}
\end{figure}

\begin{figure}[p]
    \centering
    \includegraphics[width=0.9\linewidth]{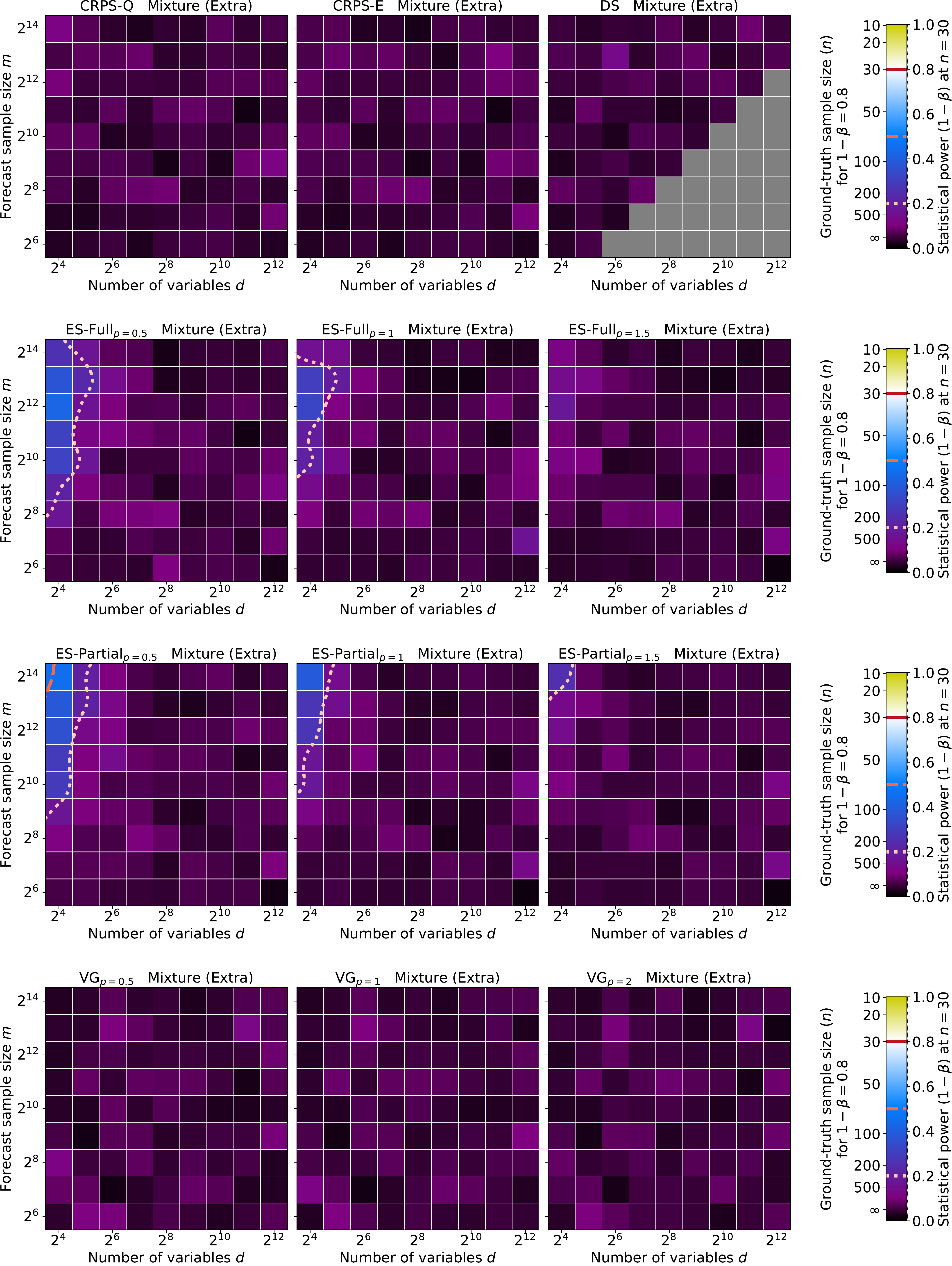}
    \caption{Statistical power of all scoring rules at correctly detecting that the forecast's distribution contains two modes, while the ground-truth's one only contains a single one.
    DS cannot be computed when $d \ge m$, so the corresponding area is greyed out.}
    \label{fig:full_page_extra_mixture}
\end{figure}

\subsection{Validity of normality assumption} \label{app:validity_normal_assumption}

\begin{figure}[H]
    \centering
    \includegraphics[width=0.9\linewidth]{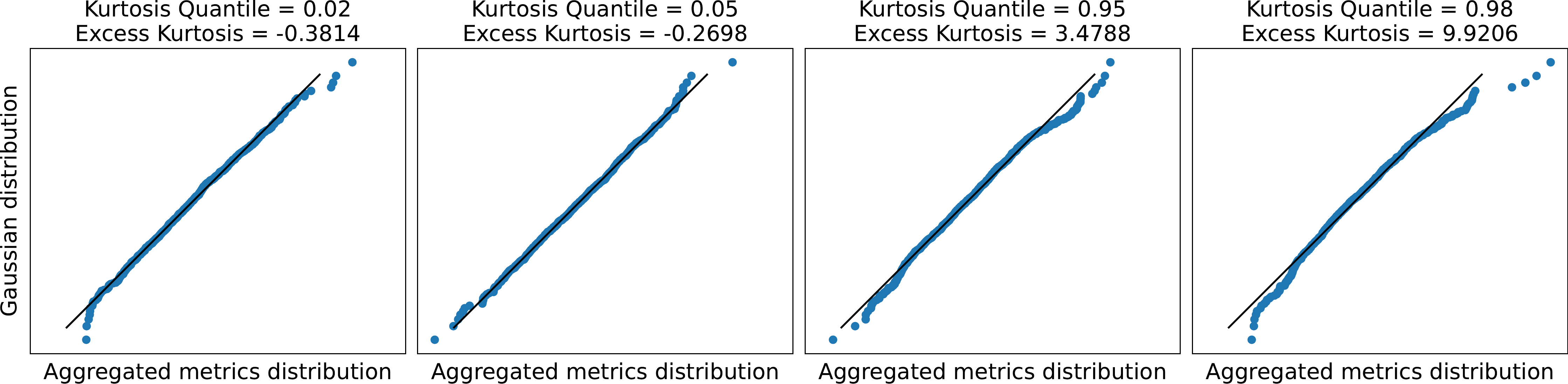}
    \caption{Q-Q plot comparing the Gaussian distribution and distribution we obtain by averaging $n=30$ independent $\Delta_m$ values from our experiments. The experiments shown are selected based on their rank when sorted by $\Delta_m$ excess kurtosis.}
    \label{fig:gaussian_check}
\end{figure}

In the statistical analysis of our results detailed in \cref{sec:power}, we have opted to assume that taking an ensemble of $n=30$ independent replications of the gap between the ground truth and forecasting scores $\Delta_m$ is sufficient to use the Normal distribution in our computations.
To test this hypothesis for a single experiment, we can use a Quantile-Quantile plot, where we compare the distribution of the ensemble average of $n$ independent $\Delta_m$, and theoretical Gaussian distribution of equal mean and standard deviation.
Instead of randomly picking from our experiments, we select a few of them according to the excess kurtosis of $\Delta_m$.
\cref{fig:gaussian_check} shows the Q-Q plot for 4 experiments, selected such that their excess kurtosis prior to ensembling are at quantiles 0.02, 0.05, 0.95, and 0.98.
From these, we can see that only the experiments with the highest 2\% of kurtosis suffer from a visible deviation between both distributions.
This means that our normality assumption is reasonable for our experiments.

\subsection{Effects of Modifying the Experiments Parameters}

In our main results, all of our statistical analysis has been done with a constant significance level $\alpha=0.05$, and our experiment tuning with a constant statistical power of $1 - \beta_{NLL} = 0.8$ for the NLL.
Since these two choices are arbitrary, we show in this section that our conclusions still hold even if we repeat our experiments with other values for $\alpha$ and $\beta$.

\begin{figure}[p]
    \centering
    \includegraphics[width=0.9\linewidth]{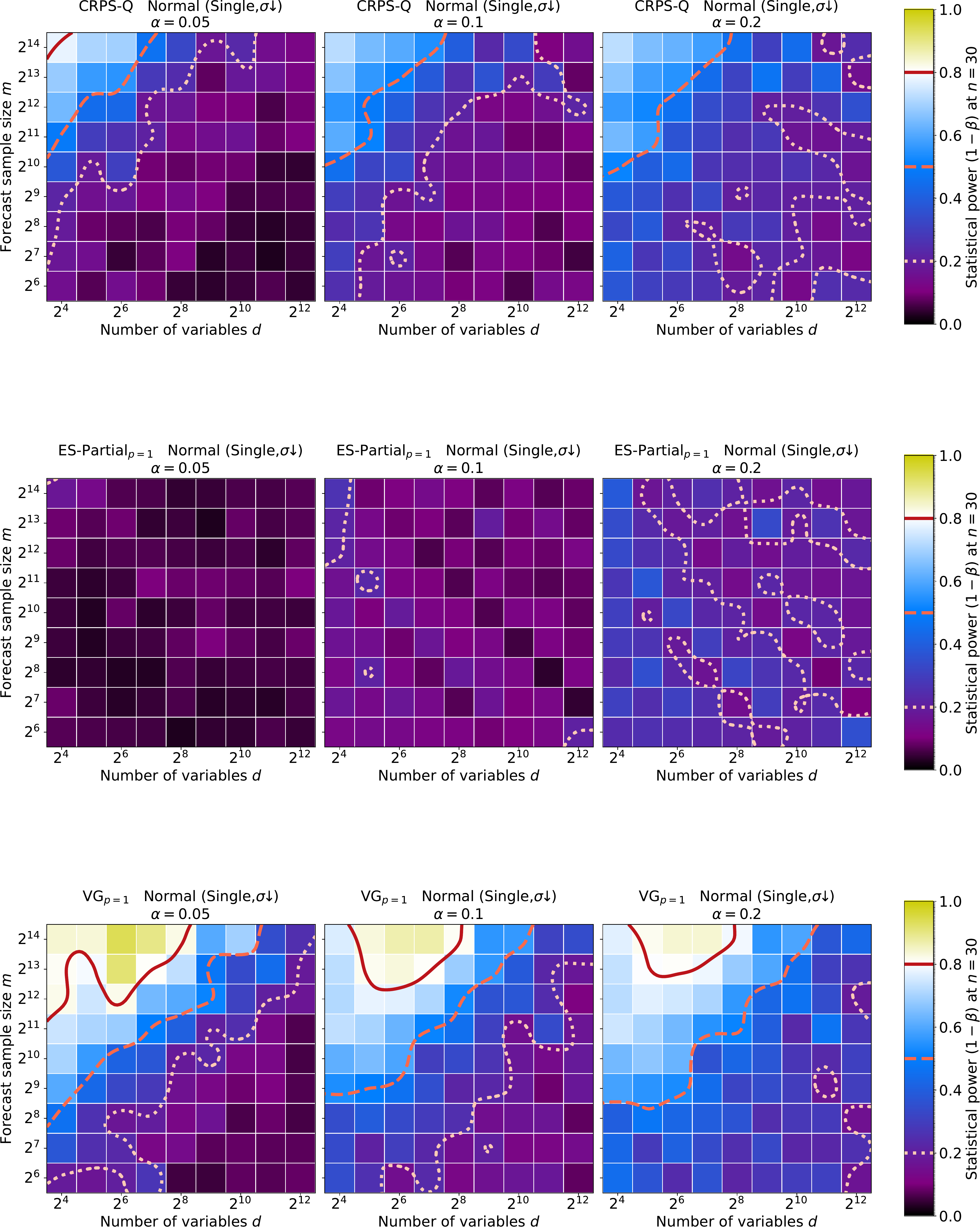}
    \caption{Statistical power of a subset of scoring rules at correctly detecting that forecast’s marginal standard deviation is lower than the ground-truth’s one for a single dimension, for Normal distribution marginals. The significance level $\alpha$ is varied for both the NLL tuning and the scoring rules under test.
    }
    \label{fig:full_page_alpha_wrong_std_single_lower}
\end{figure}

\begin{figure}[p]
    \centering
    \includegraphics[width=0.9\linewidth]{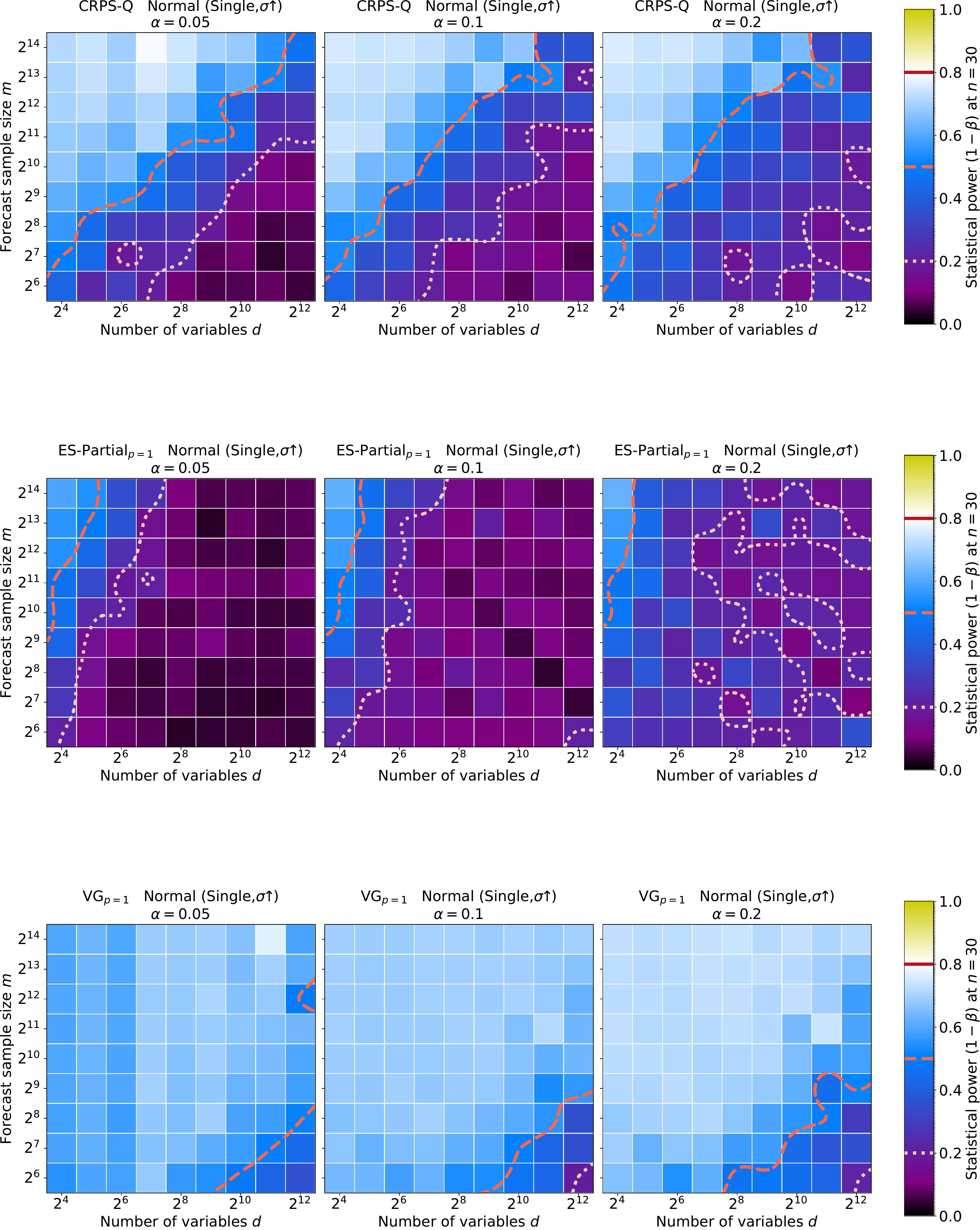}
    \caption{Statistical power of a subset of scoring rules at correctly detecting that forecast’s marginal standard deviation is higher than the ground-truth’s one for a single dimension, for Normal distribution marginals. The significance level $\alpha$ is varied for both the NLL tuning and the scoring rules under test.
    }
    \label{fig:full_page_alpha_wrong_std_single_higher}
\end{figure}

\begin{figure}[p]
    \centering
    \includegraphics[width=0.9\linewidth]{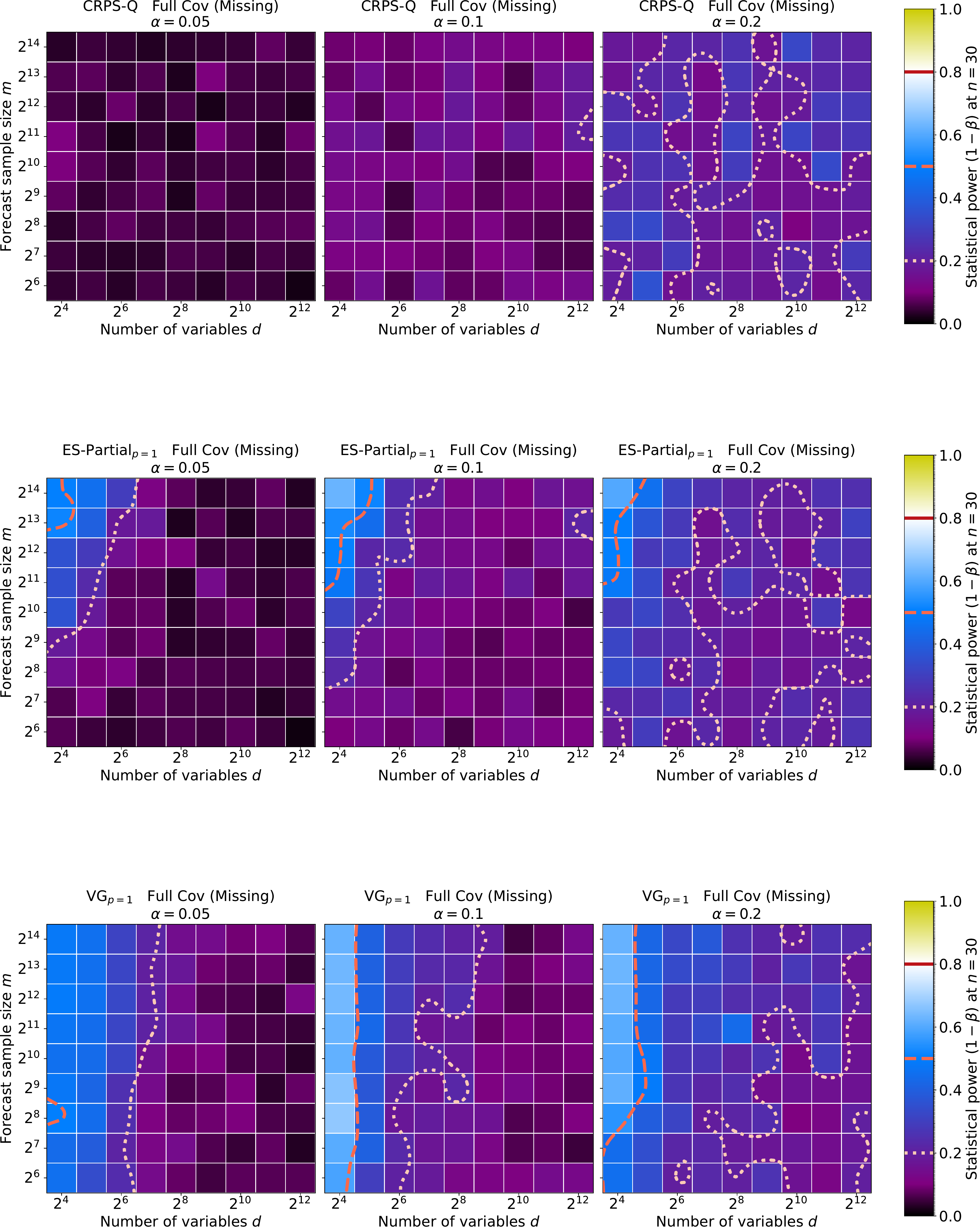}
    \caption{Statistical power of a subset of scoring rules at correctly detecting that all variables are independent in the forecast, while they are all positively correlated in the ground truth, for Normal distribution marginals. The significance level $\alpha$ is varied for both the NLL tuning and the scoring rules under test.
    }
    \label{fig:full_page_alpha_missing_covariance_full}
\end{figure}

\begin{figure}[p]
    \centering
    \includegraphics[width=0.9\linewidth]{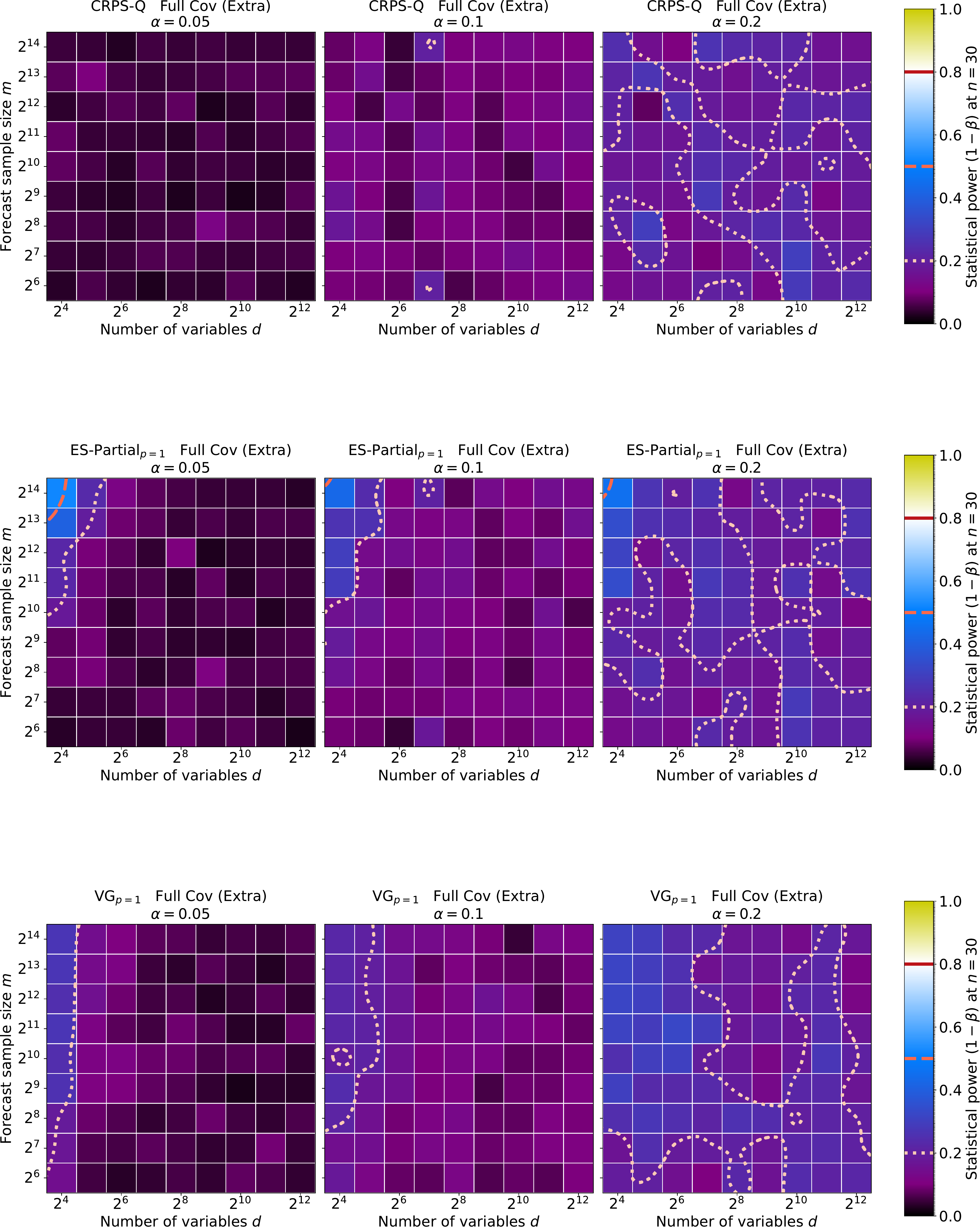}
    \caption{Statistical power of a subset of scoring rules at correctly detecting that all variables are positively correlated in the forecast, while they are all independent in the ground truth, for Normal distribution marginals. The significance level $\alpha$ is varied for both the NLL tuning and the scoring rules under test.
    }
    \label{fig:full_page_alpha_extra_covariance_full}
\end{figure}

\cref{fig:full_page_alpha_wrong_std_single_lower,fig:full_page_alpha_wrong_std_single_higher,fig:full_page_alpha_missing_covariance_full,fig:full_page_alpha_extra_covariance_full} show the impact of varying $\alpha$ on the ability of CRPS-Q, ES-Partial$_p=1$, and VG$_p=1$ to distinguish a forecast distribution from a ground-truth distribution when they differ only by their covariance.
From these, we can observe that the regions of high statistical power ($1 - \beta \ge 0.5$) stay untouched by the increased in $\alpha$.
While increasing $\alpha$ does increase statistical power, this effect is perfectly balanced by the tuning procedure in those regions.
On the flip side, regions of low statistical power saw an increase in said power.
This is expected since for purely random scoring rules, the statistical power is equal to $\alpha$.

\begin{figure}[p]
    \centering
    \includegraphics[width=0.9\linewidth]{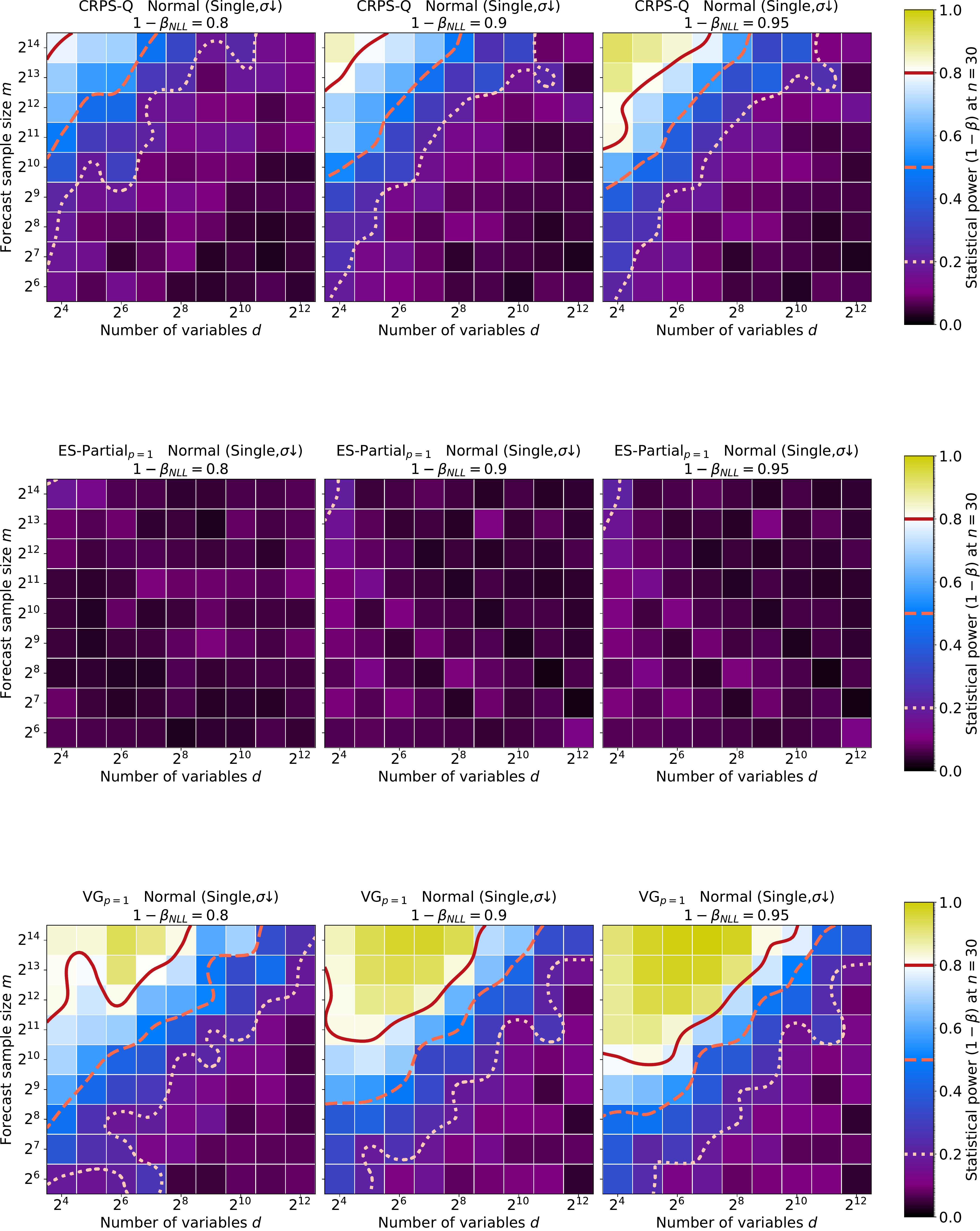}
    \caption{Statistical power of a subset of scoring rules at correctly detecting that forecast’s marginal standard deviation is lower than the ground-truth’s one for a single dimension, for Normal distribution marginals. The target statistical power $1 - \beta_{NLL}$ is varied for the NLL tuning.
    }
    \label{fig:full_page_beta_wrong_std_single_lower}
\end{figure}

\begin{figure}[p]
    \centering
    \includegraphics[width=0.9\linewidth]{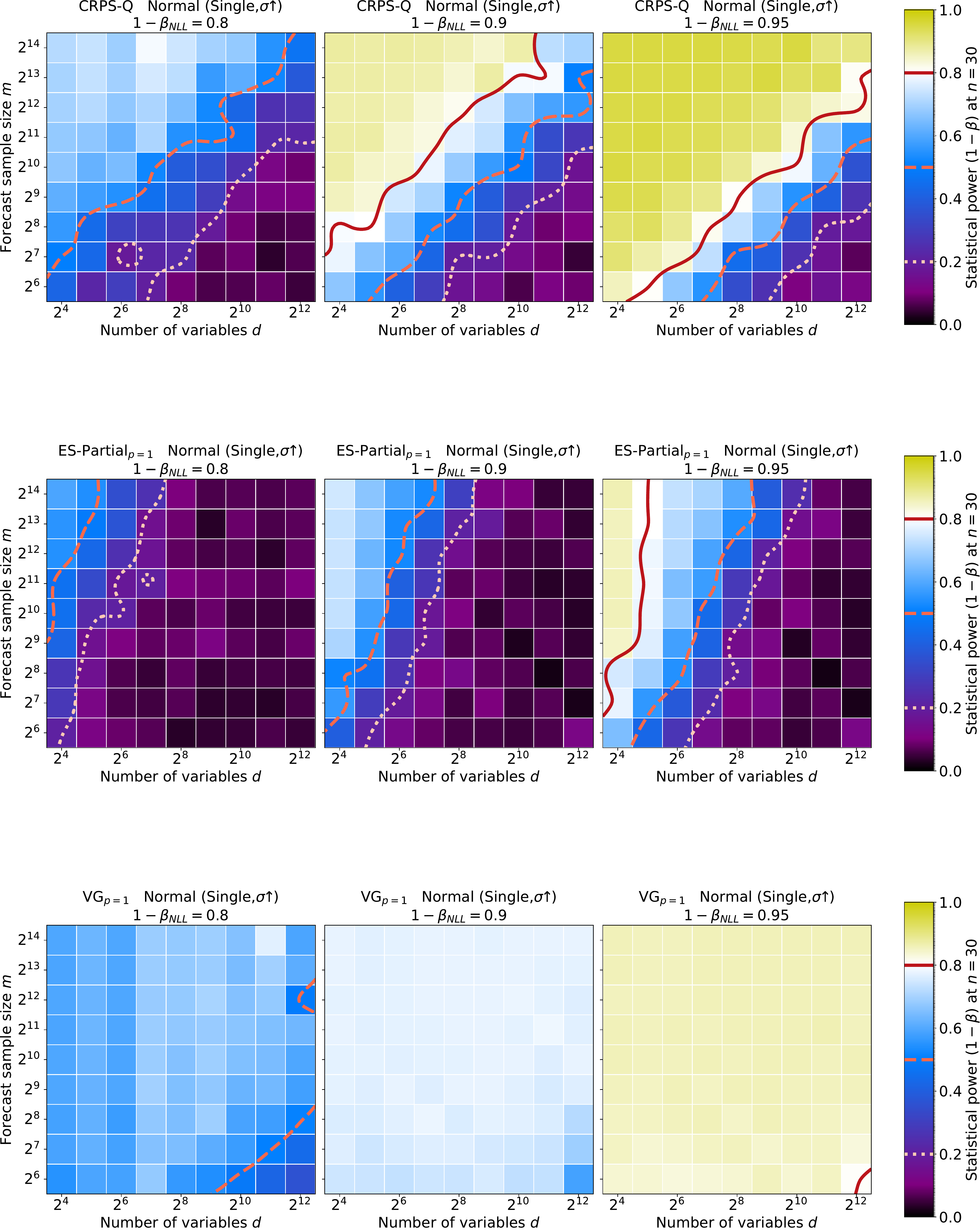}
    \caption{Statistical power of a subset of scoring rules at correctly detecting that forecast’s marginal standard deviation is higher than the ground-truth’s one for a single dimension, for Normal distribution marginals. The target statistical power $1 - \beta_{NLL}$ is varied for the NLL tuning.
    }
    \label{fig:full_page_beta_wrong_std_single_higher}
\end{figure}

\begin{figure}[p]
    \centering
    \includegraphics[width=0.9\linewidth]{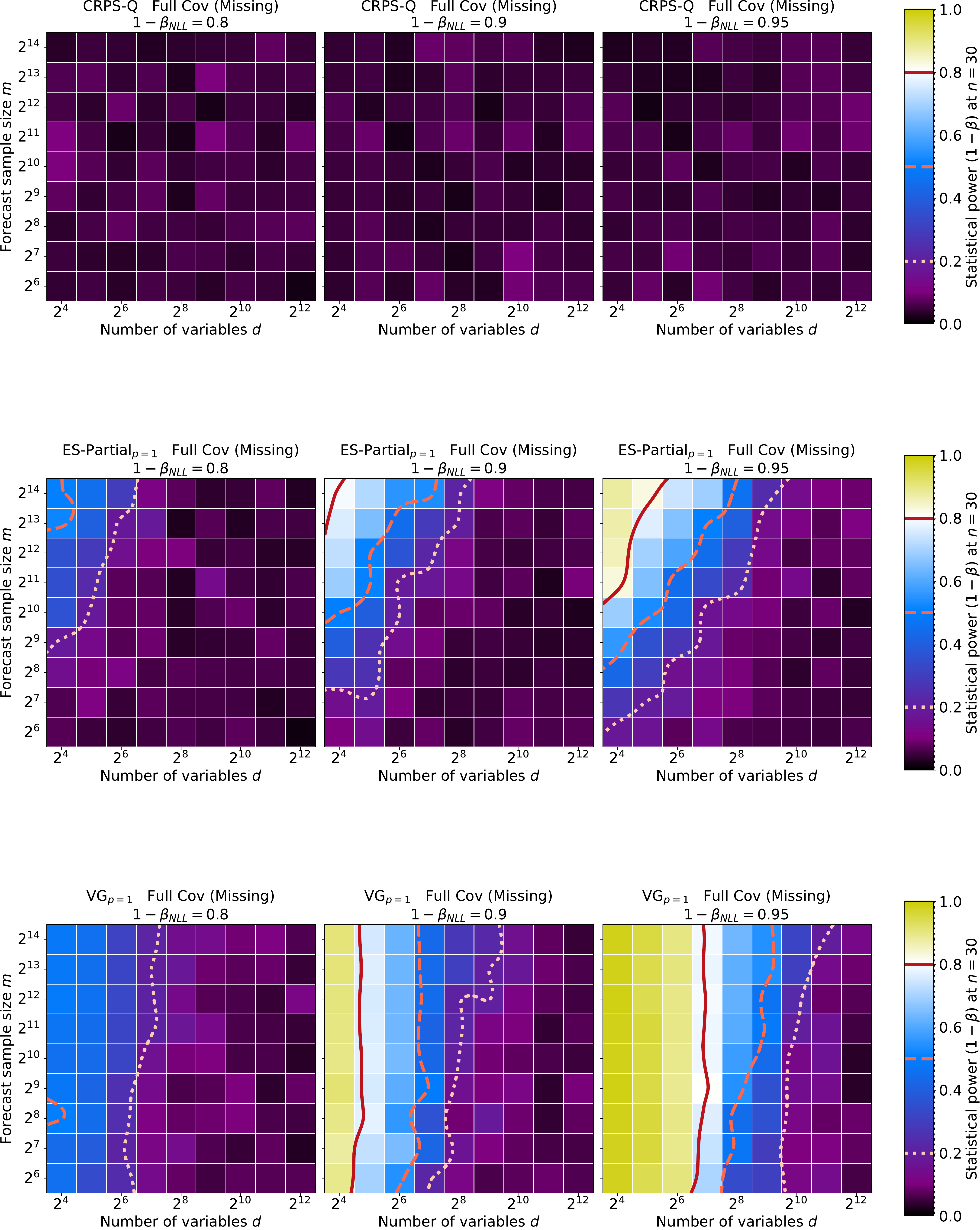}
    \caption{Statistical power of a subset of scoring rules at correctly detecting that all variables are independent in the forecast, while they are all positively correlated in the ground truth, for Normal distribution marginals. The target statistical power $1 - \beta_{NLL}$ is varied for the NLL tuning.
    }
    \label{fig:full_page_beta_missing_covariance_full}
\end{figure}

\begin{figure}[p]
    \centering
    \includegraphics[width=0.9\linewidth]{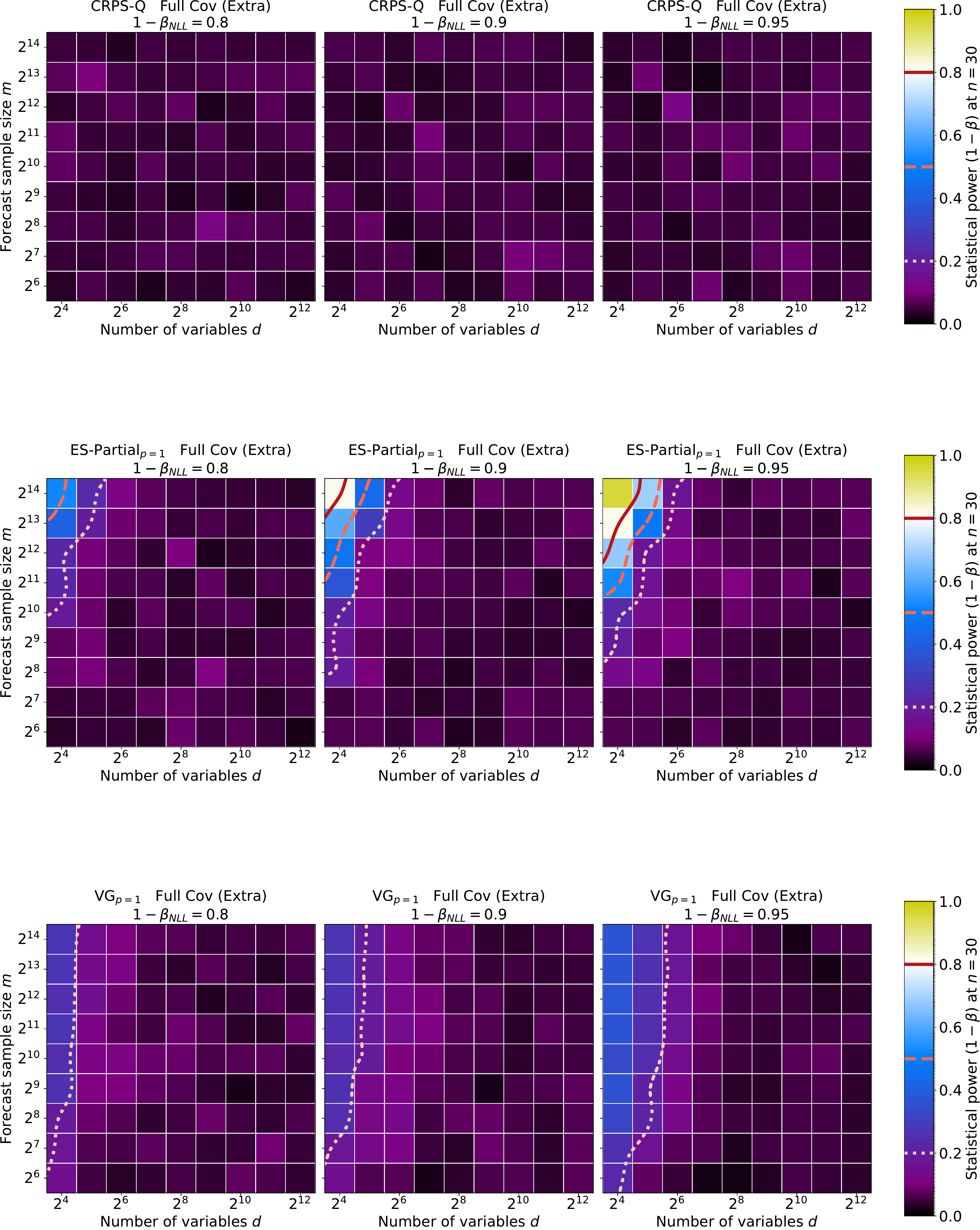}
    \caption{Statistical power of a subset of scoring rules at correctly detecting that all variables are positively correlated in the forecast, while they are all independent in the ground truth, for Normal distribution marginals. The target statistical power $1 - \beta_{NLL}$ is varied for the NLL tuning.
    }
    \label{fig:full_page_beta_extra_covariance_full}
\end{figure}

\cref{fig:full_page_beta_wrong_std_single_lower,fig:full_page_beta_wrong_std_single_higher,fig:full_page_beta_missing_covariance_full,fig:full_page_beta_extra_covariance_full} show the impact of varying $1 - \beta_{NLL}$ on the ability of CRPS-Q, ES-Partial$_p=1$, and VG$_p=1$ to distinguish a forecast distribution from a ground-truth distribution when they differ only by their covariance.
To reach an increased targeted statistical power for the NLL, the tuning requires the forecast and ground-truth distributions to be more dissimilar.
As expected, these pairs of distributions are also easier to distinguish by the scoring rules under tests, leading to better reliability.
However, the shapes of the various regions of reliability remain unchanged, so these results do not contradict our earlier conclusions.

\section{Details on Real-Data Experiments}\label{app:real-data}

For our real-data experiments, we started with TACTiS \cite{pmlr-v162-drouin22a}, a state-of-the-art timeseries forecasting model, trained on the 10-minute increment version of the \texttt{solar} dataset.
We chose this model since it allows us to compute the NLL of multiple perturbations of the forecast, including making all variables independent or adding multiplication or additive biases.
Using this model, we generated a multivariate stochastic forecast over a 12 hours period (thus 72 time steps), with a sample of size 100.

The impact of the perturbations on the revenues is computed with the model presented in \cref{eq:motivating}, with $M=50$, and $\lambda=10$.
This model can readily be converted to a Mixed Integer Programming one, and be solved exactly by a wide range of solvers.

The NLL is computed directly on each element from this sample, which we take as the ground truth sample.
However, since it is numerically prohibitive to get independent samples for the other scoring rules, we have to reuse the ground truth sample to generate the forecasting sample.
Thus, for each element from the ground truth sample, we use all other elements to generate the forecasting sample, after applying the perturbation.
To break the correlations in a sample, we shuffle the values of each variable across the sample, which keeps the marginals intact.
It should be noted that due to reusing the same sample to generate each forecasting sample, we ignore the extra variance in the scoring rules due to the sampling process, thus our estimates of $1 - \beta$ for them are biased toward higher values.
The statistical power for these experiments is shown in \cref{tab:app_real_data_solar}, but rewritten to put the emphasis on how close the NLL statistical power is to 1, which would only be possible for a perfect scoring rule with distributions with no overlap.

\subsection{Additional Remarks}
\paragraph{Multiplying by a Constant} 
The closest analog to this test case in our benchmark is \textbf{Exponential (All, $\mu \uparrow$)} since multiplying an exponential variable by some factor is the same as multiplying its mean by the same factor.
Alternatively we could have \textbf{\textbf{\textbf{\textbf{\textbf{\textbf{considered}}}}}} our test cases \textbf{Normal (All, $\mu \uparrow$)} and \textbf{Normal (All, $\sigma \uparrow$)} jointly.

\begin{table}
\centering
\caption{Statistical power $1-\beta$ for the ability of various scoring rules to distinguish a ground-truth distribution based on the \texttt{solar} dataset and a perturbation of said distribution.}
\label{tab:app_real_data_solar}
\begin{tabular}{lrrr}
\toprule
\textbf{Scoring rule} &   \textbf{Breaking correlations}   &  \textbf{Multiplying by 1.05} & \textbf{Adding 0.05} \\
\midrule
NLL & $1 - 5.6 \times 10^{-7728}$ & $1 - 1.6 \times 10^{-43}$ & $1 - 1.1 \times 10^{-200}$ \\
CRPS-Q & $0.05$ & $0.37$ & $1 - 2.2 \times 10^{-28}$ \\
ES-Partial$_{p=1}$ & $1 - 1.7 \times 10^{-3}$ & $0.30$ & $0.10$ \\
VG$_{p=1}$ & $0.99$ & $0.30$ & $0.05$ \\
\bottomrule
\end{tabular}
\end{table}

\subsection{Further Real-Data Experiments}\label{app:more_real_data}

To check whether the previous results are valid for other data than the \texttt{solar} dataset, we did the experiment with the \texttt{kdd-cup} (an air pollution dataset with hourly increments, 48 hours forecast, and $d=12960$) and \texttt{electricity} (an electricity dataset with hourly increments, 24 hours forecast, and $d=8880$) datasets.
We kept the multiplication perturbation factor equal to the one used for the \texttt{solar} dataset ($1.05$).
However, since the datasets do not share the same scale, we adjusted the additive perturbation term to roughly keep a constant ratio to the average value of the data.
This gave an additive term of $0.5$ for \texttt{kdd-cup} and $40$ for \texttt{electricity}.

The statistical power for these two experiments is shown in \cref{tab:app_real_data_kdd_cup,tab:app_real_data_electricity}.
The dominance of the NLL over the other scoring rules remains, although the three other scoring rules had very high statistical power for the \texttt{electricity} dataset when perturbing the data with either a multiplicative factor or an additive term.
On the flip side, the energy score and the variogram lost much of the statistical power they showed for the correlation-breaking perturbation on the \texttt{solar} dataset.

\subsection{Comparison with Previous Work}

Our approach to testing scoring rules using real-world data is similar to the one in \citet{Alexander2022} in that we both use forecasting models trained on real-world data as the ground truth.
Their differs from ours in how they create the erroneous forecasts in that they use alternative forecasting models to do so, while we apply explicit transformations.
Both approaches are valid, in that theirs is closer to how forecast inaccuracies will behave while ours is easier to interpret.

\begin{table}
\centering
\caption{Statistical power $1-\beta$ for the ability of various scoring rules to distinguish a ground-truth distribution based on the \texttt{kdd-cup} dataset and a perturbation of said distribution.}
\label{tab:app_real_data_kdd_cup}
\begin{tabular}{lrrr}
\toprule
\textbf{Scoring rule} &   \textbf{Breaking correlations}   &  \textbf{Multiplying by 1.05} & \textbf{Adding 0.5} \\
\midrule
NLL & $1 - 5.8 \times 10^{-301}$ & $1 - 7.0 \times 10^{-40010}$ & $1 - 3.0 \times 10^{-19042}$ \\
CRPS-Q & $0.05$ & $0.61$ & $0.96$ \\
ES-Partial$_{p=1}$ & $0.34$ & $0.24$ & $0.08$ \\
VG$_{p=1}$ & $0.30$ & $0.24$ & $0.05$ \\
\bottomrule
\end{tabular}
\end{table}

\begin{table}
\centering
\caption{Statistical power $1-\beta$ for the ability of various scoring rules to distinguish a ground-truth distribution based on the \texttt{electricity} dataset and a perturbation of said distribution.}
\label{tab:app_real_data_electricity}
\begin{tabular}{lrrr}
\toprule
\textbf{Scoring rule} &   \textbf{Breaking correlations}   &  \textbf{Multiplying by 1.05} & \textbf{Adding 40} \\
\midrule
NLL & $1 - 3.3 \times 10^{-3274}$ & $1 - 3.7 \times 10^{-11560}$ & $1 - 1.7 \times 10^{-222}$ \\
CRPS-Q & $0.05$ & $1 - 2.5 \times 10^{-30}$ & $1 - 7.3 \times 10^{-233}$ \\
ES-Partial$_{p=1}$ & $0.35$ & $1 - 5.1 \times 10^{-3}$ & $1 - 7.8 \times 10^{-3}$ \\
VG$_{p=1}$ & $0.71$ & $0.98$ & $0.05$ \\
\bottomrule
\end{tabular}
\end{table}

\end{document}